%% file: neurips_2026.tex
\pdfoutput=1
\documentclass{article}

\PassOptionsToPackage{numbers, compress, sort}{natbib}
\usepackage[main, final]{neurips_2026}

\usepackage[utf8]{inputenc} %
\usepackage[T1]{fontenc}    %
\usepackage{hyperref}       %
\usepackage{latexml}%
\usepackage{url}            %
\usepackage{booktabs}       %
\usepackage{amsfonts}       %
\usepackage{nicefrac}       %
\usepackage{microtype}      %
\usepackage{xcolor}         %

\usepackage{amsmath}
\usepackage{amssymb}
\usepackage{mathtools}
\usepackage{amsthm}

\usepackage{algorithm}
\usepackage{algorithmic}

\usepackage{graphicx}
\usepackage{subcaption}
\usepackage{makecell}
\usepackage{dsfont}
\usepackage{adjustbox}
\usepackage{etoolbox}
\usepackage{enumitem}
\iflatexml\else\usepackage{etoc}\fi%
\usepackage{placeins}
\usepackage{multirow}
\iflatexml\renewcommand{\multirow}[3]{#3}\fi%
\usepackage{wrapfig}
\usepackage{caption}
\iflatexml\usepackage{array}\newcolumntype{S}[1][]{c}\else\usepackage{siunitx}\fi%
\usepackage{colortbl}
\usepackage{wasysym}
\usepackage{pifont}

\usepackage[capitalize,noabbrev]{cleveref}

\iflatexml\else\usepackage[disable,textsize=tiny]{todonotes}\fi%

\iflatexml\else
\fi
\iflatexml\let\lxpm\pm\renewcommand{\pm}{\ensuremath{\lxpm}}\fi%

\theoremstyle{plain}
\newtheorem{theorem}{Theorem} %
\newtheorem{proposition}[theorem]{Proposition}

\theoremstyle{definition}

\theoremstyle{remark}

\newtheorem*{lemma*}{Lemma}
\newtheorem*{remark*}{Remark}

\newcounter{appitem}

\theoremstyle{plain}
\newtheorem{appproposition}[appitem]{Proposition}
\newtheorem{applemma}[appitem]{Lemma}
\newtheorem{appcorollary}[appitem]{Corollary}

\theoremstyle{definition}
\newtheorem{appdefinition}[appitem]{Definition}
\newtheorem{appassumption}[appitem]{Assumption}
\theoremstyle{plain}
\DeclareMathOperator*{\argmax}{arg\,max}

\newcommand{\update}[1]{\textcolor{black}{#1}}
\newcommand{\rev}[1]{{\color{black}#1}}  %
\newcommand{\DP}{\mathcal{D}_{\mathrm{P}}}
\newcommand{\DS}{\mathcal{D}_{\mathrm{S}}}
\newcommand{\DU}{\mathcal{D}_{\mathrm{U}}}
\newcommand{\dP}{d_{\mathrm{P}}}
\newcommand{\dS}{d_{\mathrm{S}}}
\newcommand{\dU}{d_{\mathrm{U}}}
\newcommand{\piP}{\pi_{\mathrm{P}}}
\newcommand{\piS}{\pi_{\mathrm{S}}}

\newcommand{\rP}{\mathrm{P}}
\newcommand{\rS}{\mathrm{S}}
\newcommand{\rU}{\mathrm{U}}
\newcommand{\cmark}{\checkmark}
\newcommand{\xmark}{-}

\newcommand{\bS}{\mathbf{S}}
\newcommand{\cA}{\mathcal{A}}
\newcommand{\bU}{\mathbf{N}}
\newcommand{\supp}{\operatorname{supp}}
\newcommand{\FNR}{\operatorname{FNR}}
\newcommand{\FPR}{\operatorname{FPR}}
\newcommand{\rstar}{r^*}
\newcommand{\rhat}{\hat{r}}
\newcommand{\hbS}{\hat{\mathbf{S}}}
\newcommand{\Cw}{C_w}

\iflatexml
\title{DexDICE: Reconciling Safety and Performance via Dual-Expert Offline Imitation Learning}
\else
\title{DexDICE: Reconciling Safety and Performance via\\ Dual-Expert Offline Imitation Learning}
\fi

\iflatexml
\usepackage{authblk}
\author[1,2]{Seokin Seo}
\author[2]{Sung Kuk Shyn}
\author[1,3]{Minji Seo}
\author[1]{Yisoo Lee$^{\dagger}$}
\author[4]{Jongmin Lee$^{\dagger}$}
\author[5]{Hongseok Yang$^{\dagger}$}
\author[2]{Kee-Eung Kim$^{\dagger}$\thanks{\textsuperscript{\textdagger}Equal advising.}}
\affil[1]{Korea Institute of Science and Technology (KIST)}
\affil[2]{Korea Advanced Institute of Science and Technology (KAIST)}
\affil[3]{Hyundai Motor Company}
\affil[4]{Yonsei University}
\affil[5]{Korea Institute for Advanced Study (KIAS)}
\else
\author{%
   Seokin Seo$^{1,2}$, Sung Kuk Shyn$^{2}$, Minji Seo$^{1,3}$, \\
    \textbf{Yisoo Lee}$^{1\dagger}$, \textbf{Jongmin Lee}$^{4\dagger}$, \textbf{Hongseok Yang}$^{5\dagger}$, \textbf{Kee-Eung Kim}$^{2}$\thanks{Equal advising.} \vspace{5pt} \\
  $^{1}$Korea Institute of Science and Technology (KIST), \\
  $^{2}$Korea Advanced Institute of Science and Technology (KAIST), \\
  $^{3}$Hyundai Motor Company,
  $^{4}$Yonsei University, \\
  $^{5}$Korea Institute for Advanced Study (KIAS) \vspace{5pt}\\
}
\fi

\begin{document}

\setcounter{footnote}{1}%
\maketitle
\setcounter{footnote}{0}%

\begin{abstract}
Ensuring high returns while providing strong safety guarantees in offline imitation learning (IL) is a fundamental challenge: existing approaches either rely on explicitly specified cost functions, which are difficult to design for complex real-world safety constraints, or require expert demonstrations that are both safe and high-performing, which are often too costly to collect.
We introduce \emph{dual-expert imitation learning}, which leverages two complementary data sources: (i) safety-compliant demonstrations that faithfully follow safety guidelines but may be suboptimal, and (ii) performance-oriented demonstrations that achieve high returns but may violate safety guidelines.
To address this problem, we propose DexDICE (\textbf{D}ual-\textbf{ex}pert offline imitation learning via stationary \textbf{DI}stribution \textbf{C}orrection \textbf{E}stimation).
Instead of conventional distribution constraints, DexDICE combines DICE-style offline learning with safe support constraints, enabling agents to exploit high-return behaviors while remaining within safety-compliant regions.
We prove that the ideal solution satisfying this constraint exactly has zero unsafe-region visitation with high probability.
We evaluate DexDICE on a re-curated DSRL benchmark and validate it on a real-world mobile-robot navigation task trained directly from human teleoperation demonstrations, where it substantially outperforms baselines and remains robust to data scarcity and quality degradation.\footnote{Code: \url{https://github.com/tzs930/dexdice}}
\end{abstract}

\iflatexml\else\etocdepthtag.toc{mtchapter}\fi%

\input{contents/1_introduction}
\input{contents/2_related_work}
\input{contents/3_problem_definition}
\input{contents/4_method}
\input{contents/5_experiment}
\input{contents/6_conclusion}

\begin{ack}
We thank the reviewers and the area chairs for their valuable feedback.
We also acknowledge support from the Google.org GCP Program in the form of Google Cloud credits.
S.~Seo, M.~Seo and Y.~Lee were supported by the National Research Foundation of Korea (NRF) grant funded by the Korea government (MSIT) (No.~RS-2025-25448259).
J.~Lee was supported by
the Institute of Information \& communications Technology Planning \& Evaluation (IITP) grants funded by the Korea government (MSIT) (No.~RS-2020-II201361, RS-2024-00457882),
the National Research Foundation of Korea (NRF) grants funded by the Korea government (MSIT) (No.~RS-2026-25505230, RS-2026-25619452),
and a grant from the Institute for AI and Social Innovation at Yonsei University.
H.~Yang was supported by the National Research Foundation of Korea (NRF)
grant funded by the Korean Government (MSIT) (No.~RS-2023-00279680),
and a KIAS Individual Grant (CG109601) at the Korea Institute for Advanced Study.
S.~Shyn and K.~Kim were supported by the Institute of Information \& communications Technology Planning \& Evaluation (IITP) grants funded by the Korea government (MSIT) (No. RS-2024-00457882, RS-2022-II220311, RS-2020-II200940, RS-2019-II190075).
The authors declare no competing financial interests.

\end{ack}

\bibliographystyle{plainnat}
\bibliography{references}

\newpage
\appendix
\input{contents/appendix}

\end{document}

%% file: contents/1_introduction.tex
\section{Introduction}
Imitation learning (IL) is a powerful paradigm for sequential decision-making that learns directly from expert demonstrations~\citep{hussein2017imitation,ho2016generative,zare2024survey}, bypassing the reward specification required in reinforcement learning (RL).
IL has enabled sophisticated behaviors in real-world domains such as robotic manipulation~\citep{fang2019survey,xie2020deep,johns2021coarse} and autonomous driving~\citep{le2022survey,hawke2020urban}.
In many such domains, however, collecting additional on-policy interaction data can be expensive or impractical due to hardware wear-and-tear, limited access to the system, and slow or costly resets, motivating offline IL methods that learn entirely from fixed datasets~\citep{jarrett2020strictly,seo2024mitigating}.
Since expert demonstrations are often scarce, recent offline IL methods also leverage auxiliary unlabeled trajectories that, while potentially suboptimal, provide valuable coverage of transition information~\citep{kim2022demodice,xu2022discriminator,yu2023offline, ma2022versatile}.

Separately, when IL policies are deployed in safety-critical applications, ensuring safety remains an open challenge. A common strategy is to encode safety requirements via explicit cost functions and solve a constrained optimization problem~\citep{yin2021imitation,ciftci2025safe,shao2024imitating}. However, specifying such costs can be as difficult as reward design: safety desiderata are often nuanced, context-dependent, and hard to instrument. Inverse constraint learning (ICL)~\citep{liu2024comprehensive,hugessen2024simplifying,malik2021inverse} offers an alternative by inferring safety constraints from demonstrations, but many ICL methods still rely on access to reward signals that IL does not assume. These limitations motivate methods that can jointly optimize safety and performance solely from demonstration data, without requiring an explicitly specified cost function.

In practice, demonstrations that are simultaneously safe and high-performing are difficult to obtain from a single expert.
Moreover, simply pooling demonstrations from multiple experts (e.g., some prioritizing safety and others prioritizing performance) does not resolve the issue,
since it introduces the challenging problem of disentangling safety and performance signals within the resulting dataset.
A more realistic strategy is therefore to separately collect different types of expert demonstrations, one focused on safety and the other focused on performance~\citep{kwon2024safe}.
This separation provides clearer supervision for each objective, making it easier for offline IL algorithms to leverage both effectively.
Such dual-source datasets are common in practice: for example, autonomous driving data contain high-speed but risky maneuvers, alongside safe and conservative driving behaviors~\citep{bansal2018chauffeurnet,campolettano2025potential};
similarly, in robotics and healthcare, some demonstrations are efficiency-oriented while others explicitly prioritize safety~\citep{mandlekar2021matters,lasota2014toward,nemati2016optimal,komorowski2018artificial}.

Motivated by these observations, we introduce a novel problem setting, \textbf{dual-expert imitation learning}, in which the learner has access to two types of expert demonstrations: \emph{safe demonstrations}, which adhere to safety constraints but may be suboptimal in task performance, and \emph{performant demonstrations}, which achieve high returns but may violate safety constraints. We further consider \textbf{dual-expert offline IL}, a more restrictive setting that forbids online interaction during training; in lieu of online rollouts, we incorporate \emph{unlabeled demonstrations} that provide broader transition coverage, following prior work on offline IL~\citep{kim2022demodice,jang2023safedice,xu2022discriminator}.

To address this setting, we propose \emph{DexDICE} (\textbf{D}ual-\textbf{ex}pert offline imitation learning via stationary \textbf{DI}stribution \textbf{C}orrection \textbf{E}stimation), which extends the DICE framework~\citep{kim2022demodice} with a \textit{support constraint} derived from the safe demonstrations. The resulting policy is the closest approximation to the performant expert's state-action distribution within the estimated safe support,
with a high-probability guarantee on zero unsafe-region visitation for the ideal support-constrained policy (Theorem~\ref{thm:safety-guarantee}).

\begin{figure*}
    \centering
    \includegraphics[width=\textwidth]{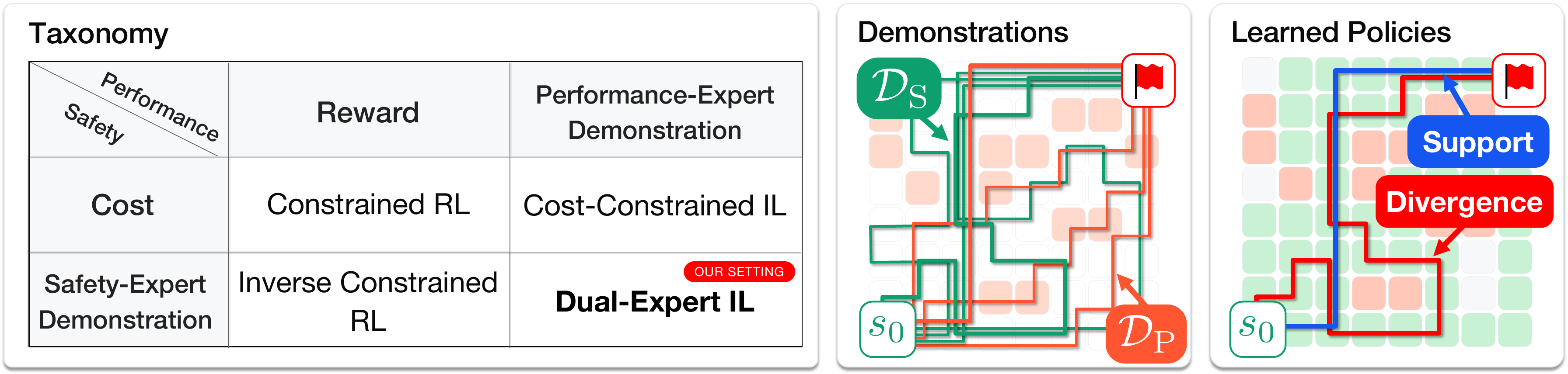}
    \caption{
    {
    {(left)} Taxonomy of safe sequential decision-making problems by available performance and safety signals. {(middle)} Dual-expert IL instance: $\mathcal{D}_\rP$ prioritizes fast goal reaching but is possibly unsafe, while $\mathcal{D}_\rS$ stays safe but possibly suboptimal. {(right)} Prior offline IL relies on \textit{divergence regularization} (e.g., behavior regularization), biasing the policy toward $\mathcal{D}_\rS$'s suboptimal behaviors (\textcolor{red}{red} path); DexDICE instead imposes a \textit{support constraint} from $\mathcal{D}_\rS$ (\textcolor{blue}{blue} path), recovering $\mathcal{D}_\rP$'s near-optimal route within the safe support.
    }
    }
    \label{fig:taxonomy_comparison}
\end{figure*}

Our main contributions are summarized below:
\vspace{-5pt}
\begin{itemize}[leftmargin=3mm]
    \item We formalize \emph{dual-expert offline imitation learning}, a problem class that disentangles safety and performance objectives through complementary demonstrations alone, without the explicit reward or cost labels required by prior methods.
    \item We propose \emph{DexDICE}, the first DICE-based offline IL algorithm to replace distributional regularization toward $d_\rS$ with an explicit \emph{safe-support constraint}, decoupling the learned policy's performance from the safe expert's suboptimality.
    \item We empirically validate DexDICE on a re-curated DSRL benchmark~\citep{liu2024datasets}, where it achieves superior (i) safety-performance trade-off, (ii) data efficiency, and (iii) robustness to variations in data quality, and further demonstrate (iv) real-robot applicability with human teleoperation demonstrations.
\end{itemize}

%% file: contents/2_related_work.tex
\section{Related Work}
\label{sec:related_work}

\noindent{\bf Offline Safe RL}\ \,
Offline safe RL (OSRL)~\citep{leecoptidice,liu2023constrained,zheng2024safe,liu2025c2iql,yang2024offline} requires per-transition reward $r(s,a)$ and cost $c(s,a)$, which the dual-expert IL setting does not provide. To naively apply these methods, we would need to estimate reward and costs from the given demonstrations.
However, such reward/cost estimates are unreliable especially in our setting where $\DU$ is a latent mixture of safe and unsafe transitions, turning reward/cost recovery into a hard PU-learning problem~\citep{bekker2020learning} (Appendix~\ref{app:offline_safe_rl_discussion}).
We evaluate recent OSRL methods CDT~\citep{liu2023constrained} and TraC~\citep{yang2024offline} with pseudo-labels $\hat r, \hat c$ estimated from the $(\DP, \DS, \DU)$ partition as matched-information baselines.
For an empirical comparison with OSRL methods under their original (ground-truth) information access, see Appendix~\ref{app:comparison_with_baselines}.

\noindent{\bf Safe Imitation Learning}\ \,
Existing safe IL methods assume either explicit cost or constraint signals~\citep{shao2024imitating,cao2025simple,ciftci2025safe} or model-based safety constructs such as Hamilton-Jacobi reachability~\citep{bansal2017hamilton} and control barrier functions~\citep{ames2019control}, both of which require system dynamics or hand-specified constraints unavailable in our offline setting. Beyond cost-aware methods, some safe IL approaches sidestep cost signals by leveraging undesirable demonstrations~\citep{jang2023safedice,burnwal2026safemil,hoang2025noexperts}, while an online dual-expert method such as Safe CoR~\citep{kwon2024safe} shares our two-set demonstration partition but additionally requires environment interaction with explicit reward and cost signals.

\noindent{\bf Offline Imitation Learning}\ \,
Offline IL is essential when environment interaction is restricted and reward specification is difficult,
and mixed-quality datasets are commonly leveraged to supplement scarce transition samples~\citep{xu2022discriminator,kim2022demodice,yu2023offline,liu2025pngail,hoang2024sprinql}.
DWBC~\citep{xu2022discriminator} reweights behavioral cloning via a PU-learning discriminator, while DICE-based methods such as DemoDICE~\citep{kim2022demodice} and RelaxDICE~\citep{yu2023offline} cast IL as convex distribution matching over stationary state-action distributions.
The closest cost-aware comparator is the offline safe IL variant of OSRL method FISOR~\citep{zheng2024safe} (henceforth \emph{FISOR-IL}), which uses per-transition cost labels; we evaluate it under matched-information access by supplying pseudo-cost $\hat c$ derived from the $(\DP, \DS, \DU)$ partition.

\noindent{\bf Inverse Constrained Learning (ICL)}\ \,
ICL methods~\citep{liu2024comprehensive,malik2021inverse,kim2023learning,subramanian2024confidence,hugessen2024simplifying} and their offline extension ICSDICE~\citep{quan2024learning} infer a latent cost function from demonstrations and then optimize against it. This indirection is inherently ill-posed due to non-uniqueness of admissible cost functions; we instead learn safe policies directly from dual-expert demonstrations, bypassing cost recovery entirely.

Figure~\ref{fig:taxonomy_comparison} positions our contribution. Prior DICE-based offline IL enforces safety via \emph{divergence regularization}, which can bias the policy toward suboptimal safe behaviors (red path). DexDICE instead enforces safety via an explicit \emph{support constraint} (blue path), and to our knowledge is the first such formulation among DICE-based offline IL methods. See Appendix~\ref{app:detailed_related_work} for extended discussion.

%% file: contents/3_problem_definition.tex
\section{Problem Definition}
\vspace{-5pt}
\subsection{Markov Decision Processes}
\vspace{-5pt}
We consider a Markov decision process (MDP) defined as a tuple $\mathcal{M} := \langle \mathcal{S}, \mathcal{A}, T, \rho_0, r, \gamma \rangle$, where $\mathcal{S}$ and $\mathcal{A}$ denote the state and action spaces respectively, $T: \mathcal{S} \times \mathcal{A} \to \Delta(\mathcal{S})$ is the transition probability function, $\rho_0 \in \Delta(\mathcal{S})$ is the initial state distribution, $r: \mathcal{S} \times \mathcal{A} \to \mathbb{R}$ is the reward function, and $\gamma \in (0,1)$ is the discount factor. We define the policy space as $\Pi := \{\pi : \mathcal{S} \to \Delta(\mathcal{A})\}$.
Given policy $\pi \in \Pi$, the stationary distribution $d_\pi$ is defined as:
\begin{equation*}
    d_\pi(s,a) := (1-\gamma) \sum_{t=0}^\infty \gamma^t \mathbb{P}(s_t=s, a_t=a),
\end{equation*}
where $s_0 \sim \rho_0$, $a_t \sim \pi(s_t)$, and $s_{t+1} \sim T(s_t,a_t)$.
Then, $d_\pi$ satisfies the Bellman flow constraint:
\begin{equation*}
    \textstyle \sum\limits_{a} d_\pi(s,a) = (1-\gamma) \rho_0(s) + \gamma \sum\limits_{\bar{s}, \bar{a}} T(s|\bar{s},\bar{a}) d_\pi(\bar{s},\bar{a}), \forall s \in \mathcal{S}.
\end{equation*}

A constrained Markov decision process (CMDP)~\citep{altman1999constrained} extends the standard MDP framework to incorporate safety constraints. We define a CMDP as $\mathcal{M}_c := \langle \mathcal{M}, c, \rev{\delta_c} \rangle$, where $\mathcal{M}$ is the unconstrained MDP, $c: \mathcal{S} \times \mathcal{A} \to \mathbb{R}_{\geq 0}$ is a non-negative cost function, and $\rev{\delta_c} \ge 0$ is a cost limit.
Since we only observe demonstrations, neither the reward function $r$ nor the cost function $c$
is available in our setting.

\vspace{-5pt}
\subsection{Imitation Learning}
\label{sec:imitation}
Imitation learning (IL) aims to recover an expert policy from demonstrations in the absence of reward signals. We denote the demonstration dataset as
$\mathcal{D} = \{(s_i,a_i)\}_{i=1}^N$, a collection of state-action pairs sampled from expert trajectories.
A straightforward approach to IL is behavioral cloning (BC), which treats the problem as supervised learning and solves it by minimizing the negative log-likelihood:
\begin{equation*}
    \min_\pi \quad \mathbb{E}_{(s,a) \sim \mathcal{D}}\left[ -\log \pi(a|s) \right].
\end{equation*}
However, BC suffers from compounding errors due to covariate shift, particularly over long horizons \citep{pmlr-v9-ross10a}.
To address this limitation, recent IL algorithms typically minimize a distribution matching objective:
\begin{equation}
    \min_\pi \quad \mathbb{D}(d_\pi(s,a), d_{\mathcal{D}}(s,a)), \notag%
\end{equation}
where $\mathbb{D}$ is a divergence measure (e.g., $f$-divergences \citep{kim2022demodice,zhang2020f,ho2016generative} and integral probability metrics \citep{kim2018imitation, dadashiprimal}) and $d_{\mathcal{D}}$ denotes the state-action distribution in $\mathcal{D}$.

In many practical applications, on-policy interaction with the environment can be
prohibitively expensive, giving rise to \emph{offline IL}.
Since expert data is often scarce, offline IL methods leverage auxiliary unlabeled demonstrations $\DU$, which may be suboptimal but provide valuable transition information~\citep{kim2022demodice,xu2022discriminator, ma2022versatile}.
While standard IL assumes demonstrations from a single expert, real-world scenarios often involve separate sources for safety and performance, motivating the dual-expert formulation introduced in the next section.

\subsection{Dual-Expert Imitation Learning}

We consider \emph{dual-expert IL} from safety-performance expert demonstrations,
consisting of a set of safety-expert (\emph{safe}) demonstrations $\DS$, which satisfy the underlying constraints but may be suboptimal, and a set of performance-expert (\emph{performant}) demonstrations $\DP$, which achieve high returns but may violate unknown safety constraints. %

Denote the underlying state-action distributions of $\DS,\DP$ as $\dS,\dP$.
Since $\DS$ may contain suboptimal demonstrations, we aim to recover only its safety-compliant support of state-action pairs rather than matching the distribution $\dS$.
Accordingly, we seek a policy $\pi$ that aligns with performant behaviors while remaining within the safe support.
Formally, the goal of dual-expert IL is to recover a policy $\pi$ whose stationary distribution $d_\pi$ matches $\dP$ with a support constraint derived from $\dS$:
\begin{equation}
\min_\pi \quad \mathbb{D}(d_\pi \| \dP)
\quad \text{s.t.} \quad \mathrm{supp}(d_\pi) \subseteq \mathrm{supp}(\dS),
\label{eq:dual_il_objective_simplified}
\end{equation}
where $\mathrm{supp}(d) = \{(s,a) \mid d(s,a) > 0\}$.
We refer to this constraint as the \emph{safe support constraint}.

To ensure the optimization problem is well-posed, we assume that there exists a policy $\pi$ whose stationary distribution $d_\pi$ satisfies
$\mathrm{supp}(d_\pi)\subseteq \mathrm{supp}(\dS)\cap \mathrm{supp}(\dP)$.
This requirement is weaker than assuming access to demonstrations that are simultaneously safe and performant, which corresponds to the special case $\dS=\dP$ \rev{(practical examples in Appendix~\ref{app:overlap_interpretation})}.
When the overlap $\mathrm{supp}(\dS)\cap \mathrm{supp}(\dP)$ is small, the optimum may be far from $\dP$ in divergence, but the formulation still returns the closest safe approximation to $\dP$.

Equivalently, this is a hard cost-constrained IL problem~\citep{shao2024imitating} with $c(s,a)=0$ on $\mathrm{supp}(\dS)$ and $c(s,a)>0$ otherwise; the support constraint enforces $\mathbb{E}_{d_\pi}[c(s,a)]=0$ implicitly (see Appendix~\ref{sec:rigorous_definition}).

%% file: contents/4_method.tex
\section{DexDICE: Dual-Expert Offline IL with Safe Support Constraint}
\label{sec:method}
We study the offline version of the dual-expert IL problem, where on-policy interaction is prohibited during training.
In addition to $\DS$ and $\DP$, we assume access to a more abundant unlabeled dataset $\DU$ sampled from distribution $\dU$, which provides coverage of transition information. Following the standard positive-unlabeled convention, $\DU$ is constructed to include $\DS$ and $\DP$ as subsets in our experiments. \rev{Accordingly, we assume $\dS\ll\dU$ and $\dP\ll\dU$ at the population level, so that the density ratios $\dS/\dU$ and $\dP/\dU$ used below are well-defined $\dU$-almost everywhere. Here $\dU$ is neither a safety nor a performance target but a common offline reference: it supplies transition coverage for learning the Bellman flow constraints offline without safety or performance annotation~\citep{kim2022demodice}, places both expert sources on a common scale through the ratios $\dS/\dU$ and $\dP/\dU$ (Section~\ref{sec:safe_support_estimation}), and defines the DICE correction $w=d/\dU$, which requires $d^\pi\ll\dU$ (Assumption~\ref{ass:bounded}).}

\subsection{Support-Constrained DICE Objective}
Since direct sampling from $d_\pi$ is not allowed in the offline setting,
we adopt the DICE (Distribution Correction Estimation) framework~\citep{nachum2019dualdice,lee2021optidice},
which formulates policy optimization as convex optimization over stationary distributions.
The key idea is to introduce a correction ratio $w(s,a) := d(s,a)/\dU(s,a)$,
enabling off-policy optimization using only static datasets~\citep{kim2022demodice,leecoptidice}.
Following this, we optimize over $d$ instead of $\pi$.

Unlike conventional DICE methods~\citep{kim2022demodice} that use a divergence regularization (e.g., $D_\mathrm{KL}(d \| \dS)$) and may bias the policy toward suboptimal safe behaviors, our support constraint restricts only \emph{where} the policy can act, not \emph{how often}, decoupling safety from suboptimality (validated in a tabular experiment in Appendix~\ref{sec:app.toy_experiment}).

To enforce the safe support constraint $\mathrm{supp}(d) \subseteq \mathrm{supp}(\dS)$ in the objective~(\ref{eq:dual_il_objective_simplified}), we reformulate the constraint as an equality condition:
\begin{equation}
    \Big(\mathrm{supp}(d) \subseteq \mathrm{supp}(\dS)\Big)
    \Longleftrightarrow
    \sum_{s,a}\mathds{1}_{\dS}(s,a) d(s,a)= 1 ,
    \notag
\end{equation}
where $\mathds{1}_{\dS}(s,a) = 1$ if $(s,a) \in \mathrm{supp}(\dS)$ and $0$ otherwise; the equivalence uses $\sum_{s,a} d(s,a) = 1$ since $d$ is a probability distribution.
Then, our objective is to find a stationary distribution $d$ that minimizes divergence to $\dP$ while satisfying the safe support constraint expressed as above.
To ensure that $d$ corresponds to a valid policy, we impose Bellman flow constraints and arrive at the following optimization problem:
\begin{align}
\small
    \min_d  \ \, D_\mathrm{KL}(d \| \dP) \qquad \text{s.t.} & \qquad \sum_{s,a} \mathds{1}_{\dS}(s,a) d(s,a) = 1 \notag \\
    & \qquad (\mathcal{B}_*d) (s) = (1\,{-}\,\gamma) \rho_0 (s) \,{+}\, \gamma (\mathcal{T}_*d)(s),\qquad \forall s \,{\in}\, \mathcal{S} \notag \\
    & \qquad d(s,a) \ge 0, \qquad \forall s \,{\in}\, \mathcal{S}, a \,{\in}\, \mathcal{A},
    \label{eq:method_consobj}
\end{align}
where $(\mathcal{B}_* d) (s) := \sum_{a} d(s,a)$ is a marginalization operator and $(\mathcal{T}_*d)(s) := \sum_{\bar{s},\bar{a}} T(s|\bar{s},\bar{a}) d(\bar{s},\bar{a})$ is a transposed Bellman operator.

To unconstrain the problem, we introduce Lagrange multipliers $\lambda$ for the safe-support constraint and $\nu(s)$ for the Bellman flow constraints, flip the sign, and apply the change of variables $w(s,a) := d(s,a)/\dU(s,a)$ together with the density ratio $r_\rP(s,a) := \dP(s,a)/\dU(s,a)$. This yields the dual objective (full derivation in Appendix~\ref{app:lagrangian_derivation}):
\begin{align}
L(w,\nu,\lambda) := (1-\gamma)\mathbb{E}_{s\sim\rho_0}[\nu(s)] - \lambda + \mathbb{E}_{(s,a)\sim \dU}\big[ w(s,a)\cdot(A_{\nu,\lambda}(s,a) - \log w(s,a)) \big],
\label{eq:loss_def}
\end{align}
with advantage $A_{\nu,\lambda}(s,a) := \gamma (\mathcal{T}\nu)(s,a) - \nu(s) + \log r_\rP(s,a) + \lambda \mathds{1}_{\dS}(s,a)$, where $(\mathcal{T}\nu)(s,a) := \mathbb{E}_{s^\prime \sim T(s,a)}[\nu(s^\prime)]$. Intuitively, $A_{\nu,\lambda}$ captures the temporal-difference structure and encourages matching $\dP$ via the performance proxy $\log r_\rP$, while $\lambda\mathds{1}_{\dS}$ relatively up-weights $w^*$ inside $\mathrm{supp}(\dS)$ by a factor $e^\lambda$ through the closed form $w^*_{\nu,\lambda}=\exp(A_{\nu,\lambda}-1)$ (Eq.~\ref{eq:optidice}).
In practice, $A_{\nu,\lambda}$ is estimated from $(s,a,s')$ tuples in $\DU$ via $(\mathcal{T}\nu)(s,a) \approx \nu(s')$, while $r_\rP$ and $\mathds{1}_{\dS}$ are obtained via discriminator training, which we revisit in Section~\ref{sec:safe_support_estimation}.

Then, our optimization objective becomes:
\begin{align}
    \min_{\nu, \lambda}\max_{w \ge 0}  L(w,\nu,\lambda). \label{eq:minimax_objective}
\end{align}

\begin{proposition}
(Strong Duality) \rev{Under the constraint qualification stated in Appendix~\ref{sec:app.proof},} $L(w,\nu,\lambda)$ satisfies strong duality, which ensures:
\begin{equation}
\max_{w \ge 0} \min_{\nu, \lambda } L(w,\nu,\lambda) =  \min_{\nu, \lambda } \max_{w \ge 0} L(w,\nu,\lambda).    \label{eq:strong_dual}
\end{equation}
\end{proposition}

\begin{proposition} %
(Closed-form solution; \citet{leecoptidice}) The inner maximization on the RHS of Eq.~(\ref{eq:strong_dual}) admits a closed-form solution:
\begin{equation}
w^*_{\nu,\lambda} (s,a) =  \exp( A_{\nu,\lambda} (s,a) - 1). \label{eq:optidice}
\end{equation}
\end{proposition}
See Appendix~\ref{sec:app.proof} for the proofs of these propositions.
Hence, the problem is converted into the following minimization problem w.r.t $\nu,\lambda$:
\begin{align}
    &\min_{\nu, \lambda} \, \mathbb{E}_{(s,a)\sim \dU}
    [ w^*(s,a) \cdot (- \log w^*(s,a) + {A}_{\nu, \lambda}(s,a))]  + (1-\gamma) \mathbb{E}_{s \sim \rho_0}[\nu (s) ] - \lambda \notag\\
=& \min_{\nu, \lambda} \, \mathbb{E}_{(s,a) \sim \dU}[ \exp({A}_{\nu,\lambda}(s,a) - 1)]  + (1-\gamma) \mathbb{E}_{s \sim \rho_0}[\nu (s) ] - \lambda
=: L(\nu, \lambda).
\label{eq:dexdice_obj}
\end{align}
We refer to the method that optimizes the objective in Eq.~(\ref{eq:dexdice_obj}) as DexDICE,
short for \textbf{D}ual-\textbf{ex}pert offline IL via stationary \textbf{DI}stribution \textbf{C}orrection \textbf{E}stimation.

\subsection{Support Estimation via Density Ratios}
\label{sec:safe_support_estimation}

To optimize the objective in Eq.~(\ref{eq:dexdice_obj}), we need an estimator $\hat{\mathds{1}}_{\dS}$ of the support indicator $\mathds{1}_{\dS}$. Directly estimating the density $\dS$ is intractable in high-dimensional state-action spaces; we instead estimate a bounded density \emph{ratio} via classification, which scales more favorably with dimensionality~\citep{hido2011statistical}.

The Bayes-optimal classifier of safe vs.\ unsafe state-action pairs is a likelihood-ratio test on $\dS/d_\mathrm{N}$, where $d_\mathrm{N}$ is the unsafe distribution. However, $d_\mathrm{N}$ is not directly observable in our setting. We therefore estimate the \emph{safe-unlabeled} ratio $\dS/\dU$ from $\DS$ and $\DU$, and relate it to $\dS/d_\mathrm{N}$ via the PU-learning mixture~\citep{scott2009novelty}: writing $\dU = \eta\,\dS + (1-\eta)\,d_\mathrm{N}$ with class prior $\eta$, rearrangement yields
\begin{align}
    \frac{d_\mathrm{N}(s,a)}{\dS(s,a)} = \frac{1}{(1-\eta)}  \frac{\dU(s,a)}{\dS(s,a)}   - \frac{\eta}{1-\eta}, \label{eq:ratio_relationship} %
\end{align}
showing that the unsafe-safe ratio is a monotone transformation of the unlabeled-safe ratio, rendering the two ratios equivalent as decision functions up to a threshold shift.

Following~\citet{kim2022demodice}, we train two discriminators: $\hat{C}_\rS$ estimates $\hat r_\rS \approx \dS/\dU$ (used to construct $\hat{\mathds{1}}_{\dS}$), and $\hat{C}_\rP$ estimates $\hat r_\rP \approx \dP/\dU$ (the performance proxy in $A_{\nu,\lambda}$). Training details are in Appendix~\ref{discriminator_estimation}. Compared to estimators that use only safe demonstrations, this safe-vs-unlabeled discriminator empirically yields more robust support estimates (Appendix~\ref{sec:supp_estimation}).

We do not directly invert Eq.~(\ref{eq:ratio_relationship}) since the class prior $\eta$ is notoriously hard to identify~\citep{elkan2008learning,pmlr-v45-Christoffel15}. Instead, we exploit \emph{rank consistency}: the monotone relationship in Eq.~(\ref{eq:ratio_relationship}) implies that $\hat{r}_\rS$ alone induces the same ordering over state-action pairs as the unsafe-safe ratio, so thresholding $\hat{r}_\rS$ suffices.
To this end, we construct a non-trivial support estimator $\hat{\mathds{1}}_{\dS}$ of $\mathds{1}_{\dS}$ by specifying a target false-negative rate (FNR) $\kappa \in (0,1)$, i.e., the fraction of safe pairs incorrectly rejected.
Specifically, we set the threshold $\tau_{\kappa}$ as the $\kappa$-quantile of $\hat{r}_\rS$ over $\mathcal D_\rS$:
\begin{align}
\tau_{\kappa}
&:= \mathrm{Quantile}_{\kappa}\!\big(\{\hat{r}_\rS(s,a):(s,a)\in\mathcal D_\rS\}\big) = \inf\Big\{\tau:\frac{1}{|\mathcal D_\rS|}\sum_{(s,a)\in\mathcal D_\rS}\mathds{1}\!\big(\hat r_\rS(s,a)\le \tau\big)\ge \kappa\Big\}. \label{eq:supp_estimator} %
\end{align}
We then define the support estimator by thresholding $\hat{r}_\rS$:
$\,\, \hat{\mathds{1}}_{\dS}(s,a)
:=\mathds{1}\!\big\{\hat r_\rS(s,a) \rev{\ge} \tau_{\kappa}\big\},$
\rev{so that a pair is accepted when $\hat r_\rS(s,a)\ge\tau_\kappa$ and rejected when $\hat r_\rS(s,a)<\tau_\kappa$.
By the definition of the empirical quantile, the empirical FNR on $\mathcal D_\rS$ (the fraction of $\mathcal D_\rS$ with $\hat r_\rS<\tau_\kappa$) is strictly below $\kappa$.}

\subsection{Theoretical Analysis}
\label{sec:theory}

Although the support estimator $\hat{\mathds{1}}_{\dS}$ in Section~\ref{sec:safe_support_estimation} is calibrated to a target FNR by construction, this does not by itself imply that \rev{a policy constrained to the estimated support} avoids the unsafe region at deployment. We now show that this calibration translates to zero unsafe-region visitation by the \rev{ideal support-constrained policy}.

\textbf{Setup.} Let $\bS=\supp(\dS)$ and $\bU=\supp(\dU)\setminus\bS$ denote the safe and (reachable) unsafe regions, and let $r^*(s,a):=\dS/(\dS{+}\dU)$ be the Bayes-optimal classifier of $\bS$ vs.\ $\bU$. The estimated safe support at threshold $\tau$ is $\hbS_\tau=\{(s,a):\hat{r}_\rS(s,a)\ge\tau\}$, with false-negative and false-positive rates $\FNR(\tau), \FPR(\tau)$ (edge cases on $\bU$ vs.\ $\bS^c\setminus\supp(\dU)$ are in Appendix~\ref{sec:app.theory}).\vspace{5pt}

\textbf{Assumption.} \rev{We assume the demonstrated safe support is safety-compliant, so that staying within $\supp(\dS)$ implies environmental safety (pointwise soundness; Appendix~\ref{sec:app.theory}). We consider an ideal support-constrained policy $\pi_\kappa$ with $\supp(d^{\pi_\kappa})\subseteq\hbS_{\tau_\kappa}$, where $\tau_\kappa$ is the empirical $\kappa$-quantile in Eq.~(\ref{eq:supp_estimator}) for a user-specified $\kappa\in(0,1)$, computed from calibration pairs in $\DS$ drawn i.i.d.\ from $\dS$ (Assumption~\ref{ass:calib}).
Let $\varepsilon>0$ bound the worst-case discriminator error over all state-action pairs, and let $\delta\in(0,1)$ be the total failure probability of the guarantee, split equally between the discriminator and the finite calibration sample. We assume $\|\hat r_\rS - r^*\|_\infty\le\varepsilon$ with probability at least $1-\delta/2$ over the training data (Assumption~\ref{ass:err}). For fixed $\delta$, $\varepsilon$ vanishes as the sample size grows.
}

The following theorem characterizes the unsafe-region visitation $d^{\pi_\kappa}(\bU)$ in terms of the discriminator error $\varepsilon$, the failure probability $\delta$, and user-specified $\kappa$.

\begin{theorem}[Safety guarantee of DexDICE; informal]
\label{thm:safety-guarantee}
Under \rev{i.i.d.\ calibration sampling and} standard regularity conditions on the discriminator estimation error, the DICE weights, \rev{and a safe-margin condition} (Assumptions~\rev{\ref{ass:calib}}--\ref{ass:margin} in Appendix~\ref{sec:app.theory}), if $\kappa > \kappa_0 + O(|\DS|^{-1/2})$, the support-constrained policy $\pi_\kappa$ visits the unsafe region with zero probability on a high-probability event:
$$d^{\pi_\kappa}(\bU) = 0 \quad \text{with probability } \geq 1 - \delta.$$
\end{theorem}
\vspace{10pt}

The \emph{boundary mass} $\kappa_0$ is the $\dS$-probability of safe pairs with low Bayes-optimal discriminator scores (formal definition in Appendix~\ref{sec:app.theory}). The condition $\kappa > \kappa_0 + O(|\DS|^{-1/2})$ requires $\kappa$ to exceed this mass by a finite-sample slack that vanishes as $|\DS|\to\infty$.

\noindent\textit{Key findings.}
\textbf{(i) \rev{Exact-zero safety with high probability}.} Once the discriminator error $\varepsilon$ falls below \rev{half} the safe margin, unsafe visitation drops to \emph{exact zero}, since $r^*\equiv 0$ on $\bU$ (Lemma~\ref{lem:fpr-zero}) converts sub-margin error into a hard zero.
\rev{This does not require perfect discriminator estimation: under the stated assumptions, $\tau_\kappa>\varepsilon$ excludes every pair in $\bU$ from the estimated support, yielding zero visitation for the ideal support-constrained policy.}

\textbf{(ii) Single-knob safety calibration.} By construction, $\kappa$ bounds the empirical FNR on $\DS$ strictly below $\kappa$. Under the conditions of Theorem~\ref{thm:safety-guarantee}, the same calibration yields $\FPR(\tau_\kappa)=0$ with high probability (Corollary~\ref{cor:simul-fnr-fpr}). Within the range covered by the theorem, varying $\kappa$ adjusts safe-pair retention while the FPR remains zero.

Detailed proofs and supporting lemmas are deferred to Appendix~\ref{sec:app.theory}.

\textbf{Practical algorithm.} In summary, $\hat r_{\rP}, \hat r_{\rS}$ are trained from $\DP, \DS, \DU$, and $\hat{\mathds{1}}_{\dS}$ is built from $\hat r_{\rS}$ via FNR calibration. Plugging these into $L(\nu,\lambda)$ (Eq.~\ref{eq:dexdice_obj}), we jointly optimize $\nu,\lambda$ to obtain $w^*_{\nu,\lambda}$ (Eq.~\ref{eq:optidice}) and recover the policy from $w^*_{\nu,\lambda}$ via weighted BC, following prior DICE frameworks~\citep{leecoptidice, kim2022demodice} (Appendix~\ref{app.policy_extraction}; full algorithm in Appendix~\ref{app.algorithm}).
\rev{This step introduces a theory--practice gap: Theorem~\ref{thm:safety-guarantee} concerns the ideal solution, whereas the extracted weighted-BC policy is a Gaussian with full support over the action space and hence cannot be hard-supported on $\hbS_{\tau_\kappa}$. A deployment-time projection of each proposed action onto $\hbS_{\tau_\kappa}$ substantially reduces the measured support mismatch %
but cannot eliminate it in general, since a reached state need not admit any in-support action and a finite-budget search may fail to find one even when it exists; the residual mismatch is therefore reported separately and is not covered by the exact guarantee (Appendix~\ref{sec:app.projection}).}

%% file: contents/5_experiment.tex
\section{Experiments}

Our experiments aim to answer the following questions: %
\begin{itemize}[leftmargin=3mm]
\item \textbf{Q1}: How effective are offline IL methods in the dual-expert demonstration setting? %
\item \textbf{Q2}: How data-efficient are these methods with respect to the sizes of $\DS$ and $\DP$? %
\item \textbf{Q3}: How robust are they to variations in demonstration quality (i.e., the proportion of safe-performant demonstrations) within $\DS$ and $\DP$? %
\item \textbf{Q4}: Does DexDICE remain effective on a real-world robot when trained from human-collected demonstrations?
\end{itemize}

Because dual-expert imitation learning is a problem setting we newly propose, no dedicated public benchmark currently exists. We therefore re-curate the offline safe RL benchmark DSRL~\citep{liu2024datasets} by relabeling each trajectory along two binary axes (performant vs.\ non-performant; safe vs.\ unsafe) based on predefined return and cost thresholds, and refer to this adapted dataset as the \emph{re-curated DSRL benchmark}. These labels are used solely to construct $\DP$ and $\DS$; DexDICE itself never consumes reward or cost signals during training.
For \textbf{Q1}, we use this re-curated DSRL benchmark, which spans a broad range of performance and safety levels. For \textbf{Q2} and \textbf{Q3}, which require precise control over dataset size and quality, we construct \emph{controlled datasets} on Safety Gymnasium~\citep{ji2023safety} by systematically mixing four distinct demonstration types. For \textbf{Q4}, we collect human teleoperation demonstrations on a physical mobile robot for real-world navigation evaluation.

Since dual-expert offline IL is a new problem setting with no directly comparable baselines, we adapt a range of established offline IL and offline safe RL methods with matched information access: %
\begin{itemize}[leftmargin=3mm]
    \item \textbf{BC}: Standard BC minimizing $\mathbb{E}_{(s,a)\sim \dP}[-\log \pi(a|s)]$. %
    \item \textbf{Filtered BC (FBC)}: BC restricted to samples retained by the estimated support $\hat{\mathds{1}}_{\dS}$. %
    \item \textbf{Jointly Filtered BC (JFBC)}: BC restricted to samples accepted by both $\hat{\mathds{1}}_{\dS}$ and $\hat{\mathds{1}}_{\dP}$. %
    \item \textbf{DemoDICE-S/-U}: Two variants of DemoDICE~\citep{kim2022demodice} that minimize $D_\mathrm{KL}(d\|\dP)+\alpha D_\mathrm{KL}(d\|\dS)$, parametrized via $w=d/\dS$ ($\DS$ substituted for $\DU$) and $w=d/\dU$ (using $\DU$), respectively; both serve as ablations of DexDICE that replace its support constraint with distribution matching to $\dS$. %
    \item \textbf{DWBC-S}: DWBC~\citep{xu2022discriminator} with $\DS$ substituted for $\DU$: expert-like samples are filtered from $\DS$ via a PU-learning-based discriminator $f$ that approximates $\DP$, and a policy is then trained by discriminator-weighted BC, minimizing $\mathbb{E}_{(s,a)\sim \dS}\left[-f(s,a) \log \pi(a\mid s) \right]$ (Appendix~\ref{app:discussion_related_work}). %
    \item \textbf{FISOR-IL}~\citep{zheng2024safe}, \textbf{CDT}~\citep{liu2023constrained}, and \textbf{TraC}~\citep{yang2024offline}: Offline safe IL/RL methods fed with binary pseudo-labels $(\hat r, \hat c)\in\{0,1\}^2$ in lieu of ground-truth reward and cost, ensuring matched information access (Appendix~\ref{app:offline_safe_rl_discussion}).
\end{itemize}
\textbf{Information access is matched.} BC, FBC, JFBC, DemoDICE-S/-U, and DWBC-S share DexDICE's $(\DS,\DP,\DU)$ partition; FISOR-IL, CDT, and TraC receive binary pseudo-labels $(\hat r,\hat c)\in\{0,1\}^2$ obtained by thresholding two discriminators trained on $\DP$ vs.\ $\DU$ and $\DS$ vs.\ $\DU$, instead of ground-truth reward and cost labels.

\input{contents/main_experiment_table}

For each task, the episode return is used as the performance metric and
normalized to $[0,1]$ using the minimum and maximum returns, and the episode cost is measured as the safety metric and normalized to $[0,1]$ by the cost upper bound \rev{$\delta_c$} (a method is deemed safe when its mean normalized cost is $\le 1$).
Each result averages 50 rollouts at the last checkpoint, across 5 random seeds.

\subsection{Effectiveness on Re-curated DSRL Benchmark}
\label{sec:dsrl_experiments}

\begin{figure*}[t]
    \centering
    \begin{subfigure}[b]{0.48\textwidth}
        \centering
        \includegraphics[width=\textwidth]{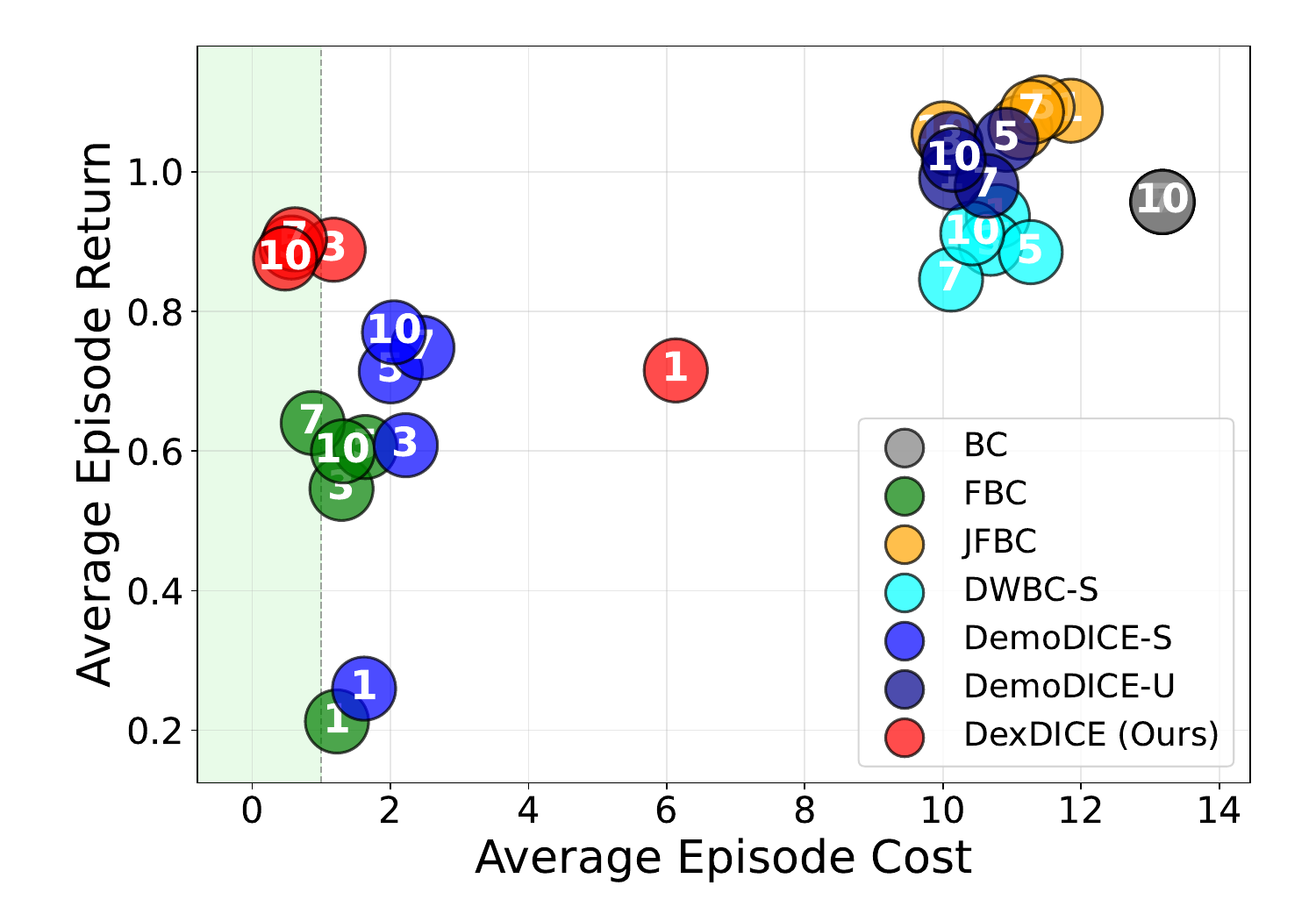}
        \caption{Data efficiency on $N_{\DS}$}
        \label{fig:data_efficiency_DS}
    \end{subfigure}
    \hspace{5pt}
    \begin{subfigure}[b]{0.48\textwidth}
        \centering
        \includegraphics[width=\textwidth]{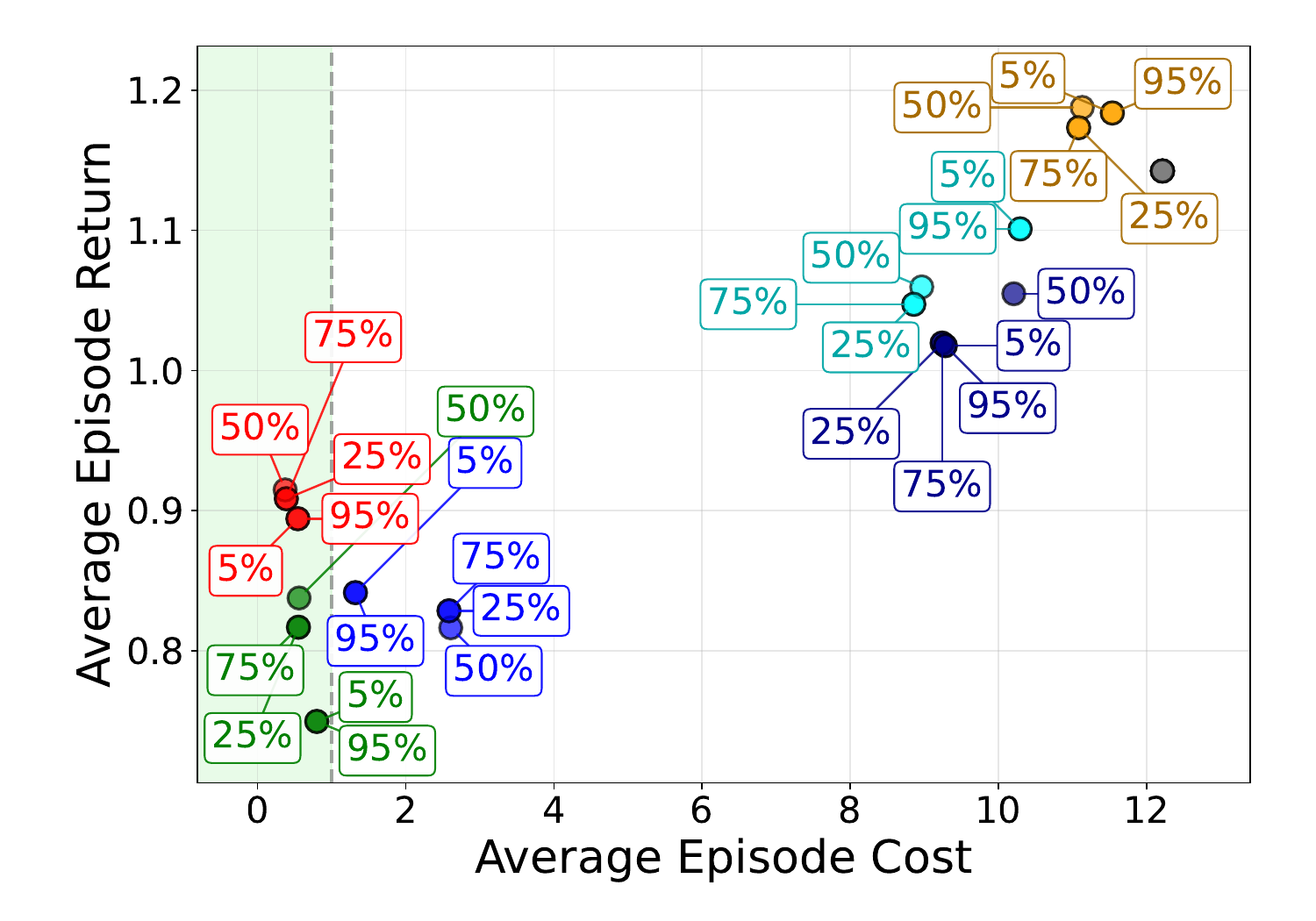}
        \caption{Dependency on $\textbf{quality}(\mathcal{D}_S)$}
        \label{fig:data_quality_dS}
    \end{subfigure}
    \begin{subfigure}[b]{0.48\textwidth}
        \centering
        \includegraphics[width=\textwidth]{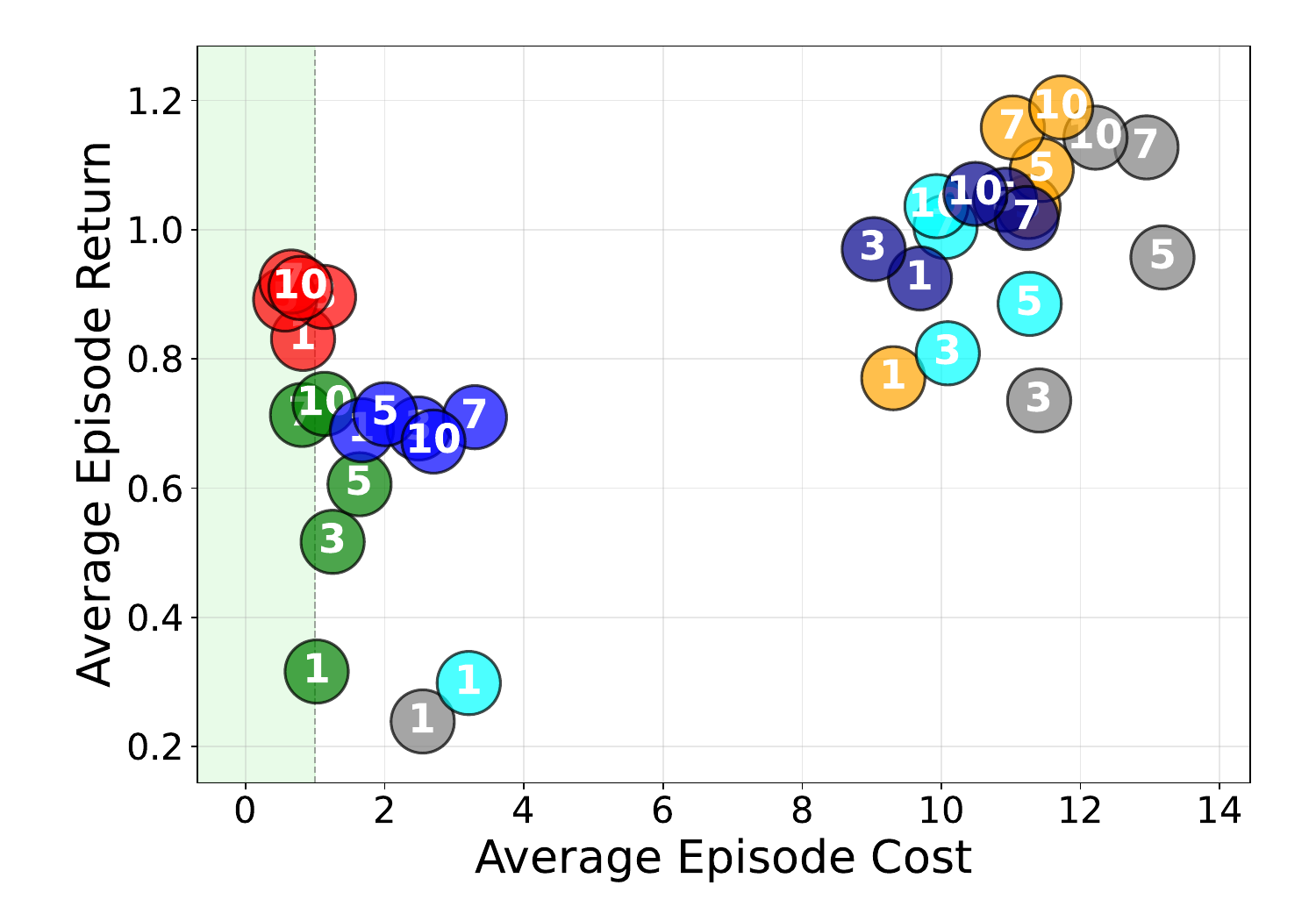}
        \caption{Data efficiency on $N_{\DP}$}
        \label{fig:data_efficiency_DP}
    \end{subfigure}
    \hspace{5pt}
    \begin{subfigure}[b]{0.48\textwidth}
        \centering
        \includegraphics[width=\textwidth]{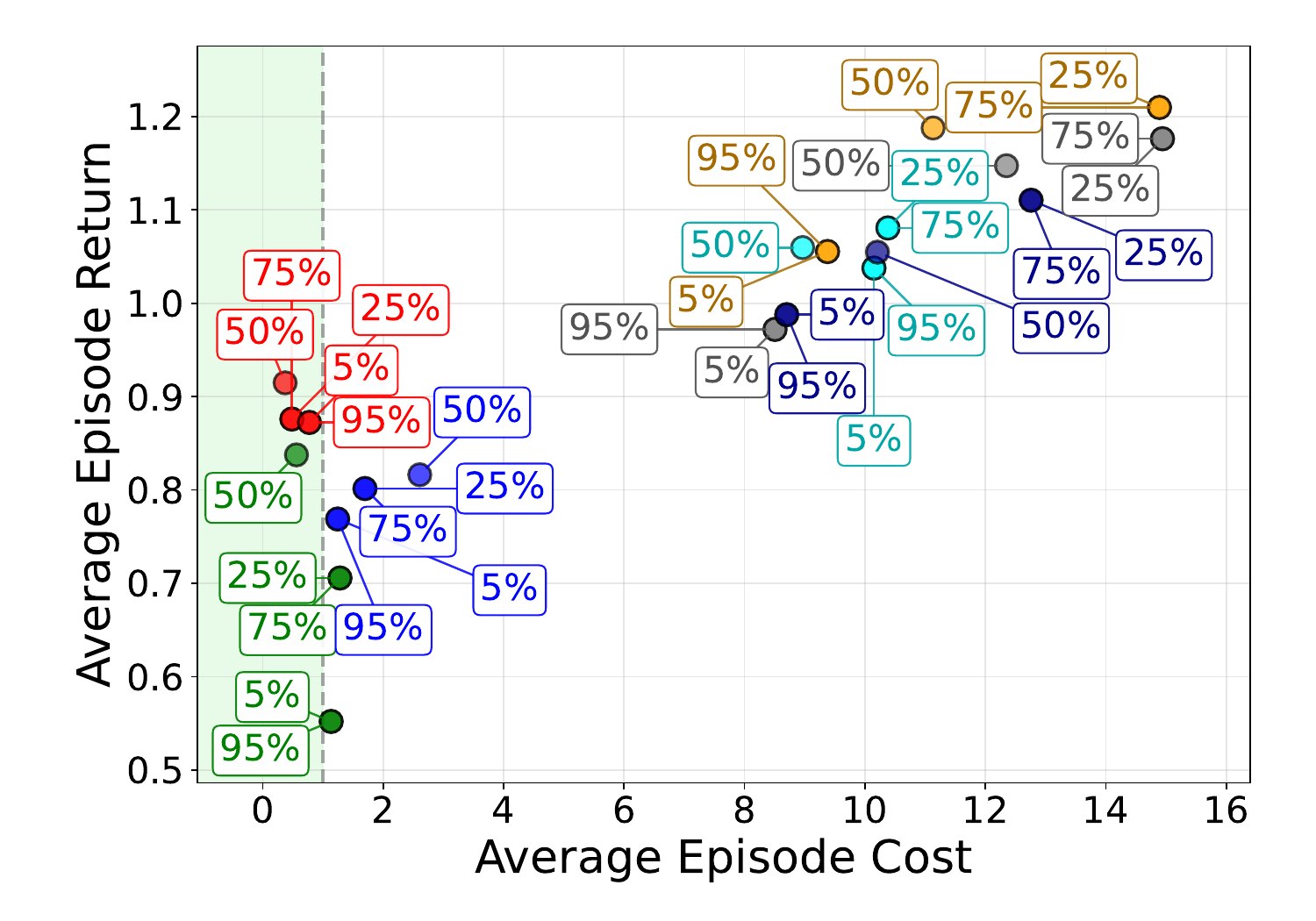}
        \caption{Dependency on $\textbf{quality}(\mathcal{D}_P)$}
        \label{fig:data_quality_dP}
    \end{subfigure}
    \vspace{15pt}
    \caption{
    Data efficiency (a) and sample quality (b) on the safe demonstration set $\DS$ across methods,
    mirroring counterpart plots on the performant demonstration set $\DP$ (c), (d) respectively.
    Results are averaged over 6 tasks.
    Each subplot examines:
    (a) varying $N_{\DS} = 2k$, $k \in\{1,3,5,7,10\}$, with $N_{\DP}=10$ fixed;
    (b) varying $\textbf{quality}(\DS) \in \{5\%, 25\%, 50\%, 75\%, 95\%\}$ with $N_{\DP}=N_{\DS}=20$ and $\textbf{quality}(\DP)=50\%$ fixed;
    (c), (d) the counterparts on $\DP$ with the roles of $\DS$ and $\DP$ exchanged.
    The green shaded region denotes the cost constraint-satisfying area. }
    \label{fig:analysis_on_data_efficiency}
\end{figure*}

To address \textbf{Q1},
we evaluate all methods on the 38 tasks of the re-curated DSRL benchmark, grouped into Safety Gymnasium (21 tasks), Bullet Safety Gym (8 tasks), and MetaDrive (9 tasks).
To construct $\DP$ and $\DS$ appropriate for the dual-expert IL problem, we label each trajectory as performant or non-performant,
and as safe or unsafe, according to predefined return and cost thresholds (see Appendix~\ref{app:DSRLDatasets}\rev{; sensitivity analysis to these thresholds in Appendix~\ref{sec:app.threshold_sensitivity}}).
Note that these are used solely for dataset construction; DexDICE does not require reward or cost during training.
We use 10 randomly sampled performant trajectories to form $\DP$, and another 10 safe trajectories to form $\DS$, while setting $\DU$ to the entire offline dataset.

Table~\ref{tab:main_comparison} reports group-wise aggregates (per-task results in Table~\ref{app:full_dsrl_table}). DexDICE keeps the group-averaged cost below the constraint and attains the highest mean episode return among safe methods on all three benchmarks; per task, it \rev{satisfies the cost constraint on $31$ of $38$ tasks and is the best safe method on $19$}. BC, JFBC, DWBC-S, FISOR-IL, and DemoDICE-U frequently violate safety; CDT and TraC become safer under pseudo-labels but at substantial return cost. DexDICE's margin over DemoDICE-S/-U confirms that distribution matching biases the policy toward suboptimal safe behaviors, whereas the support constraint preserves high performance within safe regions (further comparisons against SafeDICE, COptiDICE, ICSDICE in Appendix~\ref{app:comparison_with_baselines}).

\subsection{Data Efficiency and Quality Robustness}
\label{sec:dataefficiency}

To address \textbf{Q2} (data efficiency) and \textbf{Q3} (quality robustness), we construct \emph{controlled datasets} on six Safety Gymnasium tasks spanning the Circle and Velocity families by mixing four policy types (\texttt{safe-high}, \texttt{safe-medium}, \texttt{unsafe-high}, \texttt{unsafe-medium}): $\DS$ combines \texttt{safe-high}/\texttt{safe-medium}, $\DP$ combines \texttt{safe-high}/\texttt{unsafe-high}, and $\DU$ contains all four (Appendix~\ref{sec:controlled_dataset}). For \textbf{Q2}, we sweep $N_{\DP}$ or $N_{\DS}\in\{2,6,10,14,20\}$ with the other at 10; for \textbf{Q3}, we fix $N_{\DP}{=}N_{\DS}{=}20$ and sweep $\textbf{quality}(\DP)$ or $\textbf{quality}(\DS)\in\{5\%,25\%,50\%,75\%,95\%\}$ with the other at $50\%$, where $\textbf{quality}(\mathcal{D})$ is the \texttt{safe-high} fraction. Each controlled dataset requires training four policies per task, limiting the ablation to this representative subset.

Figure~\ref{fig:analysis_on_data_efficiency}-(a),(c) report six-task averages as $N_{\DS}$ and $N_{\DP}$ vary, respectively. DexDICE maintains safety and attains the highest episode return among safe methods, struggling only at $N_{\DS}{=}2$ where the safe support cannot be reliably estimated.
Figure~\ref{fig:analysis_on_data_efficiency}-(b),(d) show that as $\textbf{quality}(\DS)$ or $\textbf{quality}(\DP)$ degrades, baselines deteriorate sharply while DexDICE sustains both safety and high return, except when the overlap between $\dP$ and $\dS$ becomes extreme (e.g., $\textbf{quality}(\DP){\le}5\%$).
Joint $(\DS,\DP)$ scaling is reported in Figure~\ref{fig:data_efficiency_overall}.
Per-task analyses are in Appendices~\ref{sec:data_efficiency_on_DS}--\ref{app:data_quality_DP}.

In addition, we observe that DexDICE is also robust to $\kappa$ in a reasonable range (Appendix~\ref{sec:sensitivity_kappa}) and to noisy safe demonstrations under moderate corruption (Appendix~\ref{sec:robust_to_noise}).

\subsection{Real-World Mobile Robot Navigation from Human Demonstrations}
\label{sec:exp_realworld}
To address \textbf{Q4}, we next ask whether DexDICE can learn from \emph{human-collected} demonstrations and survive deployment on real hardware beyond simulation.
We use a SUMMIT-XL-Steel mobile robot with a 2D LiDAR for goal-reaching navigation with intermediate obstacles. A single operator records $N{=}114$ joystick teleoperation episodes: 10 \textit{aggressive} runs (fast, possibly unsafe) form $\DP$, 10 \textit{conservative} runs (safe, slow) form $\DS$, and the remaining 94 episodes together with $\DP\cup\DS$ form $\DU$. To smooth control outputs and prevent actuator stress, we apply action chunking~\citep{zhao2023learning} uniformly across methods. Hardware, MDP, training, and evaluation details are in Appendix~\ref{app:realworld}.

\begin{figure}[t]
    \centering
    \sbox0{\includegraphics[width=0.52\textwidth]{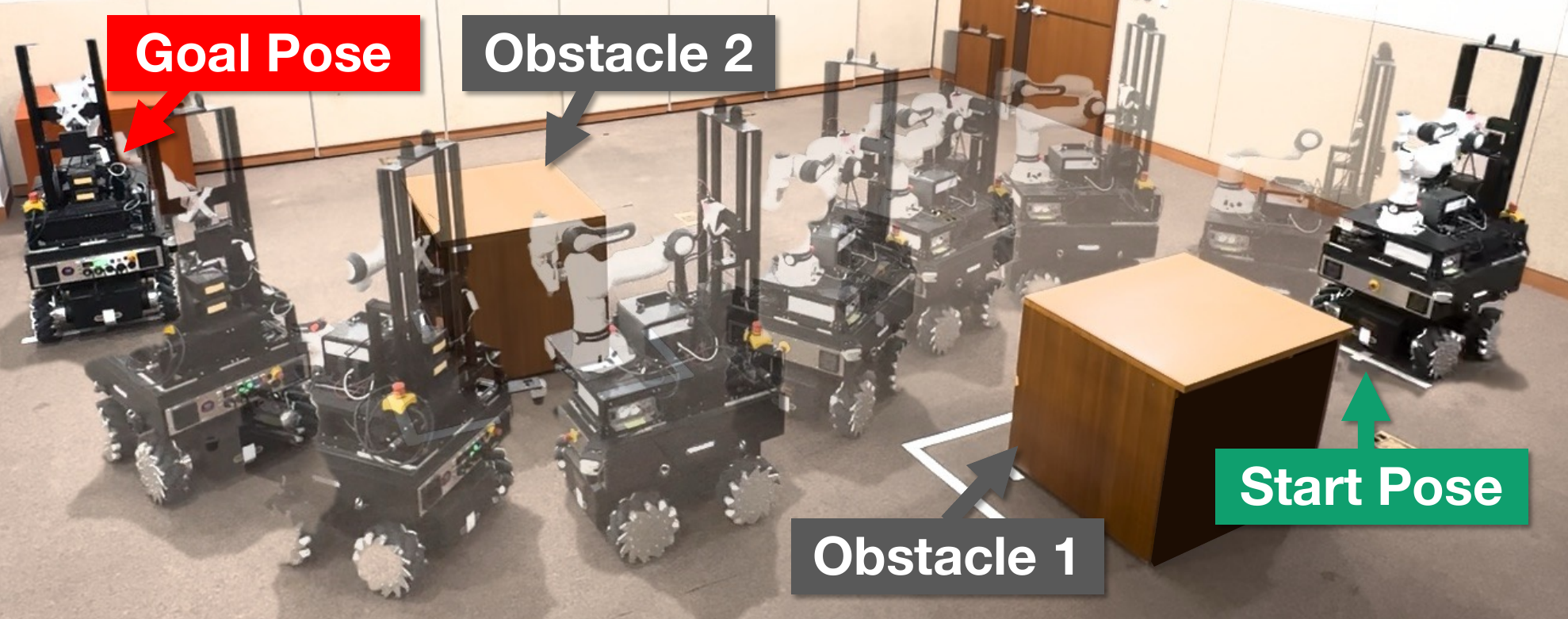}}%
    \begin{subfigure}[b]{0.52\textwidth}
        \centering
        \usebox0
        \caption{Navigation arena with obstacles.}
        \label{fig:realworld_setup}
    \end{subfigure}%
    \hspace{4pt}%
    \begin{subfigure}[b]{0.45\textwidth}
        \centering
        \begin{minipage}[b][\dimexpr\ht0+\dp0\relax][c]{\linewidth}
        \centering
        \resizebox{\linewidth}{!}{%
        \setlength{\tabcolsep}{3pt}%
        \renewcommand{\arraystretch}{1.15}%
        \begin{tabular}{l c r@{\,$\pm$\,}l r@{\,$\pm$\,}l}
            \toprule
            \textbf{Method} & \makecell{\textbf{Success}\\\textbf{Rate (\%)}} & \multicolumn{2}{c}{\makecell{\textbf{Time-}\\\textbf{to-goal}\textbf{(s)}}} & \multicolumn{2}{c}{\makecell{\textbf{Violation}\\\textbf{Rate (\%)}}} \\
            \midrule
            BC-Safe              & 8/10\,(80)            & 36.0          & 6.4          & 3.81          & 1.56 \\
            FBC                  & 6/10\,(60)            & 15.8          & 0.7          & 6.38          & 1.62 \\
            DemoDICE-U           & 8/10\,(80)            & 15.9          & 0.1          & 0.69          & 0.20 \\
            FISOR-IL ($\hat{c}$) & 8/10\,(80)            & 16.9          & 1.3          & 4.89          & 1.53 \\
            \textbf{DexDICE (ours)} & \textbf{10/10\,(100)} & {16.9} & {0.5} & \textbf{0.00} & \textbf{0.00} \\
            \bottomrule
        \end{tabular}}
        \end{minipage}
        \caption{Per-method results (10 trials).}
        \label{tab:realworld_results}
    \end{subfigure}
    \vspace{15pt}
    \caption{Real-world mobile-robot navigation. (a) Arena: start (green) to goal (red) avoiding two obstacles. (b) Per-method results over 10 trials. \textit{Success}: reached goal without collision. \textit{Time-to-goal}: on successful trials. \textit{Violation rate}: fraction of timesteps within an unsafe distance to obstacles.}
    \label{fig:realworld}
    \vspace{0pt}
\end{figure}

Despite hardware-induced uncertainty and only 20 directed human episodes, DexDICE is the only method that reaches the goal in every trial with \emph{zero} safety violations (Figure~\ref{fig:realworld}(b)). BC-Safe, which uses safe demonstrations only, is safer than FBC/FISOR-IL but its conservatism more than doubles time-to-goal; FBC and FISOR-IL match expert speed but violate safety frequently; DemoDICE-U closely tracks DexDICE yet misses the goal in $2/10$ trials.
These results show that the advantages observed on the simulated benchmarks carry over to a noisy, human-driven physical setting (per-trial analysis in Figure~\ref{fig:realworld_per_trial} in Appendix~\ref{app:realworld}).

%% file: contents/main_experiment_table.tex
\begin{table*}[!t]
    \centering
    \begin{adjustbox}{max width=\textwidth}
    \small
    \aboverulesep=0.1ex
    \belowrulesep=0.1ex
    \renewcommand{\arraystretch}{1.05}
    \begin{tabular}{l|l|c|c|c|c|c|c|c|c|c|c}
    \toprule
    \textbf{Task Group} & \textbf{Metric} & \textbf{BC} & \textbf{FBC} & \textbf{JFBC} & \makecell{\textbf{DemoDICE}\textbf{-S}} & \makecell{\textbf{DemoDICE}\textbf{-U}} & \textbf{DWBC-S} & \makecell{\textbf{FISOR-IL} \\ ($\hat{c}$)} & \makecell{\textbf{CDT} \\ ($\hat{r}, \hat{c}$)} & \makecell{\textbf{TraC} \\ ($\hat{r}, \hat{c}$)} & \makecell{\textbf{DexDICE} \\ {(Ours)}} \\
    \midrule\midrule
\multirow{4}{*}{\makecell[l]{\textbf{Safety} \\ \textbf{Gymnasium}\\(21 tasks)}} & Avg return  & \color{gray} 0.46 & \color{gray} 0.13 & \color{gray} 0.59 &  0.27 & \color{gray} 0.47 & \color{gray} 0.45 & \color{gray} 0.58 & \color{gray} 0.43 &  0.19 &  0.37 \\
  & Avg cost  & \color{gray} 1.99 & \color{gray} 1.44 & \color{gray} 2.29 &  0.68 & \color{gray} 1.84 & \color{gray} 1.74 & \color{gray} 2.25 & \color{gray} 1.46 &  0.86 & 0.71 \\
  & \# Safe & 3/21 & 7/21 & 2/21 & 18/21 & 10/21 & 5/21 & 3/21 & 7/21 & 14/21 & 17/21 \\
  & \# Best among safe & 0/21 & 1/21 & 2/21 & 4/21 & 5/21 & 2/21 & 3/21 & 2/21 & 3/21 & \bfseries 9/21 \\
    \midrule
\multirow{4}{*}{\makecell[l]{\textbf{Bullet} \\\textbf{Safety Gym}\\(8 tasks)}} & Avg return  & \color{gray} 0.57 &  0.44 & \color{gray} 0.73 & \color{gray} 0.46 & \color{gray} 0.63 & \color{gray} 0.62 & \color{gray} 0.60 & 0.44 &  0.35 &  0.59 \\
  & Avg cost  & \color{gray} 1.47 &  0.98 & \color{gray} 1.62 & \color{gray} 1.02 & \color{gray} 1.41 & \color{gray} 1.41 & \color{gray} 1.41 & 0.99 &  0.53 & 0.63 \\
  & \# Safe  & 1/8 & 4/8 & 1/8 & 5/8 & 2/8 & 2/8 & 1/8 & 4/8 & 6/8 &  7/8 \\
  & \# Best among safe & 0/8 & 2/8 & 1/8 & 0/8 & 0/8 & 1/8 & 1/8 & 2/8 & 3/8 & \bfseries 5/8 \\
    \midrule
\multirow{4}{*}{\makecell[l]{\textbf{MetaDrive}\\(9 tasks)}} & Avg return & \color{gray} 0.76 &  0.25 & \color{gray} 0.70 &  0.33 & \color{gray} 0.71 & \color{gray} 0.74 & \color{gray} 0.62 &  0.23 &  0.23 &  0.50 \\
  & Avg cost   & \color{gray} 1.68 &  0.50 & \color{gray} 1.85 &  0.28 & \color{gray} 1.79 & \color{gray} 1.69 & \color{gray} 1.68 &  0.04 &  0.04 &  0.69 \\
  & \# Safe & 1/9 & 8/9 & 0/9 &  9/9 & 0/9 & 0/9 & 0/9 & 9/9 &  9/9 & 7/9 \\
  & \# Best among safe & 1/9 & 2/9 & 0/9 & \bfseries 5/9 & 0/9 & 0/9 & 0/9 & 4/9 & 3/9 & \bfseries 5/9 \\
    \bottomrule
    \end{tabular}
    \end{adjustbox}
  \caption{Safety--performance comparison on the re-curated DSRL benchmark (38 tasks), grouped by environment family.
  Return and cost are normalized and averaged over 5 seeds and over the tasks in each group (safe when cost $\le 1$; a lower cost is not preferred once the constraint is met).
  \#~Safe counts tasks satisfying the cost constraint, and \#~Best among safe counts tasks on which a method attains the highest return among safe methods (ties within one standard error).
  All methods use $N_{\DS}{=}N_{\DP}{=}10$; DexDICE uses a single fixed $\kappa=0.01$ for all tasks without per-task tuning.
  \textcolor{gray}{Gray}: group-averaged cost $>1$ (constraint violated). \textbf{Bold}: row maximum in \# Best among safe.
  Per-task results are reported in Table~\ref{app:full_dsrl_table}.}
    \label{tab:main_comparison}
\end{table*}

%% file: contents/6_conclusion.tex
\section{Conclusion}
\vspace{-5pt}

We proposed DexDICE, a framework for learning safe and performant policies from dual-expert demonstrations in the offline regime, without relying on explicit reward or cost signals.
\rev{By replacing distributional regularization toward $d_\rS$ with a safe support constraint, DexDICE decouples safety from the safe expert's suboptimality: it attains the highest return among constraint-satisfying methods on all three families of the re-curated DSRL benchmark, remains data-efficient and robust to demonstration quality, and transfers to a real mobile robot trained from human teleoperation, reaching the goal in every trial without a safety violation. The support constraint further admits a high-probability zero-unsafe-visitation guarantee for the ideal support-constrained policy, calibrated by a single parameter.}

\rev{Our approach has two main limitations.
First, as in any constrained formulation, the dual-expert IL problem requires a feasibility condition: non-trivial overlap between safe and performant demonstrations and a separable safe support.
Our controlled studies show graceful degradation over a wide range of overlap levels, with failures only under extreme scarcity.
Second, the exact guarantee concerns the ideal policy yet the weighted BC policy in practice need not be hard-supported.
This gap cannot be eliminated in general, but deployment-time projection substantially reduces it at a modest cost in return (Appendix~\ref{sec:app.projection}).
The conditions of Theorem~\ref{thm:safety-guarantee} are population-level and are monitored through empirical diagnostics rather than certified (Appendix~\ref{sec:supp_estimation}), and results depend on the dataset-construction thresholds, to which we report sensitivity (Appendix~\ref{sec:app.threshold_sensitivity}).}
A promising direction for future work is to extend our framework to handle multiple safety constraints or multi-objective settings.

%% file: contents/appendix.tex
\renewcommand{\thetable}{\Alph{table}}
\renewcommand{\thefigure}{\Alph{figure}}
\setcounter{theorem}{0}
\setcounter{figure}{0}
\setcounter{table}{0}
\renewcommand \thepart{}
\renewcommand \partname{}
\part{Appendix}
\setcounter{tocdepth}{3}
\iflatexml\else
\etocdepthtag.toc{mtappendix}
\etocsettagdepth{mtchapter}{none}
\etocsettagdepth{mtappendix}{subsubsection}
{\hypersetup{linkcolor=black}\tableofcontents}
\fi
\newpage

\section{Problem Definition}
\label{sec:rigorous_definition}
For any function $f: \mathcal{S} \times \mathcal{A} \to \mathbb{R}$, we denote the expected value under the stationary distribution as:
\begin{equation*}
    J(\pi; f) := \mathbb{E}_{(s,a) \sim d_\pi}[f(s,a)].
\end{equation*}
Given a cost upper bound $\delta_c \in \mathbb{R}_{\geq 0}$ and a performance lower bound $\delta_r \in \mathbb{R}$, we define
$\Pi_\rS$ and $\Pi_\rP$ to be the following sets of policies:\vspace{5pt}
\begin{equation*}
    \Pi_{\rS,\delta_c} := \{ \pi_\rS \in \Pi \mid J(\pi_\rS; c) \leq \delta_c \}
    \quad\text{ and }\quad
    \Pi_{\rP,\delta_r} := \{ \pi_\rP \in \Pi \mid J(\pi_\rP; r) \geq \delta_r \}.
\end{equation*}\vspace{5pt}

Intuitively, $\Pi_\rS$ is the set of safe policies, and $\Pi_\rP$ that of performant policies. Also, for a distribution $d$ on state-action pairs, we write $\mathrm{supp}(d)$ for the support of $d$, i.e., $\mathrm{supp}(d) = \{(s,a) \mid d(s,a) > 0\}$.

\begin{appdefinition}[Dual-Expert Imitation Learning]
\label{def:ilsp}
Let $\delta_c \in \mathbb{R}_{\geq 0}$ and $\delta_r \in \mathbb{R}$ denote a cost upper bound and a performance lower bound, respectively.
Let $\pi_\rS \in \Pi_{\rS,\delta_c}$ and $\pi_\rP \in \Pi_{\rP,\delta_r}$ denote a safe policy and a performant policy, with stationary distributions $d_\rS$ and $d_\rP$.
We assume the feasibility condition
\begin{equation}
    \overline{\Pi} := \{ \pi \in \Pi \mid \mathrm{supp}(d_\pi)
    \subseteq \mathrm{supp}(d_\rS) \cap \mathrm{supp}(d_\rP) \} \neq \emptyset,
    \label{eq:feasibility_assumption}
\end{equation}
where $d_\pi$ is the stationary distribution of policy $\pi$.
Given finite demonstration datasets
\begin{equation*}
    \DS = \{\tau^{(1)}_\rS,\ldots,\tau^{(m)}_\rS\}, \quad
    \tau^{(i)}_\rS \sim \pi_\rS,
    \qquad
    \DP = \{\tau^{(1)}_\rP,\ldots,\tau^{(n)}_\rP\}, \quad
    \tau^{(i)}_\rP \sim \pi_\rP,
\end{equation*}
where each trajectory $\tau = \{s_0, a_0, s_1, a_1, \ldots\}$ is generated by
$s_0 \sim \rho_0$, $a_t \sim \pi(s_t)$, and $s_{t+1} \sim T(s_t,a_t)$, the \emph{dual-expert IL problem} seeks a policy $\pi$ that minimizes
\begin{equation}
\min_{\pi \in \Pi} \; \mathbb{D}(d_\pi \| d_\rP)
\quad \text{subject to} \quad
\mathrm{supp}(d_\pi) \subseteq \mathrm{supp}(d_\rS).
\label{eq:dual_il_objective}
\end{equation}
\end{appdefinition}
The non-emptiness of $\overline{\Pi}$ ensures that some $\pi$ satisfies the support constraint in the problem and also has a finite value when $\mathbb{D}$ is the KL divergence; these imply the feasibility of the above optimization problem in this case. If interaction with the environment is allowed during optimization, the problem is referred to as the online setting; otherwise, it is the offline setting.

Note that dual-expert IL is closely related to cost-constrained IL~\citep{shao2024imitating}.
In particular, a solution to dual-expert IL corresponds to a feasible solution of the cost-constrained IL problem with a hard constraint, where the cost function is non-negative ($c \ge 0$) and the upper bound is fixed to $\delta_c = 0$:
\begin{equation*}
\min_{\pi} \quad \mathbb{D}(d_\pi \| \dP) \quad \text{subject to} \quad J(\pi; c) = 0.
\end{equation*}
This is because the safety condition $J(\pi;c) = 0$ follows from our support constraint $\mathrm{supp}(d_\pi) \subseteq \mathrm{supp}(\dS)$.
To see why, note that $J(\pi_\rS;c) \leq \delta_c = 0$. Together with the non-negativity of $c$, this implies that $c(s,a) = 0$ for all $(s,a) \in \mathrm{supp}(d_\rS)$. Thus, also by our support constraint, we have $c(s,a) = 0$ for all $(s,a) \in \mathrm{supp}(d_\pi)$, which gives $J(\pi;c) = \mathbb{E}_{d_\pi}[c(s,a)] = 0$, that is, $\pi$ satisfies the hard constraint. \rev{The intermediate statement $c(s,a)=0$ on $\mathrm{supp}(d_\rS)$ is precisely the pointwise soundness condition $\mathrm{supp}(d_\rS)\cap\mathcal{N}_{\mathrm{env}}=\emptyset$ with $\mathcal{N}_{\mathrm{env}}:=\{(s,a): c(s,a)>0\}$, under which the safety analysis of Appendix~\ref{sec:app.theory} is stated.}

In practice, demonstrations may be collected from multiple safe policies and multiple performant policies, instead of just one $\piS$ and one $\piP$. This general case is also covered by our setup. A finite set of demonstrations from multiple policies can be viewed as a demonstration set of a single policy obtained by a convex combination of the original multiple policies. Furthermore, if all those original policies are safe (i.e., they are all in $\Pi_{\rS,\delta_c}$), so is their convex combination. Similarly, if the policies are performant (i.e., they belong to $\Pi_{\rP,\delta_r}$), so is their convex combination. Thus, the general case can be understood as an instance of our problem where $\pi_\rS$ and $\pi_\rP$ are appropriate convex combinations of multiple safe and performant policies, respectively.
\newpage
\update{
\paragraph{Dual-Expert IL with a trivial safe-support}
When $d_{\rS}$ has full support, i.e., $d_{\rS}(s,a)>0$ for all $(s,a)\in\mathcal{S}\times\mathcal{A}$, the support constraint in Eq.~\ref{eq:dual_il_objective} becomes vacuous and dual-expert IL collapses to unconstrained stationary-distribution matching, $\min_{d\in\mathcal{Q}}\mathbb{D}(d\,\|\,d_{\rP})$, i.e., DemoDICE without behavior regularization. Analogously to constrained RL with a trivial cost constraint, this reflects that the safe demonstrations carry no safety signal, which underscores the practical importance of collecting informative safe demonstrations.
}
\update{
\paragraph{Dual-Expert IL with no support overlap}
If the safe and performant stationary distributions, $d_{\rS}$ and $d_{\rP}$, have
\[
\mathrm{supp}(d_{\rS}) \cap \mathrm{supp}(d_{\rP}) = \emptyset,
\]
then the feasible region of dual-expert IL becomes severely restricted.
Under the dual-expert IL formulation
\[
\min_{d \in \mathcal{Q}} \; \mathbb{D}(d \,\|\, d_{\rP})
\quad\text{s.t.}\quad
\mathrm{supp}(d)\subseteq \mathrm{supp}(d_{\rS}),
\]
any feasible $d$ must satisfy $d(s,a)=0$ outside $\mathrm{supp}(d_{\rS})$, while $d_{\rP}$ assigns mass exclusively outside this region. Hence no feasible $d$ is absolutely continuous with respect to $d_{\rP}$, and the feasibility condition~(\ref{eq:feasibility_assumption}) is violated.
Under the KL divergence used throughout this paper, this means $D_\mathrm{KL}(d \,\|\, d_{\rP}) = +\infty$ for every feasible $d$, so the optimization is degenerate and admits no finite minimizer in this regime.
For broader choices of $\mathbb{D}$ that remain finite without absolute continuity (e.g., total variation or Wasserstein), the problem is well-posed but reduces to selecting the feasible $d$ that minimizes $\mathbb{D}(d \,\|\, d_{\rP})$ within $\mathrm{supp}(d_{\rS})$, a degenerate solution whose support is disjoint from $d_{\rP}$ and therefore cannot meaningfully imitate it.
In either case, dual-expert IL collapses to imitating only the safe demonstrations, even when $d_{\rP}$ lies entirely outside the safe support.
This limitation is also inherent to constrained optimization and is directly analogous to constrained RL: when the constraint is overly strict, no policy that attains sufficiently high return can satisfy the constrained RL objective.
}

\paragraph{Practical interpretation of the overlap condition}
\label{app:overlap_interpretation}
\rev{The feasibility condition does not require a dataset of entire trajectories that are jointly safe and performant; it requires the separately collected sources to share enough state-action support to admit a Bellman-flow-feasible policy, which can hold even when the two demonstrators differ substantially at the trajectory level. Representative cases include: (i) autonomous driving, where human driver logs contain both fast, risky maneuvers and rule-compliant driving~\citep{bansal2018chauffeurnet,campolettano2025potential}, and the overlap consists of efficient routes that still respect the limits; (ii) human-robot collaboration, where a manipulator moves fast when the workspace is clear but slows near a person~\citep{lasota2014toward}, so that efficient-and-safe motions populate the overlap; and (iii) changing safety regulations, where a revised rule (e.g., updated speed limits, no-go zones, or clearance standards) redefines the safe region, so that only the safe demonstrations need re-collecting or re-labeling while the performant demonstrations remain reusable, and the overlap is the efficient behavior that still complies. The last case is precisely where labeling demonstrations is cheaper than re-engineering a reward or cost function. Our real-robot navigation experiment (Section~\ref{sec:exp_realworld}) is a further instance: safe demonstrations keep clearance from obstacles, performant ones take faster routes, and the overlap is the set of fast routes that pass through safe corridors.}

\clearpage

\section{Detailed Discussions on Related Work}
\label{app:detailed_related_work}

\begin{table}[hpt!]
\centering
\small
\aboverulesep=0.1ex
\belowrulesep=0.1ex
\renewcommand{\arraystretch}{1.7}
\begin{adjustbox}{max width=1\textwidth}
\begin{tabular}{l | l | c c c c c | c c}
\toprule
\textbf{DICE Method} & \textbf{Target Offline Problem} &
\textbf{Reward} & \textbf{Cost} & \textbf{$\mathcal{D}_\rP$} &
\textbf{$\mathcal{D}_\rS$} & \textbf{$\mathcal{D}_\mathrm{N}$} & \textbf{Safety} &
\textbf{Constraints}
\\
\midrule\midrule
OptiDICE~\small{\citep{lee2021optidice}} & RL & $\cmark$ & $\xmark$ & $\xmark$ & $\xmark$ & $\xmark$ &  $\xmark$ & Distribution  \\
\cmidrule(lr){1-9}
\makecell[l]{DemoDICE~\small{\citep{kim2022demodice}},
\\SMODICE~\small{\citep{ma2022versatile}},
\\RelaxDICE~\small{\citep{yu2023offline}}}
& \makecell[l]{IL} & $\xmark$ & $\xmark$ & $\cmark$ & $\xmark$ & $\xmark$ & $\xmark$ & Distribution  \\ \cmidrule(lr){1-9}
SafeDICE~\small{\citep{jang2023safedice}} & \makecell[l]{IL} &  $\xmark$ &  $\xmark$ & $\xmark$ &   $\xmark$ &  $\cmark$ & $\cmark$ & Distribution \\
\cmidrule(lr){1-9}
COptiDICE~\small{\citep{leecoptidice}} & Constrained RL & $\cmark$ & $\cmark$ &  $\xmark$ & $\xmark$ &  $\xmark$ & $\cmark$ & \makecell[c]{Distribution,\\ Cost (true)}\\
\cmidrule(lr){1-9}
ICSDICE~\small{\citep{quan2024learning}} & Inverse Constrained RL & $\cmark$ &  $\xmark$ &  $\xmark$ & $\cmark$  & $\xmark$ & $\cmark$ &
\makecell[c]{Distribution, \\Cost (inferred)} \\
\midrule
\textbf{DexDICE (Ours)} & Dual-expert IL & $\xmark$ & $\xmark$ & $\cmark$ & $\cmark$ & $\xmark$ & $\cmark$ &
{Support}\\
\bottomrule
\end{tabular}
\end{adjustbox}
\vspace{5pt}
\caption{Structural comparison of DICE methods for offline problems in terms of required learning signals, safety constraints, and dataset requirements.
$\mathcal{D}_\mathrm{N}$ indicates a set of non-preferred demonstrations, which corresponds to unsafe demonstrations in our setting.
}\label{tab:dice_comparison_app}
\vspace{-15pt}
\end{table}

In this section, we provide a detailed discussion of DICE-based approaches developed for offline learning problems that are closely related to our setting. Table~\ref{tab:dice_comparison_app} summarizes representative methods in terms of required learning signals, safety considerations, and dataset requirements.
DICE methods have been widely studied in both RL and IL because they bypass explicit sequential modeling of MDP dynamics by optimizing over stationary distributions constrained to satisfy the Bellman flow condition.
This perspective leads to linear programming objectives with tractable convex dual forms, making DICE formulations particularly appealing for offline settings.

\subsection{Conceptual Comparison with DICE Methods}
\label{app:discussion_related_work}

A representative DICE method for offline RL is OptiDICE~\citep{lee2021optidice}, whose objective is
\[
\max_{d \in \mathcal{Q}} \; \mathbb{E}_{d}[r(s,a)]
\;-\;
\alpha\, D_f(d \,\|\, d_\rU),
\]
where $\mathcal{Q}$ denotes the set of stationary distributions. The $f$-divergence regularizer plays a central role by enforcing conservatism, preventing the learned distribution from drifting outside the support of the offline dataset, which is a standard requirement for stable offline RL.
COptiDICE~\citep{leecoptidice} extends OptiDICE to the CMDP setting.
DemoDICE~\citep{kim2022demodice} adapts this idea to offline IL using the objective
\[
\max_{d \in \mathcal{Q}}
\;-\; D_{\mathrm{KL}}(d \,\|\, d_\rP)
\;-\;
\alpha\, D_{\mathrm{KL}}(d \,\|\, d_\rU),
\]
which encourages the learned stationary distribution to remain close to both the performant demonstrations $d_\rP$ and the offline dataset $d_\rU$. As shown in Table~\ref{tab:dice_comparison_app}, DemoDICE also relies on distributional proximity to $d_\rU$, which may be problematic in dual-expert IL;
if the offline dataset contains suboptimal or unsafe behavior, forcing closeness to $d_\rU$ can degrade either safety or performance.
SafeDICE~\citep{jang2023safedice} leverages non-preferred demonstrations $\mathcal{D}_\mathrm{N}$ and assumes that $\DU$ is a mixture of expert and non-preferred data.
ICSDICE~\cite{quan2024learning} assumes that $\dS$ is induced from an optimal CMDP policy and deems $(s,a)$ risky if it is more likely under a superior stationary distribution, namely, the stationary distribution of a policy that outperforms $\dS$.
As discussed in Section~\ref{sec:related_work},
replacing $d_\rU$ with $d_\rS$ (safe demonstrations) in DemoDICE can partially mitigate safety concerns, but still requires staying close to possibly low-quality safe demonstrations, making the method sensitive to data quality.

\update{
To address this limitation, DexDICE explicitly replaces the  distributional regularizer with a support constraint:
\[
\max_{d \in \mathcal{Q}} \; - D_{\mathrm{KL}}(d \,\|\, d_\rP)
\quad\text{subject to}\quad
\mathrm{supp}(d) \subseteq \mathrm{supp}(d_\rS).
\]
This formulation seeks the stationary distribution that is closest to $d_\rP$ while remaining entirely within the safe support defined by $d_\rS$. Among the DICE methods in Table~\ref{tab:dice_comparison_app}, DexDICE alone enforces safety through an explicit support constraint without requiring reward signals, cost signals, or non-preferred demonstrations. In the tabular comparison of Appendix~\ref{sec:app.toy_experiment}, this avoids the return degradation caused by distributional penalties toward suboptimal $d_\rS$.
}

Although it is not a DICE-based method, DWBC~\citep{xu2022discriminator} is closely related to our setting as an offline IL approach that leverages unlabeled demonstrations. In its original formulation, DWBC filters expert-like samples from unlabeled data by optimizing the discriminator $f$ by using the following PU-learning objective:
\begin{align*}
&\min_{f} \; \mathbb{E}_{(s,a)\sim \dP}[- \log f(s,a)] + \mathbb{E}_{(s,a)\sim d_\mathrm{N}}[- \log (1-f(s,a))] \\
= \;&\min_{f} \; \eta\, \mathbb{E}_{(s,a)\sim \dP}[-\log f(s,a)] + \mathbb{E}_{(s,a)\sim \dU}[-\log(1-f(s,a))] - \eta \,
\mathbb{E}_{(s,a)\sim \dP}[- \log (1-f(s,a))],
\end{align*}
where $\eta$ denotes the class prior. The policy is then obtained via discriminator-weighted behavior cloning, i.e.,
$\min_\pi \mathbb{E}_{(s,a)\sim \DU}[ - f(s,a)\log\pi(a|s)]$.
In the context of safe IL, we compare against DWBC-S, a variant of DWBC that substitutes $\DS$ for $\DU$ throughout, so that expert-like samples are filtered from $\DS$ rather than from $\DU$ and the policy is trained by minimizing $\mathbb{E}_{(s,a)\sim \DS}[- f(s,a)\log\pi(a|s)]$.

\subsection{Detailed Discussion of Offline Safe RL Baselines}
\label{app:offline_safe_rl_discussion}

This section elaborates on the offline safe RL methods discussed in Section~\ref{sec:related_work}, why they are not directly applicable to the dual-expert IL setting, and how we adapt them for the comparison reported in Section~\ref{sec:dsrl_experiments}.

\paragraph{Offline safe RL methods.}
These methods take a constrained MDP as input, with explicit per-transition reward $r(s,a)$ and cost $c(s,a)$ provided in the offline dataset, and learn a policy that maximizes return subject to a cost budget.
\begin{itemize}[leftmargin=6mm]
    \item \textbf{COptiDICE}~\citep{leecoptidice} extends OptiDICE~\citep{lee2021optidice} to constrained MDPs by introducing a Lagrangian term for the cost expectation.
    \item \textbf{CDT}~\citep{liu2023constrained} conditions a Decision Transformer~\citep{chen2021decision} on both return-to-go and cost-to-go targets computed from each trajectory's reward and cost signals.
    \item \textbf{FISOR}~\citep{zheng2024safe} casts safety as a reachability problem, learning a cost value function $V_c$ from $c(s,a)$ and using it to mask infeasible actions for a guided diffusion policy.
    \item \textbf{C2IQL}~\citep{liu2025c2iql} extends IQL with a Lagrangian dual: separate Q-functions for reward and cost are learned via expectile regression and balanced by a learnable multiplier.
    \item \textbf{TraC}~\citep{yang2024offline} reframes offline safe RL as trajectory classification: trajectories are scored by predicted return and total cost, then weighted behavioral cloning recovers the policy.
\end{itemize}
All five methods consume per-transition $(r, c)$ supervision, which our dual-expert IL setting does not provide.

\paragraph{Why naive labeling fails.}
A natural attempt to apply these methods to our dual-expert IL setting is to construct labels from the dataset identity alone, e.g.,
\begin{align*}
\hat r(s,a) = \mathds{1}[(s,a) \in \DP], \qquad \hat c(s,a) = \mathds{1}[(s,a) \notin \DS].
\end{align*}
This labeling is biased in two ways. First, $\mathds{1}[(s,a) \notin \DS]$ marks every transition outside $\DS$ as unsafe, including a substantial number of safe transitions in $\DU$, since $\DU$ is a latent mixture of safe and unsafe behavior. Second, $\DS$ contains only positive (safe) examples while $\DU$ is unlabeled rather than negative; this is exactly the positive--unlabeled (PU) learning regime~\citep{bekker2020learning,elkan2008learning}, in which naive binary classification systematically over-estimates the negative class. Cost Q-functions trained on such biased labels inflate the unsafe region, leading either to overly conservative behavior or to unreliable feasibility masks that admit unsafe actions.

\paragraph{Pseudo-label adaptation.}
For empirical comparison under matched information access, we threshold DexDICE's learned discriminators $\hat{r}_\rP$ and $\hat{r}_\rS$ to obtain binary reward and cost pseudo-labels. Let $\tau_\rP$ and $\tau_\rS$ denote the corresponding discriminator-score thresholds:
\begin{align*}
\hat r(s,a) := \mathds{1}\{\hat{r}_\rP(s,a) \ge \tau_\rP\}, \qquad
\hat c(s,a) := \mathds{1}\{\hat{r}_\rS(s,a) < \tau_\rS\}.
\end{align*}
These labels assign $(\hat r,\hat c)\in\{0,1\}^2$ to transitions in $\DU$ using only $(\DP,\DS,\DU)$, without ground-truth reward or cost signals.
Section~\ref{sec:dsrl_experiments} reports the resulting performance for FISOR, CDT, and TraC under this pseudo-label scheme, and Appendix~\ref{app:comparison_with_baselines} under the oracle $(r, c)$ scheme; DexDICE outperforms the pseudo-label variants on the majority of tasks and remains competitive with the oracle variants despite using strictly less information.

\section{Tabular Experiment: Distribution vs.\ Support Constraint}

\label{sec:app.toy_experiment}

This section provides a controlled comparison between distribution constraints and support constraints using a tabular domain where exact solutions can be computed.
The key question is: \emph{how do these two constraint types behave when the quality of $\dS$ or $\dP$ varies?}

\begin{figure*}[hpt!]
    \vspace{-10pt}
    \centering
    \begin{subfigure}[b]{0.27\textwidth}
        \centering
        \includegraphics[width=\textwidth]{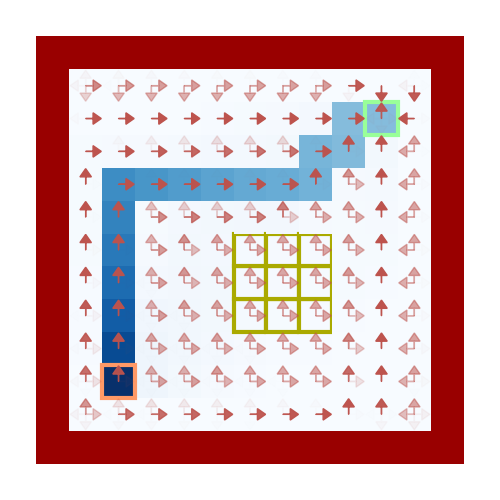}
        \caption{Constrained gridworld.}
        \label{fig:toy_problem}
    \end{subfigure}
    \hfill
    \begin{subfigure}[b]{0.72\textwidth}
        \centering
        \includegraphics[width=\textwidth]{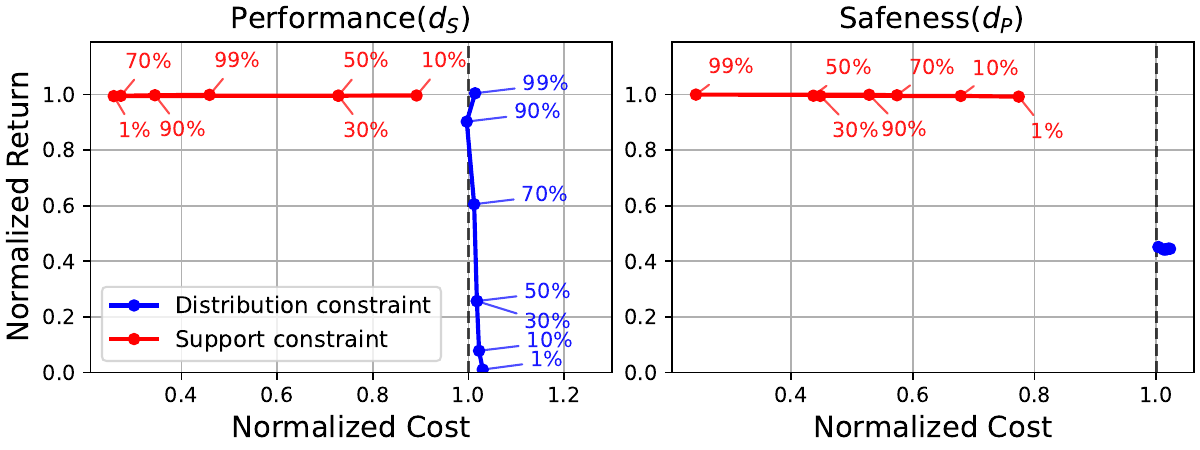}
        \caption{Return vs.\ cost under varying $\dS$ and $\dP$ quality.}
        \label{fig:toy_result}
    \end{subfigure}
    \caption{An illustration on tabular experiment.
    (a) The agent must reach the goal while avoiding unsafe zones.
    (b) Each point shows the mean return and cost (averaged over 50 seeds) for a given demonstration quality (indicated by labels).
    The support constraint remains in the high return, low cost region across all settings,
    while the distribution constraint degrades when $\dS$ is suboptimal.}
    \label{fig:tabular_experiments}
    \vspace{-10pt}
\end{figure*}

\subsection{Environment Setup}
We design an $11 \times 11$ constrained gridworld (Figure~\ref{fig:toy_problem}) where the agent navigates from start (orange) to goal (green) while avoiding $3 \times 3$ unsafe zones (olive) along the diagonal.
The environment has stochastic transitions ($0.99$ success), $\gamma = 0.95$, reward $1$ at the goal, cost $1$ in unsafe zones, and cost threshold $\delta_c = 10^{-4}$.
This tabular domain enables exact computation of stationary distributions, isolating constraint effects without function approximation.

\paragraph{Constructing $\dP$ and $\dS$}
To systematically vary demonstration quality, we define:
\begin{itemize}[leftmargin=6mm]\vspace{-5pt}
    \item $d^*_{\mathcal{M}}$: stationary distribution of the unconstrained MDP-optimal policy.  %
    \item $d^*_{\mathcal{M}_c}$: stationary distribution of the CMDP-optimal policy.  %
    \item $d^o_{\mathcal{M}_c}$: stationary distribution of a safe policy with performance level $o \in [0,1]$.\vspace{-5pt}
\end{itemize}
We construct $d_\rP = (1-\xi)\, d^*_{\mathcal{M}} + \xi\, d^*_{\mathcal{M}_c}$ where $\xi \in [0,1]$ controls safeness, and $d_\rS = d^o_{\mathcal{M}_c}$ where $o$ controls performance.
This allows us to independently control the \emph{safeness} of $\dP$ and the \emph{performance} of $\dS$.

\paragraph{Hyperparameters}
For the distribution constraint, we sweep $\alpha \in [10^{-2}, 10^{2}]$.
For the support constraint, we define $\mathds{1}_{\dS}$ by thresholding low-probability state-action pairs, with thresholds in $[10^{-10}, 10^{-4}]$.
We report the highest return among feasible policies (i.e., $J(\pi;c) \leq \delta_c$); if none exist, we report the policy with the lowest cost.

\subsection{Results}

We systematically vary the safeness of $\dP$ and the performance of $\dS$, then solve each objective exactly.
From each solution $d$, we extract a policy via $\pi(a|s) = d(s,a) / \sum_{\bar{a}} d(s,\bar{a})$ and evaluate its return and cost.
All results are averaged over 50 random seeds with randomized constraint zones.

Figure~\ref{fig:toy_result} reveals a clear behavioral difference between the two constraint types:
\begin{itemize}[leftmargin=6mm]
    \item \textbf{Distribution constraint} pulls the learned distribution toward $\dS$, causing the policy to inherit suboptimal behaviors when $\dS$ has low performance. Even with high-quality $\dP$, the regularization toward $\dS$ limits the achievable return.
    \item \textbf{Support constraint} restricts only \emph{where} the policy can act, not \emph{how often}. This allows the policy to align with $\dP$ within safe regions, achieving high returns even when $\dS$ is suboptimal.
\end{itemize}
This confirms the key advantage of support constraints: they decouple safety from suboptimality, enabling robust performance across varying demonstration qualities.

\section{Proofs of Propositions}
\label{sec:app.proof}

\subsection{Derivation of the Lagrangian Objective $L(w,\nu,\lambda)$}
\label{app:lagrangian_derivation}

\rev{Throughout this section, $d$ ranges over the domain $\mathcal{C}$ of normalized nonnegative state-action distributions that are supported on $\supp(\dS)$, absolutely continuous with respect to $\dP$, and have finite $D_\mathrm{KL}(d\|\dP)$. Every $d\in\mathcal{C}$ therefore assigns zero mass outside $\supp(\dS)\cap\supp(\dP)$. Since $\dP\ll\dU$ (Section~\ref{sec:method}), $d\ll\dP$ implies $d\ll\dU$, so the correction ratio $w = d/\dU$ is well-defined $\dU$-almost everywhere.}
Starting from the constrained problem in Eq.~(\ref{eq:method_consobj}), we introduce Lagrange multipliers $\lambda \in \mathbb{R}$ for the safe-support constraint and $\nu(s) \in \mathbb{R}$ for the Bellman flow constraints:
\begin{align}
\min_{d\ge 0} \max_{\nu,\lambda}\, & D_\mathrm{KL}(d \| \dP) + \lambda \left( 1 - \sum_{s,a} \mathds{1}_{\dS}(s,a) d(s,a) \right) \notag \\
& + \sum_{s} \nu(s) \big( (\mathcal{B}_* d)(s) - (1-\gamma) \rho_0(s) - \gamma (\mathcal{T}_* d)(s) \big). \label{eq:app.lagrangian}
\end{align}
Flipping the sign and rearranging terms yields the equivalent max-min form
\begin{align}
\max_{d\ge 0} \min_{\nu, \lambda}\, & (1-\gamma) \mathbb{E}_{s \sim \rho_0}[\nu(s)] - \lambda \notag \\
& + \mathbb{E}_{(s,a)\sim d} \Big[ \gamma (\mathcal{T}\nu)(s,a) - \nu(s) - \log \frac{d(s,a)}{\dP(s,a)} + \lambda \mathds{1}_{\dS}(s,a) \Big],
\end{align}
where $(\mathcal{T}\nu)(s,a) := \mathbb{E}_{s' \sim T(s,a)}[\nu(s')]$. Applying the change of variables $w(s,a) := d(s,a)/\dU(s,a)$ and defining $r_\rP(s,a) := \dP(s,a)/\dU(s,a)$, we obtain
\begin{align}
& (1-\gamma) \mathbb{E}_{s \sim \rho_0}[\nu(s)] - \lambda + \mathbb{E}_{(s,a)\sim \dU} \big[ w(s,a) \cdot ( -\log w(s,a) + A_{\nu,\lambda}(s,a) ) \big] \notag \\
&\qquad =: L(w,\nu,\lambda),
\end{align}
\rev{which is Eq.~(\ref{eq:loss_def}) in the main text, and} where the advantage term is
\begin{align}
A_{\nu,\lambda}(s,a) := \gamma (\mathcal{T}\nu)(s,a) - \nu(s) + \log r_\rP(s,a) + \lambda \mathds{1}_{\dS}(s,a).
\end{align}
The DexDICE objective is then $\min_{\nu,\lambda} \max_{w \ge 0} L(w,\nu,\lambda)$, which by strong duality (Proposition below) and the closed-form solution $w^*_{\nu,\lambda}(s,a) = \exp(A_{\nu,\lambda}(s,a) - 1)$ reduces to the convex minimization in Eq.~(\ref{eq:dexdice_obj}).

\subsection{Strong Duality and Closed-Form Solution}

Rewrite an objective~(\ref{eq:dexdice_obj}):
\begin{equation*}
L(w,\nu, \lambda) = (1-\gamma) \mathbb{E}_{s \sim \rho_0}[\nu (s) ] + \mathbb{E}_{(s,a)\sim d_U}\left[ w(s,a) (A_{\nu,\lambda}(s,a) -\log w(s,a) ) \right]  - \lambda.
\end{equation*}

\setcounter{theorem}{0}%
\begin{proposition}
(Strong Duality) $L(w,\nu,\lambda)$ satisfies strong duality which ensures:
\begin{equation*}
\max_{w \ge 0} \min_{\nu, \lambda } L(w,\nu,\lambda) =  \min_{\nu, \lambda } \max_{w \ge 0} L(w,\nu,\lambda)
\end{equation*}
\end{proposition}
\begin{proof}
\rev{On the domain $\mathcal{C}$ defined in Appendix~\ref{app:lagrangian_derivation}, the objective $D_\mathrm{KL}(d\|\dP)$ is convex in $d$, the safe-support constraint $\sum_{s,a}\mathds{1}_{\dS}(s,a)d(s,a)=1$ holds automatically, and the Bellman flow constraints are affine in $d$. Problem~(\ref{eq:method_consobj}) is therefore a convex program with affine constraints only. We assume the standard constraint qualification for this case~\citep{boyd2004convex}: the affine set $\{d : (\mathcal{B}_*d)(s) = (1-\gamma)\rho_0(s) + \gamma(\mathcal{T}_*d)(s)\ \ \forall s\}$ intersects the relative interior of $\mathcal{C}$. Under this condition, strong duality holds, so the primal optimum equals the dual optimum and the max--min and min--max values of $L(w,\nu,\lambda)$ coincide. We note that reachability of every state alone does not provide this condition once the support constraint is active, since a strictly feasible $d$ must additionally be supported on $\supp(\dS)\cap\supp(\dP)$; the constraint qualification above makes the required feasibility explicit.}
\end{proof}

\begin{proposition}[Closed-form solution]
For any dual variables $\nu$ and $\lambda$, the inner maximization
\[
\max_{w \ge 0} L(w, \nu, \lambda)
\]
admits the closed-form solution
\[
w^*_{\nu,\lambda}(s,a) = \exp\bigl( A_{\nu,\lambda}(s,a) - 1 \bigr).
\]
\end{proposition}

\begin{proof}
Recall the following lemma (Proposition 1 in \citep{leecoptidice}):

\smallskip
\noindent
\begin{lemma*}
Let $f$ be a convex function (e.g. for KL divergence, $f(x)=x\log x$) with strictly increasing derivative $f'$.
Given $\nu$, $\lambda$, and $\alpha \ge 0$, define
\[
e_{\nu,\lambda}(s,a) := r(s,a) - \lambda c(s,a) + \gamma \, \mathbb{E}_{s' \sim T(s,a)}[\nu(s')] - \nu(s).
\]
Then, the maximization
\[
\max_{w \ge 0} \;
\mathbb{E}_{(s,a)\sim d_U}\bigl[\,w(s,a)\,e_{\nu,\lambda}(s,a) - \alpha\,f(w(s,a))\bigr]
\]
is achieved by
\[
w^*_{\nu,\lambda}(s,a) = \bigl( f' \bigr)^{-1}\!\left( \tfrac{1}{\alpha}\, e_{\nu,\lambda}(s,a) \right)_{+},
\]
where $(\cdot)_{+} = \max(0, \cdot)$.
\end{lemma*}
\smallskip
Applying this lemma in our setting: we substitute
\[
r = \log r_P,\quad c = -\mathds{1}_{d_S},\quad \alpha = 1,\quad \delta_c = -1,
\]
which yields $e_{\nu,\lambda} = A_{\nu,\lambda}$ and $L = \mathcal{L}$ in the canonical form.
Since for KL divergence one has $f(x) = x \log x$ and thus $(f')^{-1}(y) = \exp(y - 1)$, we conclude
\[
w^*_{\nu,\lambda}(s,a) = \exp\bigl(A_{\nu,\lambda}(s,a) - 1 \bigr).
\]
\end{proof}

\section{Safety Guarantee of DexDICE}
\label{sec:app.theory}

This section provides the full statement and proof of the safety guarantee informally stated as Theorem~\ref{thm:safety-guarantee} in the main paper, relating the policy's unsafe-region visitation to the discriminator's error rates.

\paragraph{Setup and notation.}
Consider a constrained MDP with state space $\mathcal{S}$ and action space $\cA$. We are given three offline datasets $\DS, \DP, \DU$, with stationary state-action distributions $\dS, \dP, \dU$. \rev{Let $\mathcal{N}_{\mathrm{env}}\subseteq\mathcal{S}\times\cA$ denote the set of state-action pairs that are unsafe in the environment. We assume \emph{pointwise soundness} of the safe demonstrations, $\supp(\dS)\cap\mathcal{N}_{\mathrm{env}}=\emptyset$, i.e., every demonstrated safe pair is environmentally safe. Under this assumption, agreement with the demonstrated support implies environmental safety, so the zero-visitation guarantee below for the unsafe region $\bU$ (the reachable complement of the demonstrated safe support) is a statement about environmental safety; without it, the guarantee concerns support agreement only.} Define:
\begin{itemize}[leftmargin=6mm,itemsep=2pt]
  \item \textbf{Safe region:} $\bS = \supp(\dS) = \{(s,a) : \dS(s,a) > 0\}$ and \textbf{(reachable) unsafe region} $\bU = \supp(\dU)\setminus\bS = \{(s,a) : \dS(s,a)=0,\ \dU(s,a)>0\}$. %
Restricting $\bU$ to $\supp(\dU)$ guarantees $\dU(s,a)>0$ on $\bU$; the residual set $\bS^c\setminus\supp(\dU)$ is never visited under Assumption~\ref{ass:bounded} (Proposition~\ref{prop:visitation}).
  \item \textbf{Bayes-optimal discriminator:} $\rstar(s,a) := \dS(s,a) / (\dS(s,a) + \dU(s,a))$, well-defined on $\bS\cup\bU = \supp(\dS)\cup\supp(\dU)$ since the denominator is strictly positive there. By construction $\rstar(s,a) > 0$ on $\bS$ and $\rstar(s,a) = 0$ for all $(s,a)\in\bU$ (where $\dS=0$ and $\dU>0$).
  \item \textbf{Estimated safe support:} $\hbS_\tau = \{(s,a) : \rhat(s,a) \ge \tau\}$, where $\rhat$ is the learned discriminator.
  \item \textbf{Conditional unsafe distribution:} $d_\bU(s,a) := \dU(s,a)\,\mathds{1}[(s,a)\in\bU] / \dU(\bU)$, the conditional distribution of $\dU$ given $(s,a)\in\bU$. We assume $\dU(\bU) > 0$ throughout; if $\dU(\bU) = 0$, then $\supp(\dU)\subseteq \bS$, $\bU=\emptyset$, and the safety guarantee holds vacuously.
  \item \textbf{False-positive rate:} $\FPR(\tau) = \mathbb{P}_{(s,a)\sim d_\bU}[(s,a) \in \hbS_\tau] = \dU(\bU \cap \hbS_\tau)/\dU(\bU)$.
\end{itemize}
\rev{Let $n_S := |\DS|$ and $\hat F(\tau) := \frac{1}{n_S}\sum_{(s,a)\in\DS}\mathds{1}\{\rhat(s,a)\le\tau\}$ denote the empirical CDF of the scores over $\DS$. For a user-specified $\kappa\in(0,1)$, $\tau_\kappa = \inf\{\tau : \hat F(\tau)\ge\kappa\}$ is the empirical $\kappa$-quantile of Eq.~(\ref{eq:supp_estimator}). A pair is accepted when $\rhat(s,a)\ge\tau_\kappa$ and rejected when $\rhat(s,a)<\tau_\kappa$.
Our analysis considers an ideal support-constrained policy $\pi_\kappa$ whose stationary distribution satisfies $\supp(d^{\pi_\kappa})\subseteq\hbS_{\tau_\kappa}$;
The policy extracted by weighted BC need not satisfy this inclusion exactly (quantified and mitigated in Appendix~\ref{sec:app.projection}), and the exact guarantee below is stated for $\pi_\kappa$.}

\rev{
\begin{appassumption}[i.i.d.\ calibration sample]
\label{ass:calib}
The $n_S$ calibration pairs $\{(s_i,a_i)\}_{i=1}^{n_S}$ used to compute $\tau_\kappa$ are i.i.d.\ draws from $\dS$.
\end{appassumption}
}

\begin{appassumption}[Uniform estimation error]
\label{ass:err}
The learned discriminator $\rhat$ satisfies $\|\rhat - \rstar\|_\infty \le \varepsilon$ with probability at least $1-\delta/2$, where $\varepsilon=\varepsilon(n)$, $n=|\DS|+|\DU|$, and $\varepsilon(n)\to0$ as $n\to\infty$. Denote this event by $\mathcal{E}:=\{\|\rhat-\rstar\|_\infty\le\varepsilon\}$.
\end{appassumption}

\begin{appassumption}[Bounded DICE weights]
\label{ass:bounded}
The learned policy $\pi$ satisfies $d^\pi \ll \dU$ (absolute continuity) and $d^\pi(s,a) \le \Cw \cdot \dU(s,a)$ for all $(s,a)$, where $\Cw < \infty$ can be controlled by gradient-penalty regularization in the DICE objective.
\end{appassumption}

\begin{appassumption}[Safe margin]
\label{ass:margin}
Define the \emph{safe margin}
$\mu = \inf\{\rstar(s,a) : (s,a) \in \bS,\ \dS(s,a) \ge \rho\}$
for a regularity parameter $\rho > 0$, and the \emph{boundary mass}
$\kappa_0 = \mathbb{P}_{\dS}(\rstar(s,a) \rev{\le} \mu)$.
The safe margin satisfies $\mu > 2\varepsilon$.
\end{appassumption}

\rev{
\begin{remark*} %
None of the assumptions requires exact recovery of the safe region, its density, or a cost function. Pointwise soundness requires only that $\DS$ be uncontaminated, not that it cover the full safe region. Assumption~\ref{ass:err} requires a uniform bound on the discriminator score rather than recovery of either density, and Assumption~\ref{ass:margin} requires a margin only on the retained safe interior. Both are standard plug-in conditions for thresholded density-ratio classifiers~\citep{tong2013plugin}. Since these are population-level conditions, their empirical diagnostics are reported in Appendices~\ref{sec:supp_estimation} and \ref{sec:sensitivity_kappa}.
\end{remark*}
}

\begin{appproposition}[Unsafe-visitation decomposition]
\label{prop:visitation}
Let $\pi$ be any policy supported on $\hbS_\tau$, and suppose Assumption~\ref{ass:bounded} holds. Then
\[
d^\pi(\bS^c) \;=\; d^\pi(\bU) \;\le\; \Cw \cdot \FPR(\tau) \cdot \dU(\bU).
\]
\end{appproposition}
\begin{proof}
Under Assumption~\ref{ass:bounded}, $d^\pi\ll\dU$, so $d^\pi$ assigns zero mass to $\bS^c\setminus\supp(\dU)$ and $d^\pi(\bS^c)=d^\pi(\bU)$. Since $\pi$ is supported on $\hbS_\tau$, we have $d^\pi(s,a) = 0$ for $(s,a) \notin \hbS_\tau$. Therefore
\begin{align*}
d^\pi(\bU)
&= \sum_{(s,a) \in \bU \cap \hbS_\tau} d^\pi(s,a)
\;\le\; \Cw \sum_{(s,a) \in \bU \cap \hbS_\tau} \dU(s,a) \\
&= \Cw \cdot \dU(\bU \cap \hbS_\tau)
\;=\; \Cw \cdot \FPR(\tau) \cdot \dU(\bU),
\end{align*}
where the inequality uses Assumption~\ref{ass:bounded} and the last equality is the definition of $\FPR$.
\end{proof}

\begin{applemma}[Structural FPR bound]
\label{lem:fpr-zero}
On the good event $\mathcal{E}$, for any threshold $\tau > \varepsilon$, $\FPR(\tau) = 0$.
\end{applemma}
\begin{proof}
For any $(s,a) \in \bU$, $\rstar(s,a) = 0$ by construction. On $\mathcal{E}$, for every $(s,a) \in \bU$,
\[
\rhat(s,a) = \rhat(s,a) - \rstar(s,a) \;\le\; |\rhat(s,a) - \rstar(s,a)| \;\le\; \varepsilon \;<\; \tau.
\]
Hence $(s,a) \notin \hbS_\tau$ for every $(s,a) \in \bU$, giving $\FPR(\tau) = 0$.
\end{proof}

\rev{
\begin{applemma}[$\kappa$-threshold-FPR connection]
\label{lem:kappa-tau}
Under Assumptions~\ref{ass:calib} and \ref{ass:margin}, if $\kappa > \kappa_0 + \sqrt{\log(2/\delta)/(2n_S)}$, then on the intersection of the good event $\mathcal{E}$ with a calibration event $\mathcal{E}_{\mathrm{cal}}$ of probability at least $1-\delta/2$ over the draw of $\DS$,
\[
\tau_\kappa \;\ge\; \mu - \varepsilon \;>\; \varepsilon.
\]
\end{applemma}
\begin{proof}
On $\mathcal{E}$, any pair $(s,a)$ with $\rhat(s,a) \le \mu - \varepsilon$ satisfies $\rstar(s,a) \le \rhat(s,a) + \varepsilon \le \mu$, hence lies in the fixed boundary set $B := \{(s,a) : \rstar(s,a) \le \mu\}$, whose $\dS$-mass is $\kappa_0$ by definition.

Since $B$ depends only on $\rstar$ and not on $\rhat$, the indicators $\mathds{1}\{(s_i,a_i) \in B\}$ are i.i.d.\ Bernoulli variables with mean $\kappa_0$ under Assumption~\ref{ass:calib}. By Hoeffding's inequality, the event
$\mathcal{E}_{\mathrm{cal}} := \big\{\frac{1}{n_S}\sum_{i=1}^{n_S}\mathds{1}\{(s_i,a_i)\in B\} \le \kappa_0 + \sqrt{\log(2/\delta)/(2n_S)}\big\}$
has probability at least $1 - \delta/2$. On $\mathcal{E} \cap \mathcal{E}_{\mathrm{cal}}$, every sample counted by $\hat F(\mu - \varepsilon)$ lies in $B$ by Step~1, so
\[
\hat F(\mu - \varepsilon) \;\le\; \kappa_0 + \sqrt{\log(2/\delta)/(2n_S)} \;<\; \kappa,
\]
where the last inequality is the assumed strict slack condition on $\kappa$.

By construction, $\hat F(\tau_\kappa) \ge \kappa$. If $\tau_\kappa < \mu - \varepsilon$, monotonicity of $\hat F$ would give $\kappa \le \hat F(\tau_\kappa) \le \hat F(\mu - \varepsilon) < \kappa$, a contradiction. Hence $\tau_\kappa \ge \mu - \varepsilon$, and $\mu > 2\varepsilon$ (Assumption~\ref{ass:margin}) yields $\tau_\kappa > \varepsilon$.
\end{proof}
}

\rev{
\begin{theorem}[Safety guarantee of DexDICE; restated]
\label{thm:guarantee-full}
Suppose Assumptions~\ref{ass:calib}--\ref{ass:margin} hold, with Assumption~\ref{ass:err} at failure level $\delta/2$. Let $\pi_\kappa$ be a support-constrained solution with $\supp(d^{\pi_\kappa}) \subseteq \hbS_{\tau_\kappa}$, where $\tau_\kappa$ is the empirical $\kappa$-quantile of $\rhat$ over $\DS$ (Eq.~\ref{eq:supp_estimator}) and
\[
\kappa \;>\; \kappa_0 + \sqrt{\log(2/\delta)/(2n_S)}.
\]
Then
\[
d^{\pi_\kappa}(\bU) \;=\; 0 \quad \text{with probability at least } 1 - \delta.
\]
\end{theorem}
\begin{proof}
By Assumption~\ref{ass:err}, the good event $\mathcal{E}$ has probability at least $1 - \delta/2$, and by Lemma~\ref{lem:kappa-tau} the calibration event $\mathcal{E}_{\mathrm{cal}}$ has probability at least $1 - \delta/2$; a union bound gives $\Pr(\mathcal{E} \cap \mathcal{E}_{\mathrm{cal}}) \ge 1 - \delta$. On $\mathcal{E} \cap \mathcal{E}_{\mathrm{cal}}$, Lemma~\ref{lem:kappa-tau} gives $\tau_\kappa > \varepsilon$, and Lemma~\ref{lem:fpr-zero} then yields $\FPR(\tau_\kappa) = 0$. Applying Proposition~\ref{prop:visitation} to $\pi_\kappa$, which is supported on $\hbS_{\tau_\kappa}$,
\[
d^{\pi_\kappa}(\bU) \;\le\; \Cw \cdot \FPR(\tau_\kappa) \cdot \dU(\bU) \;=\; 0.
\]
The square-root term in the condition on $\kappa$ is a finite-sample slack; as $n_S \to \infty$, the condition reduces to the strict population condition $\kappa > \kappa_0$.
\end{proof}
}

\begin{appcorollary}[Simultaneous FNR-FPR control]
\label{cor:simul-fnr-fpr}
Under the conditions of Theorem~\ref{thm:guarantee-full}, the FNR parameter $\kappa$ controls both error rates:
\begin{enumerate}[leftmargin=8mm,itemsep=2pt]
\item \textbf{FNR control (direct):} \rev{Under acceptance $\rhat \ge \tau_\kappa$, the rejected set is $\{\rhat < \tau_\kappa\}$, and the empirical FNR on $\DS$, $\frac{1}{n_S}\sum_{(s,a)\in\DS}\mathds{1}\{\rhat(s,a) < \tau_\kappa\}$, is strictly below $\kappa$ by construction; hence $\hbS_{\tau_\kappa}$ retains more than a $(1-\kappa)$-fraction of the calibration pairs in $\DS$.}
\item \textbf{FPR control (indirect):} On $\mathcal{E}$, $\FPR(\tau_\kappa) = 0$ as a consequence of $\tau_\kappa > \varepsilon$ (Lemma~\ref{lem:kappa-tau}) and the structural property $\rstar \equiv 0$ on $\bU$ (Lemma~\ref{lem:fpr-zero}).
\end{enumerate}
\end{appcorollary}

\section{Details on Practical Algorithm}

\subsection{Discriminator-based Estimation}
\label{discriminator_estimation}
Following~\citet{kim2022demodice}, we train discriminators $\hat{C}_\rP, \hat{C}_S : \mathcal{S}\times\mathcal{A} \to [0,1]$ with neural networks to estimate the density ratio $r_\rP = \dP/\dU$ and $r_\rS = \dS/\dU$ via the following objectives:
\begin{align}
    \hat{C}_\rP \in & \argmax_{C: \mathcal{S}\times \mathcal{A} \to [0,1]} \mathbb{E}_{d_\rP}[\log C(s,a)] + \mathbb{E}_{d_\rU}[\log (1-C(s,a))] \label{eq:C_E_obj}\\
    \hat{C}_\rS \in & \argmax_{C: \mathcal{S}\times \mathcal{A} \to [0,1]} \mathbb{E}_{d_\rS}[\log C(s,a)] + \mathbb{E}_{d_\rU}[\log (1-C(s,a))] \label{eq:C_S_obj}
\end{align}
The solutions correspond to ${C}^*_\rP(s,a) = \frac{\dP(s,a)}{\dP(s,a)+\dU(s,a)}, {C^*_\rS}(s,a) = \frac{\dS(s,a)}{\dS(s,a)+\dU(s,a)}$ respectively.
From $\hat{C}_\rP, \hat{C}_\rS$, we estimate the density ratios $\hat{r}_\rP, \hat{r}_\rS$ as follows:
\begin{equation}
\hat{r}_\rP(s,a) = \frac{\dP (s,a)}{\dU(s,a)} = \frac{\hat{C}_\rP(s,a)}{1-\hat{C}_\rP(s,a)}, \quad \hat{r}_\rS(s,a) = \frac{\dS (s,a)}{\dU(s,a)} = \frac{\hat{C}_\rS(s,a)}{1-\hat{C}_\rS(s,a)} \label{eq:obtain_ratios}
\end{equation}

\subsection{Numerical Stability}
To improve numerical stability during optimization arising from summations of exponentials, we adopt an alternative log-sum-exp formulation, as proposed by \citet{kim2022demodice}.

\begin{appproposition}
    (\citet{kim2022demodice}) Define an alternative objective $\tilde{L}(\nu,\lambda)$ as follows:
    \begin{equation}
    \label{eq:practical_obj}
        \tilde{L}(\nu,\lambda) := (1-\gamma) \mathbb{E}_{\rho_0}[\nu(s)] + \log \mathbb{E}_{(s,a)\sim \dU}\left[ \exp\left( A_{\nu,\lambda}(s,a)\right)\right]  -\lambda
    \end{equation}
    Then, the following holds:
    \begin{equation*}
        \min_{\nu} L(\nu,\lambda) = \min_{\nu} \tilde{L}(\nu,\lambda).
    \end{equation*}
\end{appproposition}

\subsection{Policy Extraction}
\label{app.policy_extraction}
In tabular settings, we can directly recover the policy from the optimal stationary distribution $d^*$ as:
\begin{equation*}
    d^*(s,a) = w^*_{\nu,\lambda} (s,a) \dU(s,a) , \quad \pi(a|s) = \frac{d^*(s,a)}{\sum_{\bar{a}\in \mathcal{A}}d^*(s,\bar{a})}.
\end{equation*}
However, in continuous spaces, computing $d^*(s) = \int d^*(s,a) da$ is intractable. Thus, following \citet{lee2021optidice, kim2022demodice}, we extract the policy via weighted behavioral cloning:
\begin{align}
    \min_{\pi} \quad J(\pi;w^*_{\nu,\lambda}) := \mathbb{E}_{(s,a) \sim \dU}\left[ - w^*_{\nu,\lambda}(s,a) \log \pi( a| s)\right].\label{eq:policy_extract}
\end{align}

\subsection{Algorithm Summary}
\label{app.algorithm}
We summarize a practical version of the DexDICE algorithm in \textbf{Algorithm 1}.

\begin{algorithm}[hbt!]
\caption{DexDICE (Safe offline imitation learning from \textbf{D}ual \textbf{ex}pert demonstrations via \textbf{DICE})}
\label{alg:cap}
\iflatexml
\noindent\algorithmicinput{} Safe, performant, unlabeled demonstration sets $\DS, \DP, \mathcal{D}_{\rU}$, and initial-state set $\mathcal{D}_0$\par
\noindent\algorithmicinput{} Discriminator networks $C_{\rP}, C_{\rS}$ with parameters $\zeta_{\rP}, \zeta_{\rS}$, and percentile cutoff $\kappa$ (e.g., $0.01$)\par
\noindent\algorithmicinput{} Value network $\nu_\theta(s)$ and policy network $\pi_\phi$ with parameters $\theta, \phi$, and Lagrange multiplier $\lambda$\par
\begin{algorithmic}[1]
\STATE Initialize $\zeta_{\rP}, \zeta_{\rS}$.
\STATE Pretrain $\hat C_{\rP}$ and $\hat C_{\rS}$ by optimizing objectives~(\ref{eq:C_E_obj}, \ref{eq:C_S_obj}).
\STATE Obtain $\hat r_{\rP}$ and $\hat r_{\rS}$ from Eq.~(\ref{eq:obtain_ratios}).
\STATE Obtain $\hat{\mathds{1}}_{\dS}$ using $\kappa$ from Eq.~(\ref{eq:supp_estimator}).
\STATE Initialize $\theta, \phi, \lambda$.
\FOR{each iteration}
    \STATE Sample mini-batches from $\DP, \DS, \mathcal{D}_{\rU}, \mathcal{D}_0$.
    \STATE Compute gradients for objective~(\ref{eq:practical_obj}):
    \STATE \hspace{1em} $g_\theta \gets \nabla_{\theta}\,\tilde{L}(\nu_\theta,\lambda)$, \quad
                       $g_\lambda \gets \nabla_{\lambda}\,L(\nu_\theta,\lambda)$.
    \STATE Compute gradient for objective~(\ref{eq:policy_extract}):
    \STATE \hspace{1em} $g_\phi \gets \nabla_{\phi}\,J\!\left(\pi_\phi; w^*_{\nu_\theta,\lambda}\right)$.
    \STATE Perform SGD updates:
    \STATE \hspace{1em} $\theta \gets \theta - \eta g_\theta$, \quad
                       $\phi \gets \phi - \eta g_\phi$, \quad
                       $\lambda \gets \lambda - \eta g_\lambda$.
\ENDFOR
\end{algorithmic}
\noindent\algorithmicoutput{} $\nu_\theta \approx \nu^*$, $\pi_\phi \approx \pi^*$\par
\else
\begin{algorithmic}[1]
\INPUT Safe, performant, unlabeled demonstration sets $\DS, \DP, \mathcal{D}_{\rU}$, and initial-state set $\mathcal{D}_0$
\INPUT Discriminator networks $C_{\rP}, C_{\rS}$ with parameters $\zeta_{\rP}, \zeta_{\rS}$, and percentile cutoff $\kappa$ (e.g., $0.01$)
\INPUT Value network $\nu_\theta(s)$ and policy network $\pi_\phi$ with parameters $\theta, \phi$, and Lagrange multiplier $\lambda$

\STATE Initialize $\zeta_{\rP}, \zeta_{\rS}$.
\STATE Pretrain $\hat C_{\rP}$ and $\hat C_{\rS}$ by optimizing objectives~(\ref{eq:C_E_obj}, \ref{eq:C_S_obj}).
\STATE Obtain $\hat r_{\rP}$ and $\hat r_{\rS}$ from Eq.~(\ref{eq:obtain_ratios}).
\STATE Obtain $\hat{\mathds{1}}_{\dS}$ using $\kappa$ from Eq.~(\ref{eq:supp_estimator}).
\STATE Initialize $\theta, \phi, \lambda$.

\FOR{each iteration}
    \STATE Sample mini-batches from $\DP, \DS, \mathcal{D}_{\rU}, \mathcal{D}_0$.
    \STATE Compute gradients for objective~(\ref{eq:practical_obj}):
    \STATE \hspace{1em} $g_\theta \gets \nabla_{\theta}\,\tilde{L}(\nu_\theta,\lambda)$, \quad
                       $g_\lambda \gets \nabla_{\lambda}\,L(\nu_\theta,\lambda)$.
    \STATE Compute gradient for objective~(\ref{eq:policy_extract}):
    \STATE \hspace{1em} $g_\phi \gets \nabla_{\phi}\,J\!\left(\pi_\phi; w^*_{\nu_\theta,\lambda}\right)$.
    \STATE Perform SGD updates:
    \STATE \hspace{1em} $\theta \gets \theta - \eta g_\theta$, \quad
                       $\phi \gets \phi - \eta g_\phi$, \quad
                       $\lambda \gets \lambda - \eta g_\lambda$.
\ENDFOR

\OUTPUT $\nu_\theta \approx \nu^*$, $\pi_\phi \approx \pi^*$
\end{algorithmic}
\fi
\end{algorithm}

\section{Details on Simulation Experiments}
\label{sec:datasetdescription}
\update{This section details the simulation experiments on 38 re-curated DSRL tasks and six controlled Safety Gymnasium tasks. It covers dataset construction (Sections~\ref{app:DSRLDatasets} and \ref{sec:controlled_dataset}), extended baseline comparisons, hyperparameters, data efficiency and quality, support-estimator comparisons, sensitivity to the target FNR and dataset construction thresholds, robustness to noisy safe demonstrations, and deployment-time projection.}

\subsection{Re-curated DSRL Datasets}
\label{app:DSRLDatasets}

The original DSRL datasets~\citep{liu2024datasets} were constructed by aggregating demonstrations from policies trained with a range of safe RL algorithms, resulting in a diverse set of pre-collected trajectories. Each trajectory records both its episode return and cost, which allows us to partition the dataset into dual-expert subsets through thresholding.
To define these subsets, we adopted cost thresholds from the original DSRL paper (specifically, one of $\{20, 40, 80\}$) and classified trajectories as \emph{safe} if their cumulative cost was below the corresponding threshold, or \emph{unsafe} otherwise. For the return threshold, we selected the 90th percentile of returns among the safe trajectories. \rev{Sensitivity to both thresholds is analyzed in Appendix~\ref{sec:app.threshold_sensitivity}.}
With these criteria, each trajectory is categorized into one of four groups: \texttt{performant} ($\DP$), \texttt{safe} ($\DS$), \texttt{non-preferred} ($\mathcal{D}_{\mathrm{N}}$), or \texttt{unlabeled} ($\DU$), based on its safety and performance relative to the thresholds. The \texttt{non-preferred} set is exclusively used in the SafeDICE experiments.
Table~\ref{table:dsrldescription} summarizes the characteristics and statistics for each task category in the re-curated DSRL benchmark.
For brevity, task names in the tables and analyses are abbreviated where appropriate by omitting repeated prefixes (e.g., `Safety', `Safe') and environment-version suffixes (e.g., `-v0', `-v1').

\vspace{10pt}
\begin{table}[hpt!]
\centering
\begin{adjustbox}{max width=1\textwidth}
    \small
    \centering
    \aboverulesep=0.1ex
    \belowrulesep=0.1ex
    \renewcommand{\arraystretch}{1.1}
\begin{tabular}
{l|
S[table-format=3.0,mode=text]
S[table-format=2.0,mode=text]
S[table-format=2.0,mode=text]
S[table-format=4.2,mode=text]
S[table-format=4.2,mode=text]
S[table-format=3.2,mode=text]}
\toprule
\textbf{Task} & \textbf{Obs. Dim} & \textbf{Act. Dim} & \textbf{Cost Threshold} & \textbf{Return Threshold} & \textbf{Max Return} &\textbf{Min Return} \\
\midrule
\midrule
\multicolumn{7}{l}{\textbf{Safety Gymnasium}} \\
\midrule
PointCircle1 & 28 & 2 & 40 & 44.85 & 61.73 & 20.07 \\
\midrule
PointGoal1 & 60 & 2 & 40 & 21.67 & 30.07 & 0.01 \\
\midrule
PointButton1 & 76 & 2 & 40 & 27.12  & 41.19 & 0.01\\
\midrule
PointPush1 & 76 & 2 & 40 & 9.91 & 16.52 & 0.001 \\
\midrule
PointCircle2 & 28 & 2 & 80 & 43.10 & 54.02 & 20.48 \\
\midrule
PointGoal2 & 60 & 2 & 80 & 20.40 & 27.72 & 0.003 \\
\midrule
PointButton2 & 76 & 2 & 80 & 29.75 & 42.90 & 0.01 \\
\midrule
PointPush2 & 76 & 2 & 40 & 8.15 & 14.69 & 0.001 \\
\midrule
CarCircle1 & 40 & 2 & 80 & 17.47 & 24.94 & 8.04 \\
\midrule
CarGoal1 & 72 & 2 & 40 & 30.08 & 39.91 & 0.01 \\
\midrule
CarButton1 & 88 & 2 & 80 & 25.15 & 44.42 & 0.003 \\
\midrule
CarPush1 & 88 & 2 & 40 & 8.24 & 16.30 & 0.01 \\
\midrule
CarCircle2 & 40 & 2 & 80 & 18.14 & 25.95 & 5.11 \\
\midrule
CarGoal2 & 72 & 2 & 40 & 18.89 & 28.90 & 0.001 \\
\midrule
CarButton2 & 88 & 2 & 80 & 23.46 & 42.00 & 0.002 \\
\midrule
CarPush2 & 88 & 2 & 40 & 6.46 & 15.14 & 0.00 \\
\midrule
HopperVelocity & 11 & 3 & 40 & 1607.84 & 1911.40 & 37.05  \\
\midrule
Walker2dVelocity & 17 & 6 & 40 & 2718.04 & 3418.22 & 18.67  \\
\midrule
HalfCheetahVelocity & 17 & 6 & 40 & 2563.55 & 2806.93 & 5.75  \\
\midrule
AntVelocity & 27 & 8 & 40 & 2743.78 & 2976.28 & 6.15 \\
\midrule
SwimmerVelocity & 8 & 2 & 40 & 158.47 & 238.96 & 0.07  \\
\midrule
\midrule
\multicolumn{7}{l}{\textbf{Bullet Safety Gym}} \\
\midrule
BallRun & 8 & 2 & 40  & 556.19 & 1327.45 & 26.34 \\
\midrule
BallCircle & 7 & 2 & 40 & 712.45 & 881.46 & 0.38 \\
\midrule
CarRun & 8 & 2 & 20 & 566.21 & 574.65 & 204.29 \\
\midrule
CarCircle & 7 & 2 & 40 & 435.21 & 534.31 & 3.48 \\
\midrule
DroneRun & 18 & 4 & 40 & 411.47 & 682.83 & 10.56 \\
\midrule
DroneCircle & 17 & 4 & 40 & 713.46 & 996.39 & 207.79 \\
\midrule
AntRun & 34 & 8 & 40 & 664.83 & 955.48 & 0.00 \\
\midrule
AntCircle & 33 & 8 & 40 & 225.71 & 460.71 & 0.02 \\
\midrule
\midrule
\multicolumn{7}{l}{\textbf{MetaDrive}} \\
\midrule
MetaDrive-easydense & 259 & 2 & 40 & 325.86 & 425.82 & 19.71 \\
\midrule
MetaDrive-easymean & 259 & 2 & 40 & 339.41 & 425.82 & 22.14 \\
\midrule
MetaDrive-easysparse & 259 & 2 & 40 & 327.70 & 425.80 & 20.09 \\
\midrule
MetaDrive-mediumdense & 259 & 2 & 20 & 263.84 & 269.79 & 15.10  \\
\midrule
MetaDrive-mediummean & 259 & 2 & 20 & 262.39 & 269.82 & 17.03 \\
\midrule
MetaDrive-mediumsparse & 259 & 2 & 20 & 263.81 & 268.80  & 17.45  \\
\midrule
MetaDrive-harddense & 259 & 2 & 20 & 283.47 & 486.64  & 15.39  \\
\midrule
MetaDrive-hardmean & 259 & 2 & 20 & 309.68 & 486.81 & 18.11 \\
\midrule
MetaDrive-hardsparse & 259 & 2 & 20 & 311.44 & 486.81 & 17.25 \\
\bottomrule
\end{tabular}
\end{adjustbox}
\vspace{5pt}
\caption{Descriptions of Tasks in Re-curated DSRL Datasets.}
\label{table:dsrldescription}
\end{table}

\newpage

We detail the data labeling process for each demonstration category below:

\begin{itemize}
    [leftmargin=6mm]
    \item \update{\texttt{safe}: demonstrations whose cumulative cost remains below the predefined cost threshold, regardless of their return.}
    \item \update{\texttt{performant}: demonstrations whose return exceeds the predefined return threshold, regardless of their cost.}
    \item \update{\texttt{non-preferred}: demonstrations whose cumulative cost exceeds the cost threshold (i.e., unsafe trajectories).}
    \item \update{\texttt{unlabeled}: the full set of offline trajectories.}
\end{itemize}
Tables~\ref{table:SafetyGym_statistics-1},
\ref{table:SafetyGym_statistics-2}, \ref{table:Bullet_statistics}, and \ref{table:MetaDrive_statistics} summarize the statistics of the re-curated DSRL datasets across 38 tasks with three different environments, reporting the safe ratio, performant ratio, safe-performant ratio, and the total number of trajectories for each demonstration category.

\begin{table}[hpt!]
\centering
\begin{adjustbox}{max width=0.85\textwidth}
    \small
    \centering
    \aboverulesep=0.1ex
    \belowrulesep=0.1ex
    \renewcommand{\arraystretch}{1.1}
\begin{tabular}
{l|l|
S[table-format=1.2,mode=text]
S[table-format=1.2,mode=text]
S[table-format=1.2,mode=text]
S[table-format=4,mode=text]}
\toprule
\textbf{Task} & \textbf{Dataset} & \textbf{Safe Ratio} & \textbf{Performant Ratio} & \textbf{Safe-Performant Ratio} & \textbf{\# traj.} \\
\midrule
\midrule
\multicolumn{6}{l}{\textbf{Safety Gymnasium}} \\
\midrule
\multirow{4}{*}{\shortstack[l]{PointCircle1}}
 & \texttt{safe}           & 1.0  & 0.10 & 0.10  & 214  \\
 & \texttt{performant}     & 0.03 & 1.0  & 0.03  & 793 \\
 & \texttt{non-preferred}   & 0.0 & 0.87 & 0.0 & 884 \\
 & \texttt{unlabeled}      & 0.19 & 0.72 & 0.02 & 1098 \\
\midrule
\multirow{4}{*}{\shortstack[l]{PointGoal1}}
 & \texttt{safe} & 1.0 & 0.10 & 0.10 & 1364 \\
 & \texttt{performant} & 0.44 & 1.0 & 0.44 & 313  \\
 & \texttt{non-preferred} & 0.0 & 0.27 & 0.0 & 658 \\
 & \texttt{unlabeled} & 0.67 & 0.15 & 0.07 & 2022 \\
\midrule
\multirow{4}{*}{\shortstack[l]{PointButton1}}
 & \texttt{safe} & 1.0 & 0.10 & 0.10 & 783 \\
 & \texttt{performant} & 0.24 & 1.0 & 0.24 & 327 \\
 & \texttt{non-preferred} & 0.0 & 0.17 & 0.0 & 1485 \\
 & \texttt{unlabeled} & 0.35 & 0.14 & 0.03 & 2268 \\
\midrule
\multirow{4}{*}{\shortstack[l]{PointPush1}}
 & \texttt{safe} & 1.0 & 0.10 & 0.10 & 1095 \\
 & \texttt{performant} & 0.47 & 1.0 & 0.47 & 235 \\
 & \texttt{non-preferred} & 0.0 & 0.10 & 0.0 & 1284 \\
 & \texttt{unlabeled} & 0.46 & 0.10 & 0.05 & 2379 \\
\midrule
\multirow{4}{*}{\shortstack[l]{PointCircle2}}
 & \texttt{safe} & 1.0 & 0.10 & 0.10 & 349 \\
 & \texttt{performant} & 0.09 & 1.0 & 0.09 & 409 \\
 & \texttt{non-preferred} & 0.0 & 0.68 & 0.0 & 546 \\
 & \texttt{unlabeled} & 0.39 & 0.46 & 0.04 & 895 \\
\midrule
\multirow{4}{*}{\shortstack[l]{PointGoal2}}
 & \texttt{safe} & 1.0 & 0.10 & 0.10 & 1963 \\
 & \texttt{performant} & 0.38 & 1.0 & 0.38 & 525 \\
 & \texttt{non-preferred} & 0.0 & 0.22 & 0.0 & 1479 \\
 & \texttt{unlabeled} & 0.57 & 0.15 & 0.06 & 3442 \\
\midrule
\multirow{4}{*}{\shortstack[l]{PointButton2}}
 & \texttt{safe} & 1.0 & 0.10 & 0.10 & 1513 \\
 & \texttt{performant} & 0.39 & 1.0 & 0.39 & 386 \\
 & \texttt{non-preferred} & 0.0 & 0.13 & 0.0 & 1775 \\
 & \texttt{unlabeled} & 0.46 & 0.12 & 0.05 & 3288 \\
\midrule
\multirow{4}{*}{\shortstack[l]{PointPush2}}
 & \texttt{safe} & 1.0 & 0.10 & 0.10 & 1218 \\
 & \texttt{performant} & 0.47 & 1.0 & 0.47 & 260 \\
 & \texttt{non-preferred} & 0.0 & 0.07 & 0.0 & 2024  \\
 & \texttt{unlabeled} & 0.38 & 0.08 & 0.04 & 3242 \\
\bottomrule
\end{tabular}
\end{adjustbox}
\vspace{5pt}
\caption{Statistics for 4 types of datasets for Point tasks of Safety Gymnasium environment in DSRL Datasets.}
\label{table:SafetyGym_statistics-1}
\end{table}

\begin{table}[t!]
\centering
\begin{adjustbox}{max width=0.85\textwidth}
    \small
    \centering
    \aboverulesep=0.1ex
    \belowrulesep=0.1ex
    \renewcommand{\arraystretch}{1.1}
\begin{tabular}
{l|l|
S[table-format=1.2,mode=text]
S[table-format=1.2,mode=text]
S[table-format=1.2,mode=text]
S[table-format=4,mode=text]}
\toprule
\textbf{Task} & \textbf{Dataset} & \textbf{Safe Ratio} & \textbf{Performant Ratio} & \textbf{Safe-Performant Ratio} & \textbf{\# traj.} \\
\midrule
\midrule
\multicolumn{6}{l}{\textbf{Safety Gymnasium}} \\
\midrule
\multirow{4}{*}{\shortstack[l]{CarCircle1}}
 & \texttt{safe} & 1.0 & 0.10 & 0.10 & 313 \\
 & \texttt{performant} & 0.04 & 1.0 & 0.04 & 908 \\
 & \texttt{non-preferred} & 0.0 & 0.91 & 0.0 & 958 \\
 & \texttt{unlabeled} & 0.25 & 0.71 & 0.03 & 1271 \\
\midrule
\multirow{4}{*}{\shortstack[l]{CarGoal1}}
 & \texttt{safe} & 1.0 & 0.10 & 0.10 & 905  \\
 & \texttt{performant} & 0.47 & 1.0 & 0.47 & 193 \\
 & \texttt{non-preferred} & 0.0 & 0.13 & 0.0 & 766  \\
 & \texttt{unlabeled} & 0.54 & 0.12 & 0.05 &  1671 \\
\midrule
\multirow{4}{*}{\shortstack[l]{CarButton1}}
 & \texttt{safe} & 1.0 & 0.10 & 0.10 & 1094 \\
 & \texttt{performant} & 0.30 & 1.0 & 0.30 & 368 \\
 & \texttt{non-preferred} & 0.0 & 0.17 & 0.0 & 1562 \\
 & \texttt{unlabeled} & 0.41 & 0.14 & 0.04 & 2656 \\
\midrule
\multirow{4}{*}{\shortstack[l]{CarPush1}}
 & \texttt{safe} & 1.0 & 0.10 & 0.10 & 1459 \\
 & \texttt{performant} & 0.37 & 1.0 & 0.37 & 399 \\
 & \texttt{non-preferred} & 0.0 & 0.18 & 0.0 & 1412  \\
 & \texttt{unlabeled} & 0.51 & 0.14 & 0.05 & 2871 \\
\midrule
\multirow{4}{*}{\shortstack[l]{CarCircle2}}
 & \texttt{safe} & 1.0 & 0.10 & 0.10 & 266 \\
 & \texttt{performant} & 0.06 & 1.0 & 0.06 & 435 \\
 & \texttt{non-preferred} & 0.0 & 0.61 & 0.0 & 674 \\
 & \texttt{unlabeled} & 0.28 & 0.46 & 0.03 & 940 \\
\midrule
\multirow{4}{*}{\shortstack[l]{CarGoal2}}
 & \texttt{safe} & 1.0 & 0.10 & 0.10 & 1709 \\
 & \texttt{performant} & 0.22 & 1.0 & 0.22 & 788 \\
 & \texttt{non-preferred} & 0.0 & 0.26 & 0.0 & 2396 \\
 & \texttt{unlabeled} & 0.42 & 0.19 & 0.04 & 4105 \\
\midrule
\multirow{4}{*}{\shortstack[l]{CarButton2}}
 & \texttt{safe} & 1.0 & 0.10 & 0.10 & 1468 \\
 & \texttt{performant} & 0.20 & 1.0 & 0.20 & 745 \\
 & \texttt{non-preferred} & 0.0 & 0.26 & 0.0 & 2287  \\
 & \texttt{unlabeled} & 0.39 & 0.20 & 0.04 & 3755 \\
\midrule
\multirow{4}{*}{\shortstack[l]{CarPush2}}
 & \texttt{safe} & 1.0 & 0.10 & 0.10 & 2004 \\
 & \texttt{performant} & 0.34 & 1.0 & 0.34 & 588 \\
 & \texttt{non-preferred} & 0.0 & 0.16  & 0.0 & 2403 \\
 & \texttt{unlabeled} & 0.45 & 0.13 & 0.05 & 4407 \\
 \midrule
\multirow{4}{*}{\shortstack[l]{HopperVelocity}}
 & \texttt{safe} & 1.0 & 0.10 & 0.10 & 461 \\
 & \texttt{performant} & 0.09 & 1.0 & 0.09 & 501 \\
 & \texttt{non-preferred} & 0.0 & 0.26 & 0.0 & 1779 \\
 & \texttt{unlabeled} & 0.21 & 0.22 & 0.02 & 2240 \\
\midrule
\multirow{4}{*}{\shortstack[l]{Walker2dVelocity}}
 & \texttt{safe} & 1.0 & 0.10 & 0.10 & 1020 \\
 & \texttt{performant} & 0.17 & 1.0 & 0.17 & 584 \\
 & \texttt{non-preferred} & 0.0 & 0.28 & 0.0 & 1709 \\
 & \texttt{unlabeled} & 0.37 & 0.21 & 0.04 & 2729 \\
\midrule
\multirow{4}{*}{\shortstack[l]{HalfCheetahVelocity}}
 & \texttt{safe} & 1.0 & 0.10 & 0.10 & 664 \\
 & \texttt{performant} & 0.13 & 1.0 & 0.13 & 505 \\
 & \texttt{non-preferred} & 0.0 & 0.24 & 0.0 & 1831 \\
 & \texttt{unlabeled} &  0.27 & 0.20 & 0.03 & 2495 \\
\midrule
\multirow{4}{*}{\shortstack[l]{AntVelocity}}
 & \texttt{safe} & 1.0 & 0.10 & 0.10 & 802 \\
 & \texttt{performant} & 0.10 & 1.0 & 0.10 & 782 \\
 & \texttt{non-preferred} & 0.0 & 0.48 & 0.0 & 1447 \\
 & \texttt{unlabeled} & 0.36 & 0.35 & 0.04 & 2249 \\
 \midrule
\multirow{4}{*}{\shortstack[l]{SwimmerVelocity}}
 & \texttt{safe} & 1.0 & 0.10 & 0.10 & 422 \\
 & \texttt{performant} & 0.15 & 1.0 & 0.15 & 290 \\
 & \texttt{non-preferred} & 0.0 & 0.20 & 0.0 & 1264  \\
 & \texttt{unlabeled} & 0.25 & 0.17 & 0.03 & 1686 \\
\bottomrule
\end{tabular}
\end{adjustbox}
\vspace{5pt}
\caption{Statistics for 4 types of datasets for remaining tasks of Safety Gymnasium environment in DSRL Datasets.}
\label{table:SafetyGym_statistics-2}
\end{table}

\begin{table}[t!]
\centering
\begin{adjustbox}{max width=1\textwidth}
    \small
    \centering
    \aboverulesep=0.1ex
    \belowrulesep=0.1ex
    \renewcommand{\arraystretch}{1.1}
\begin{tabular}
{l|l|
S[table-format=1.2,mode=text]
S[table-format=1.2,mode=text]
S[table-format=1.2,mode=text]
S[table-format=4,mode=text]}
\toprule
\textbf{Task} & \textbf{Dataset} & \textbf{Safe Ratio} & \textbf{Performant Ratio} & \textbf{Safe-Performant Ratio} & \textbf{\# traj.} \\
\midrule
\midrule
\multicolumn{6}{l}{\textbf{Bullet Safety Gym}} \\
\midrule
\multirow{4}{*}{\shortstack[l]{BallRun}}
 & \texttt{safe} & 1.0 & 0.10 & 0.10 & 315 \\
 & \texttt{performant} & 0.07 & 1.0 & 0.07 & 488 \\
 & \texttt{non-preferred} & 0.0 & 0.73 & 0.0 & 625 \\
 & \texttt{unlabeled} &  0.34 & 0.52 & 0.03 & 940 \\
\midrule
\multirow{4}{*}{\shortstack[l]{BallCircle}}
 & \texttt{safe} & 1.0 & 0.10 & 0.10 & 385 \\
 & \texttt{performant} & 0.12 & 1.0 & 0.12 & 338 \\
 & \texttt{non-preferred} & 0.0 & 0.60 & 0.0 & 501 \\
 & \texttt{unlabeled} & 0.43 & 0.38 & 0.04 & 886 \\
\midrule
\multirow{4}{*}{\shortstack[l]{CarRun}}
 & \texttt{safe} & 1.0 & 0.10 & 0.10 & 470 \\
 & \texttt{performant} & 0.52 & 1.0 & 0.52 & 91 \\
 & \texttt{non-preferred} & 0.0 & 0.24 & 0.0 & 181 \\
 & \texttt{unlabeled} & 0.72 & 0.14 & 0.07 & 651 \\
\midrule
\multirow{4}{*}{\shortstack[l]{CarCircle}}
 & \texttt{safe} & 1.0 & 0.10 & 0.10 & 634 \\
 & \texttt{performant} & 0.10 & 1.0 & 0.10 & 633 \\
 & \texttt{non-preferred} & 0.0 & 0.70 & 0.0 & 816 \\
 & \texttt{unlabeled} & 0.44 & 0.44 & 0.04 & 1450 \\
\midrule
\multirow{4}{*}{\shortstack[l]{DroneRun}}
 & \texttt{safe} & 1.0 & 0.10 & 0.10 & 933 \\
 & \texttt{performant} & 0.14 & 1.0 & 0.14 & 682 \\
 & \texttt{non-preferred} & 0.0 & 0.56 & 0.0 & 1057 \\
 & \texttt{unlabeled} & 0.47 & 0.34 & 0.05 & 1990 \\
\midrule
\multirow{4}{*}{\shortstack[l]{DroneCircle}}
 & \texttt{safe} & 1.0 & 0.10 & 0.10 & 728 \\
 & \texttt{performant} & 0.06 & 1.0 & 0.06 & 1219 \\
 & \texttt{non-preferred} & 0.0 & 0.96 & 0.0 & 1195 \\
 & \texttt{unlabeled} & 0.38 & 0.63 & 0.04 & 1923 \\
\midrule
\multirow{4}{*}{\shortstack[l]{AntRun}}
 & \texttt{safe} & 1.0 & 0.10 & 0.10 & 654 \\
 & \texttt{performant} & 0.09 & 1.0 & 0.09 & 727 \\
 & \texttt{non-preferred} & 0.0 & 0.57 & 0.0 & 1162 \\
 & \texttt{unlabeled} & 0.36 & 0.40 & 0.04 & 1816 \\
 \midrule
 \multirow{4}{*}{\shortstack[l]{AntCircle}}
 & \texttt{safe} & 1.0 & 0.10 & 0.10 & 1611 \\
 & \texttt{performant} & 0.05 & 1.0 & 0.05 & 3244 \\
 & \texttt{non-preferred} & 0.0 & 0.75 & 0.0 & 4117 \\
 & \texttt{unlabeled} & 0.28 & 0.57 & 0.03 & 5728 \\
\bottomrule
\end{tabular}
\end{adjustbox}
\vspace{5pt}
\caption{Statistics for 4 types of datasets for Bullet Safety Gym environment in DSRL Datasets.}
\label{table:Bullet_statistics}
\end{table}

\clearpage

\begin{table}[t!]
\centering
\begin{adjustbox}{max width=1\textwidth}
    \small
    \centering
    \aboverulesep=0.1ex
    \belowrulesep=0.1ex
\renewcommand{\arraystretch}{1.1}
\begin{tabular}
{l|l|
S[table-format=1.2,mode=text]
S[table-format=1.2,mode=text]
S[table-format=1.2,mode=text]
S[table-format=4,mode=text]}
\toprule
\textbf{Task} & \textbf{Dataset} & \textbf{Safe Ratio} & \textbf{Performant Ratio} & \textbf{Safe-Performant Ratio} & \textbf{\# traj.} \\
\midrule
\midrule
\multicolumn{6}{l}{\textbf{MetaDrive}} \\
\midrule
\multirow{4}{*}{\shortstack[l]{easydense}}
 & \texttt{safe} & 1.0 & 0.10 & 0.10 & 675 \\
 & \texttt{performant} & 0.21 & 1.0 & 0.21 & 331  \\
 & \texttt{non-preferred} & 0.0 & 0.81 & 0.0 & 325 \\
 & \texttt{unlabeled} & 0.68 & 0.33 & 0.07 & 1000 \\
\midrule
\multirow{4}{*}{\shortstack[l]{easymean}}
 & \texttt{safe} & 1.0 & 0.10 & 0.10 & 634 \\
 & \texttt{performant} & 0.16 & 1.0 & 0.16 & 388 \\
 & \texttt{non-preferred} & 0.0 & 0.89 & 0.0 & 366 \\
 & \texttt{unlabeled} & 0.63 & 0.39 & 0.06 & 1000 \\
 \midrule
\multirow{4}{*}{\shortstack[l]{easysparse}}
 & \texttt{safe} & 1.0 & 0.10 & 0.10 & 626 \\
 & \texttt{performant} & 0.16 & 1.0 & 0.16 & 388 \\
 & \texttt{non-preferred} & 0.0 & 0.87 & 0.0 & 374 \\
 & \texttt{unlabeled} & 0.63 & 0.39 & 0.06 & 1000 \\
 \midrule
\multirow{4}{*}{\shortstack[l]{mediumdense}}
 & \texttt{safe} & 1.0 & 0.10 & 0.10 & 576 \\
 & \texttt{performant} & 0.25 & 1.0 & 0.25 & 230 \\
 & \texttt{non-preferred} & 0.0 & 0.41 & 0.0 & 424 \\
 & \texttt{unlabeled} & 0.58 & 0.23 & 0.06 & 1000 \\
 \midrule
\multirow{4}{*}{\shortstack[l]{mediummean}}
 & \texttt{safe} & 1.0 & 0.10 & 0.10 & 870 \\
 & \texttt{performant} & 0.52 & 1.0 & 0.52 & 166 \\
 & \texttt{non-preferred} & 0.0 & 0.61 & 0.0 & 130 \\
 & \texttt{unlabeled} & 0.87 & 0.17 & 0.09 & 1000 \\
 \midrule
\multirow{4}{*}{\shortstack[l]{mediumsparse}}
 & \texttt{safe} & 1.0 & 0.10 & 0.10 & 601 \\
 & \texttt{performant} & 0.34 & 1.0 & 0.34 & 178 \\
 & \texttt{non-preferred} & 0.0 & 0.29 & 0.0 & 399 \\
 & \texttt{unlabeled} & 0.60 & 0.18 & 0.06 & 1000 \\
 \midrule
\multirow{4}{*}{\shortstack[l]{harddense}}
 & \texttt{safe} & 1.0 & 0.10 & 0.10 & 582 \\
 & \texttt{performant} & 0.23 & 1.0 & 0.23 & 260  \\
 & \texttt{non-preferred} & 0.0 & 0.48 & 0.0 & 418 \\
 & \texttt{unlabeled} & 0.58 & 0.26 & 0.06 & 1000 \\
 \midrule
\multirow{4}{*}{\shortstack[l]{hardmean}}
 & \texttt{safe} & 1.0 & 0.10 & 0.10 & 527 \\
 & \texttt{performant} & 0.22 & 1.0 & 0.22 & 236  \\
 & \texttt{non-preferred} & 0.0 & 0.39 & 0.0 & 473 \\
 & \texttt{unlabeled} & 0.53 & 0.24 & 0.05 & 1000 \\
 \midrule
\multirow{4}{*}{\shortstack[l]{hardsparse}}
  & \texttt{safe} & 1.0 & 0.10 & 0.10 & 523 \\
 & \texttt{performant} & 0.18 & 1.0 & 0.18 & 287 \\
 & \texttt{non-preferred} & 0.0 & 0.49 & 0.0 & 477 \\
 & \texttt{unlabeled} & 0.52 & 0.29 & 0.05 & 1000 \\
\bottomrule
\end{tabular}
\end{adjustbox}
\vspace{5pt}
\caption{Statistics for 4 types of datasets for MetaDrive environment in DSRL Datasets.}
\vspace{-5pt}
\label{table:MetaDrive_statistics}
\end{table}
\subsection{Controlled Datasets}
\label{sec:controlled_dataset}

To systematically study the influence of (1) the number of trajectories and (2) the quality of datasets, we construct controlled datasets with varying levels of safety and performance\rev{; the Safety Gymnasium tasks used for this purpose are described in Table~\ref{table:safetygymdescription}}. This involves collecting demonstrations from policies exhibiting different performance (e.g., near-optimal, medium) and safety (e.g., safe, unsafe) characteristics. We utilize the OmniSafe framework~\citep{ji2024omnisafe}, which provides a suite of safe RL algorithms for the Safety Gymnasium benchmark~\citep{ji2023safety}, to generate policies for data collection.

Specifically, we employ PPO~\citep{schulman2017proximal} for unconstrained MDPs and PPO-Lag~\citep{ray2019benchmarking} for constrained MDPs (CMDPs) with explicit cost limits. Using these algorithms, we train near-optimal policies for both settings: PPO-Lag yields the \texttt{safe-high} dataset (cost-constrained, high return) and PPO yields the \texttt{unsafe-high} dataset (unconstrained, high return).

To acquire the \texttt{safe-medium} and \texttt{unsafe-medium} datasets, we identify policies whose performance is approximately half that of the corresponding near-optimal policy. We ensure that the \texttt{safe-medium} policies continue to satisfy cost constraints, whereas the \texttt{unsafe-medium} policies violate these constraints by design.

For the data efficiency (\textbf{Q2}) and data quality (\textbf{Q3}) experiments (see Section~\ref{sec:dataefficiency}), we conduct evaluations on six representative tasks: SafetyPointCircle1, SafetyPointCircle2, SafetyHopperVelocity, SafetyHalfCheetahVelocity, SafetyWalker2dVelocity, and SafetyAntVelocity.

\begin{table}[thb!]
\centering
    \small
    \centering
    \aboverulesep=0.1ex
    \belowrulesep=0.1ex
    \renewcommand{\arraystretch}{1.1}
\begin{tabular}
{l|
S[table-format=3.0,mode=text]
S[table-format=2.0,mode=text]
S[table-format=2.0,mode=text]}
\toprule
\textbf{Task} & \textbf{Obs. Dim} & \textbf{Act. Dim} & \textbf{Cost limit} \\
\midrule
\midrule
\multicolumn{4}{l}{\textbf{Safe Navigation Tasks}} \\
\midrule
SafetyPointCircle1 & 28 & 2 & 25 \\
\midrule
SafetyPointCircle2 & 28 & 2 & 25 \\
\midrule
\midrule
\multicolumn{4}{l}{\textbf{Safe Velocity Tasks}} \\
\midrule
SafetyHopperVelocity & 11 & 3 & 25 \\
\midrule
SafetyWalker2dVelocity & 17 & 6 & 25 \\
\midrule
SafetyHalfCheetahVelocity & 17 & 6 & 25 \\
\midrule
SafetyAntVelocity & 27 & 8 & 25 \\
\bottomrule
\end{tabular}
\vspace{5pt}
\caption{Descriptions of Tasks in Safety Gymnasium.}
\label{table:safetygymdescription}
\end{table}

\begin{table*}[p]
  \centering
  \begin{adjustbox}{width=\textwidth, totalheight=0.92\textheight, keepaspectratio}
  \aboverulesep=0.1ex
  \belowrulesep=0.1ex
  \renewcommand{\arraystretch}{1.1}
  \begin{tabular}
  {l|l|
   S[table-format=1.2(1.2),mode=text]
   S[table-format=1.2(1.2),mode=text]
   S[table-format=1.2(1.2),mode=text]
   S[table-format=1.2(1.2),mode=text]
   S[table-format=1.2(1.2),mode=text]
   S[table-format=1.2(1.2),mode=text]
   S[table-format=1.2(1.2),mode=text]
   S[table-format=1.2(1.2),mode=text]
   S[table-format=1.2(1.2),mode=text]
   S[table-format=1.2(1.2),mode=text]
   |S[table-format=1.2(1.2),mode=text]
   }
     \toprule
  \textbf{Environment} & \textbf{Metric} & \textbf{BC} & \textbf{FBC} & \textbf{JFBC} & \textbf{DemoDICE-S} & \textbf{DemoDICE-U} & \textbf{DWBC-S} & \textbf{FISOR-IL ($\hat{c}$)} & \textbf{CDT ($\hat{r}, \hat{c}$)} & \textbf{TraC ($\hat{r}, \hat{c}$)} & \textbf{DexDICE} & \textbf{DexDICE (best $\kappa$)} \\
  \hline\hline
\multirow{2}{*}{PointButton1} & Episode return & 0.49 \pm 0.01 & 0.05 \pm 0.04 & 0.59 \pm 0.01 & \bfseries 0.05 \pm 0.02 & \color{blue}\bfseries 0.11 \pm 0.02 & 0.48 \pm 0.01 & 0.57 \pm 0.01 & 0.40 \pm 0.10 & 0.10 \pm 0.13 & \color{blue}\bfseries 0.15 \pm 0.02 & \bfseries 0.15 \pm 0.02 \\
 & Episode cost & 2.93 \pm 0.09 & 1.81 \pm 0.28 & 3.23 \pm 0.10 & \bfseries 0.51 \pm 0.07 & \color{blue}\bfseries 0.95 \pm 0.14 & 3.02 \pm 0.10 & 3.10 \pm 0.11 & 2.23 \pm 0.38 & 1.81 \pm 1.31 & \color{blue}\bfseries 0.95 \pm 0.09 & \bfseries 0.95 \pm 0.09 \\
\cmidrule(lr){1-13}
\multirow{2}{*}{PointButton2} & Episode return & 0.41 \pm 0.03 & 0.17 \pm 0.03 & 0.56 \pm 0.01 & \bfseries 0.21 \pm 0.02 & \bfseries 0.22 \pm 0.02 & 0.42 \pm 0.02 & 0.56 \pm 0.01 & 0.37 \pm 0.09 & 0.23 \pm 0.14 & \color{blue}\bfseries 0.27 \pm 0.02 & \bfseries 0.27 \pm 0.02 \\
 & Episode cost & 1.58 \pm 0.04 & 1.90 \pm 0.09 & 1.70 \pm 0.06 & \bfseries 0.69 \pm 0.06 & \bfseries 0.73 \pm 0.06 & 1.60 \pm 0.08 & 1.82 \pm 0.02 & 1.14 \pm 0.25 & 1.33 \pm 0.27 & \color{blue}\bfseries 0.92 \pm 0.08 & \bfseries 0.92 \pm 0.08 \\
\cmidrule(lr){1-13}
\multirow{2}{*}{PointCircle1} & Episode return & 0.75 \pm 0.01 & -0.27 \pm 0.15 & 0.87 \pm 0.01 & \color{blue}\bfseries 0.29 \pm 0.09 & 0.81 \pm 0.01 & 0.73 \pm 0.04 & 0.84 \pm 0.01 & 0.30 \pm 0.08 & \bfseries 0.09 \pm 0.02 & \color{blue}\bfseries 0.27 \pm 0.03 & \bfseries 0.27 \pm 0.03 \\
 & Episode cost & 4.37 \pm 0.15 & 5.10 \pm 0.84 & 4.73 \pm 0.06 & \color{blue}\bfseries 0.93 \pm 0.38 & 4.74 \pm 0.20 & 3.49 \pm 0.46 & 4.46 \pm 0.03 & 1.41 \pm 0.45 & \bfseries 0.01 \pm 0.03 & \color{blue}\bfseries 0.27 \pm 0.07 & \bfseries 0.27 \pm 0.07 \\
\cmidrule(lr){1-13}
\multirow{2}{*}{PointCircle2} & Episode return & 0.58 \pm 0.13 & -0.36 \pm 0.21 & 0.82 \pm 0.01 & \color{blue}\bfseries 0.51 \pm 0.02 & 0.82 \pm 0.01 & 0.55 \pm 0.06 & 0.63 \pm 0.04 & 0.65 \pm 0.05 & \bfseries -0.04 \pm 0.22 & 0.61 \pm 0.01 & \bfseries 0.60 \pm 0.02 \\
 & Episode cost & 2.77 \pm 0.11 & 1.93 \pm 0.58 & 2.94 \pm 0.05 & \color{blue}\bfseries 0.85 \pm 0.09 & 2.94 \pm 0.05 & 2.64 \pm 0.27 & 3.03 \pm 0.08 & 1.26 \pm 0.58 & \bfseries 0.73 \pm 0.29 & 1.30 \pm 0.07 & \bfseries 0.93 \pm 0.19 \\
\cmidrule(lr){1-13}
\multirow{2}{*}{PointGoal1} & Episode return & \bfseries 0.57 \pm 0.01 & 0.18 \pm 0.03 & 0.71 \pm 0.01 & \bfseries 0.25 \pm 0.03 & \bfseries 0.67 \pm 0.01 & \bfseries 0.58 \pm 0.02 & \color{blue}\bfseries 0.72 \pm 0.00 & \bfseries 0.57 \pm 0.01 & \bfseries 0.15 \pm 0.11 & \bfseries 0.35 \pm 0.01 & \bfseries 0.41 \pm 0.02 \\
 & Episode cost & \bfseries 0.92 \pm 0.03 & 1.15 \pm 0.16 & 1.00 \pm 0.04 & \bfseries 0.35 \pm 0.04 & \bfseries 0.88 \pm 0.04 & \bfseries 0.94 \pm 0.04 & \color{blue}\bfseries 0.99 \pm 0.02 & \bfseries 0.75 \pm 0.07 & \bfseries 0.63 \pm 0.34 & \bfseries 0.40 \pm 0.04 & \bfseries 0.44 \pm 0.05 \\
\cmidrule(lr){1-13}
\multirow{2}{*}{PointGoal2} & Episode return & 0.52 \pm 0.02 & 0.15 \pm 0.02 & 0.66 \pm 0.01 & \bfseries 0.18 \pm 0.01 & 0.55 \pm 0.01 & 0.49 \pm 0.01 & 0.63 \pm 0.01 & \color{blue}\bfseries 0.41 \pm 0.06 & \bfseries 0.18 \pm 0.08 & \color{blue}\bfseries 0.35 \pm 0.02 & \bfseries 0.38 \pm 0.02 \\
 & Episode cost & 1.16 \pm 0.09 & 1.30 \pm 0.10 & 1.35 \pm 0.01 & \bfseries 0.40 \pm 0.04 & 1.14 \pm 0.05 & 1.19 \pm 0.05 & 1.34 \pm 0.01 & \color{blue}\bfseries 0.81 \pm 0.10 & \bfseries 0.93 \pm 0.58 & \color{blue}\bfseries 0.61 \pm 0.05 & \bfseries 0.68 \pm 0.02 \\
\cmidrule(lr){1-13}
\multirow{2}{*}{PointPush1} & Episode return & \bfseries 0.20 \pm 0.01 & \bfseries 0.03 \pm 0.01 & 0.26 \pm 0.01 & \bfseries 0.07 \pm 0.01 & \color{blue}\bfseries 0.23 \pm 0.00 & \bfseries 0.15 \pm 0.01 & \color{blue}\bfseries 0.24 \pm 0.01 & \bfseries 0.17 \pm 0.03 & \bfseries 0.01 \pm 0.03 & \bfseries 0.16 \pm 0.01 & \bfseries 0.19 \pm 0.01 \\
 & Episode cost & \bfseries 0.89 \pm 0.03 & \bfseries 0.98 \pm 0.19 & 1.13 \pm 0.09 & \bfseries 0.19 \pm 0.01 & \color{blue}\bfseries 0.78 \pm 0.04 & \bfseries 0.76 \pm 0.03 & \color{blue}\bfseries 0.80 \pm 0.05 & \bfseries 0.73 \pm 0.11 & \bfseries 0.72 \pm 0.24 & \bfseries 0.50 \pm 0.03 & \bfseries 0.57 \pm 0.08 \\
\cmidrule(lr){1-13}
\multirow{2}{*}{PointPush2} & Episode return & 0.17 \pm 0.01 & -0.01 \pm 0.01 & 0.20 \pm 0.01 & \bfseries 0.09 \pm 0.01 & \color{blue}\bfseries 0.17 \pm 0.01 & 0.16 \pm 0.01 & 0.16 \pm 0.01 & \bfseries 0.09 \pm 0.03 & \bfseries -0.21 \pm 0.27 & 0.12 \pm 0.04 & \bfseries 0.15 \pm 0.03 \\
 & Episode cost & 1.49 \pm 0.14 & 1.29 \pm 0.27 & 1.35 \pm 0.12 & \bfseries 0.50 \pm 0.07 & \color{blue}\bfseries 0.98 \pm 0.09 & 1.09 \pm 0.04 & 1.34 \pm 0.04 & \bfseries 0.80 \pm 0.29 & \bfseries 0.60 \pm 0.33 & 1.08 \pm 0.19 & \bfseries 0.96 \pm 0.10 \\
\cmidrule(lr){1-13}
\multirow{2}{*}{CarButton1} & Episode return & 0.17 \pm 0.01 & 0.01 \pm 0.02 & 0.30 \pm 0.02 & \color{blue}\bfseries 0.03 \pm 0.05 & \bfseries -0.06 \pm 0.03 & 0.15 \pm 0.01 & 0.32 \pm 0.01 & 0.18 \pm 0.08 & -0.02 \pm 0.17 & \color{blue}\bfseries 0.04 \pm 0.05 & \bfseries 0.13 \pm 0.02 \\
 & Episode cost & 1.91 \pm 0.08 & 1.62 \pm 0.17 & 1.88 \pm 0.06 & \color{blue}\bfseries 0.91 \pm 0.14 & \bfseries 0.63 \pm 0.05 & 1.77 \pm 0.07 & 2.33 \pm 0.06 & 1.31 \pm 0.36 & 1.37 \pm 0.85 & \color{blue}\bfseries 0.63 \pm 0.05 & \bfseries 0.81 \pm 0.09 \\
\cmidrule(lr){1-13}
\multirow{2}{*}{CarButton2} & Episode return & 0.25 \pm 0.02 & -0.02 \pm 0.01 & 0.38 \pm 0.02 & 0.08 \pm 0.01 & \bfseries -0.09 \pm 0.03 & 0.23 \pm 0.01 & 0.43 \pm 0.00 & 0.26 \pm 0.05 & -0.05 \pm 0.10 & \color{blue}\bfseries 0.08 \pm 0.02 & \bfseries 0.08 \pm 0.02 \\
 & Episode cost & 2.15 \pm 0.05 & 1.64 \pm 0.17 & 2.85 \pm 0.11 & 1.13 \pm 0.12 & \bfseries 0.50 \pm 0.03 & 2.31 \pm 0.08 & 3.05 \pm 0.05 & 2.10 \pm 0.27 & 1.56 \pm 0.24 & \color{blue}\bfseries 0.86 \pm 0.13 & \bfseries 0.86 \pm 0.13 \\
\cmidrule(lr){1-13}
\multirow{2}{*}{CarCircle1} & Episode return & 0.25 \pm 0.04 & -0.06 \pm 0.05 & 0.72 \pm 0.01 & 0.21 \pm 0.05 & 0.70 \pm 0.01 & 0.29 \pm 0.06 & 0.67 \pm 0.00 & 0.64 \pm 0.04 & -0.09 \pm 0.12 & 0.53 \pm 0.03 & 0.56 \pm 0.03 \\
 & Episode cost & 1.46 \pm 0.20 & 1.69 \pm 0.45 & 2.20 \pm 0.12 & 1.13 \pm 0.26 & 2.52 \pm 0.06 & 1.56 \pm 0.14 & 2.62 \pm 0.04 & 2.44 \pm 0.22 & 1.05 \pm 0.42 & 1.27 \pm 0.22 & 1.37 \pm 0.17 \\
\cmidrule(lr){1-13}
\multirow{2}{*}{CarCircle2} & Episode return & 0.36 \pm 0.02 & 0.19 \pm 0.04 & 0.63 \pm 0.04 & 0.49 \pm 0.02 & 0.71 \pm 0.01 & 0.34 \pm 0.02 & 0.75 \pm 0.00 & 0.65 \pm 0.05 & -0.33 \pm 0.04 & 0.58 \pm 0.02 & 0.63 \pm 0.01 \\
 & Episode cost & 1.88 \pm 0.21 & 2.66 \pm 0.23 & 2.87 \pm 0.15 & 1.55 \pm 0.20 & 2.90 \pm 0.08 & 1.24 \pm 0.16 & 3.28 \pm 0.05 & 2.38 \pm 0.76 & 3.75 \pm 0.95 & 2.00 \pm 0.12 & 2.01 \pm 0.13 \\
\cmidrule(lr){1-13}
\multirow{2}{*}{CarGoal1} & Episode return & 0.48 \pm 0.01 & 0.25 \pm 0.02 & 0.68 \pm 0.01 & \bfseries 0.14 \pm 0.01 & \color{blue}\bfseries 0.45 \pm 0.01 & \color{blue}\bfseries 0.43 \pm 0.00 & 0.71 \pm 0.01 & \color{blue}\bfseries 0.48 \pm 0.09 & \bfseries -0.02 \pm 0.06 & \bfseries 0.31 \pm 0.01 & \bfseries 0.31 \pm 0.01 \\
 & Episode cost & 1.22 \pm 0.06 & 1.39 \pm 0.08 & 1.23 \pm 0.05 & \bfseries 0.24 \pm 0.03 & \color{blue}\bfseries 0.62 \pm 0.02 & \color{blue}\bfseries 0.96 \pm 0.05 & 1.38 \pm 0.06 & \color{blue}\bfseries 0.77 \pm 0.22 & \bfseries 0.65 \pm 0.47 & \bfseries 0.32 \pm 0.02 & \bfseries 0.32 \pm 0.02 \\
\cmidrule(lr){1-13}
\multirow{2}{*}{CarGoal2} & Episode return & 0.36 \pm 0.01 & 0.02 \pm 0.02 & 0.51 \pm 0.01 & \bfseries 0.18 \pm 0.01 & \color{blue}\bfseries 0.29 \pm 0.01 & 0.36 \pm 0.01 & 0.50 \pm 0.01 & 0.28 \pm 0.04 & \bfseries -0.07 \pm 0.14 & \bfseries 0.24 \pm 0.02 & \bfseries 0.26 \pm 0.01 \\
 & Episode cost & 2.71 \pm 0.13 & 1.80 \pm 0.22 & 2.68 \pm 0.09 & \bfseries 0.97 \pm 0.10 & \color{blue}\bfseries 0.99 \pm 0.08 & 2.58 \pm 0.12 & 2.42 \pm 0.06 & 1.24 \pm 0.35 & \bfseries 0.58 \pm 0.60 & \bfseries 0.81 \pm 0.08 & \bfseries 0.90 \pm 0.09 \\
\cmidrule(lr){1-13}
\multirow{2}{*}{CarPush1} & Episode return & \bfseries 0.20 \pm 0.01 & \bfseries 0.03 \pm 0.02 & \color{blue}\bfseries 0.28 \pm 0.01 & \bfseries 0.14 \pm 0.01 & \bfseries 0.22 \pm 0.01 & \bfseries 0.18 \pm 0.01 & \color{blue}\bfseries 0.27 \pm 0.01 & \bfseries 0.21 \pm 0.02 & \bfseries -0.01 \pm 0.02 & \bfseries 0.19 \pm 0.01 & \bfseries 0.21 \pm 0.00 \\
 & Episode cost & \bfseries 0.95 \pm 0.05 & \bfseries 0.35 \pm 0.06 & \color{blue}\bfseries 0.92 \pm 0.09 & \bfseries 0.30 \pm 0.03 & \bfseries 0.43 \pm 0.05 & \bfseries 0.85 \pm 0.07 & \color{blue}\bfseries 0.92 \pm 0.06 & \bfseries 0.52 \pm 0.10 & \bfseries 0.22 \pm 0.23 & \bfseries 0.33 \pm 0.04 & \bfseries 0.30 \pm 0.02 \\
\cmidrule(lr){1-13}
\multirow{2}{*}{CarPush2} & Episode return & 0.07 \pm 0.00 & 0.02 \pm 0.02 & 0.17 \pm 0.01 & \bfseries 0.03 \pm 0.02 & 0.11 \pm 0.00 & 0.09 \pm 0.00 & 0.17 \pm 0.00 & 0.10 \pm 0.02 & 0.04 \pm 0.03 & \color{blue}\bfseries 0.10 \pm 0.00 & \bfseries 0.12 \pm 0.01 \\
 & Episode cost & 2.49 \pm 0.21 & 2.55 \pm 0.28 & 2.05 \pm 0.13 & \bfseries 0.61 \pm 0.03 & 1.15 \pm 0.10 & 1.96 \pm 0.11 & 2.08 \pm 0.08 & 1.20 \pm 0.44 & 1.34 \pm 1.05 & \color{blue}\bfseries 0.91 \pm 0.10 & \bfseries 0.93 \pm 0.10 \\
\cmidrule(lr){1-13}
\multirow{2}{*}{SwimmerVelocity} & Episode return & 0.73 \pm 0.01 & \bfseries 0.68 \pm 0.00 & 0.73 \pm 0.01 & \bfseries 0.28 \pm 0.04 & 0.81 \pm 0.01 & \color{blue}\bfseries 0.69 \pm 0.00 & 0.70 \pm 0.02 & 0.66 \pm 0.01 & \bfseries 0.67 \pm 0.00 & \bfseries 0.62 \pm 0.01 & \bfseries 0.62 \pm 0.01 \\
 & Episode cost & 2.71 \pm 0.23 & \bfseries 0.35 \pm 0.03 & 3.55 \pm 0.46 & \bfseries 0.58 \pm 0.07 & 4.66 \pm 0.36 & \color{blue}\bfseries 0.60 \pm 0.13 & 2.35 \pm 0.12 & 3.11 \pm 0.50 & \bfseries 0.14 \pm 0.23 & \bfseries 0.65 \pm 0.28 & \bfseries 0.65 \pm 0.28 \\
\cmidrule(lr){1-13}
\multirow{2}{*}{HopperVelocity} & Episode return & 0.35 \pm 0.11 & \color{blue}\bfseries 0.84 \pm 0.00 & 0.59 \pm 0.12 & \bfseries 0.08 \pm 0.03 & 0.29 \pm 0.09 & 0.38 \pm 0.03 & 0.70 \pm 0.02 & \bfseries 0.15 \pm 0.02 & \color{blue}\bfseries 0.73 \pm 0.13 & \bfseries 0.36 \pm 0.09 & \bfseries 0.47 \pm 0.07 \\
 & Episode cost & 1.71 \pm 0.32 & \color{blue}\bfseries 0.00 \pm 0.00 & 2.97 \pm 0.60 & \bfseries 0.52 \pm 0.23 & 2.39 \pm 0.96 & 2.28 \pm 0.21 & 3.30 \pm 0.12 & \bfseries 0.64 \pm 0.12 & \color{blue}\bfseries 0.51 \pm 0.41 & \bfseries 0.40 \pm 0.16 & \bfseries 0.60 \pm 0.12 \\
\cmidrule(lr){1-13}
\multirow{2}{*}{HalfCheetahVelocity} & Episode return & 0.98 \pm 0.00 & \bfseries 0.66 \pm 0.07 & 0.99 \pm 0.00 & \bfseries 0.76 \pm 0.01 & 0.99 \pm 0.00 & 0.98 \pm 0.00 & 0.97 \pm 0.00 & 0.92 \pm 0.01 & \color{blue}\bfseries 0.96 \pm 0.01 & \bfseries 0.76 \pm 0.01 & \bfseries 0.76 \pm 0.01 \\
 & Episode cost & 1.92 \pm 0.41 & \bfseries 0.13 \pm 0.06 & 2.82 \pm 0.73 & \bfseries 0.74 \pm 0.14 & 4.30 \pm 0.80 & 1.44 \pm 0.28 & 1.61 \pm 0.14 & 1.74 \pm 0.78 & \color{blue}\bfseries 0.01 \pm 0.02 & \bfseries 0.45 \pm 0.12 & \bfseries 0.45 \pm 0.12 \\
\cmidrule(lr){1-13}
\multirow{2}{*}{Walker2dVelocity} & Episode return & 0.83 \pm 0.02 & \bfseries 0.25 \pm 0.04 & 0.80 \pm 0.02 & \bfseries 0.73 \pm 0.02 & 0.77 \pm 0.01 & 0.78 \pm 0.02 & 0.75 \pm 0.03 & 0.60 \pm 0.02 & \color{blue}\bfseries 0.78 \pm 0.00 & \color{blue}\bfseries 0.79 \pm 0.01 & \bfseries 0.79 \pm 0.01 \\
 & Episode cost & 3.27 \pm 0.07 & \bfseries 0.49 \pm 0.06 & 3.35 \pm 0.04 & \bfseries 0.94 \pm 0.26 & 2.25 \pm 0.21 & 2.96 \pm 0.09 & 3.05 \pm 0.09 & 2.40 \pm 0.34 & \color{blue}\bfseries 0.00 \pm 0.01 & \color{blue}\bfseries 0.07 \pm 0.02 & \bfseries 0.07 \pm 0.02 \\
\cmidrule(lr){1-13}
\multirow{2}{*}{AntVelocity} & Episode return & 0.99 \pm 0.00 & \bfseries -0.02 \pm 0.01 & 0.99 \pm 0.00 & \color{blue}\bfseries 0.96 \pm 0.00 & 1.00 \pm 0.00 & 0.97 \pm 0.01 & 0.99 \pm 0.00 & 0.91 \pm 0.02 & \bfseries 0.90 \pm 0.01 & \color{blue}\bfseries 0.96 \pm 0.00 & \bfseries 0.96 \pm 0.00 \\
 & Episode cost & 1.34 \pm 0.22 & \bfseries 0.04 \pm 0.01 & 1.31 \pm 0.15 & \color{blue}\bfseries 0.30 \pm 0.05 & 2.18 \pm 0.38 & 1.33 \pm 0.13 & 2.03 \pm 0.04 & 1.58 \pm 0.35 & \bfseries 0.08 \pm 0.01 & \color{blue}\bfseries 0.15 \pm 0.03 & \bfseries 0.15 \pm 0.03 \\
\cmidrule(lr){1-13}
\multirow{2}{*}{BallRun} & Episode return & 0.44 \pm 0.08 & 0.32 \pm 0.10 & 0.70 \pm 0.07 & 0.64 \pm 0.19 & \bfseries -0.01 \pm 0.00 & 0.52 \pm 0.03 & 0.57 \pm 0.01 & \color{blue}\bfseries 0.43 \pm 0.21 & \color{blue}\bfseries 0.31 \pm 0.00 & \color{blue}\bfseries 0.38 \pm 0.05 & \bfseries 0.41 \pm 0.04 \\
 & Episode cost & 1.18 \pm 0.16 & 1.01 \pm 0.05 & 1.61 \pm 0.15 & 1.36 \pm 0.38 & \bfseries 0.00 \pm 0.00 & 1.14 \pm 0.11 & 1.41 \pm 0.03 & \color{blue}\bfseries 0.79 \pm 0.78 & \color{blue}\bfseries 0.48 \pm 0.09 & \color{blue}\bfseries 0.89 \pm 0.18 & \bfseries 0.86 \pm 0.19 \\
\cmidrule(lr){1-13}
\multirow{2}{*}{CarRun} & Episode return & \bfseries 0.97 \pm 0.01 & \color{blue}\bfseries 0.99 \pm 0.00 & \color{blue}\bfseries 0.99 \pm 0.00 & 0.64 \pm 0.05 & \bfseries 0.36 \pm 0.06 & \color{blue}\bfseries 0.99 \pm 0.00 & \color{blue}\bfseries 0.99 \pm 0.00 & 0.97 \pm 0.00 & \color{blue}\bfseries 0.99 \pm 0.00 & \bfseries 0.96 \pm 0.00 & \bfseries 0.97 \pm 0.01 \\
 & Episode cost & \bfseries 0.45 \pm 0.15 & \color{blue}\bfseries 0.51 \pm 0.04 & \color{blue}\bfseries 0.37 \pm 0.05 & 1.79 \pm 0.58 & \bfseries 0.42 \pm 0.32 & \color{blue}\bfseries 0.16 \pm 0.02 & \color{blue}\bfseries 0.37 \pm 0.02 & 1.19 \pm 0.35 & \color{blue}\bfseries 0.48 \pm 0.38 & \bfseries 0.07 \pm 0.03 & \bfseries 0.79 \pm 0.30 \\
\cmidrule(lr){1-13}
\multirow{2}{*}{DroneRun} & Episode return & 0.56 \pm 0.09 & 0.59 \pm 0.03 & 0.66 \pm 0.07 & 0.48 \pm 0.03 & 0.77 \pm 0.01 & 0.64 \pm 0.05 & 0.74 \pm 0.01 & \color{blue}\bfseries 0.53 \pm 0.06 & \color{blue}\bfseries 0.55 \pm 0.01 & 0.60 \pm 0.01 & \bfseries 0.60 \pm 0.01 \\
 & Episode cost & 2.34 \pm 0.16 & 2.57 \pm 0.21 & 2.41 \pm 0.36 & 2.55 \pm 0.14 & 2.15 \pm 0.06 & 2.67 \pm 0.13 & 2.48 \pm 0.04 & \color{blue}\bfseries 0.67 \pm 0.88 & \color{blue}\bfseries 0.00 \pm 0.00 & 1.07 \pm 0.29 & \bfseries 0.68 \pm 0.20 \\
\cmidrule(lr){1-13}
\multirow{2}{*}{AntRun} & Episode return & 0.54 \pm 0.02 & \bfseries 0.10 \pm 0.01 & 0.68 \pm 0.03 & \bfseries 0.42 \pm 0.03 & 0.77 \pm 0.01 & 0.56 \pm 0.02 & 0.56 \pm 0.03 & \bfseries 0.26 \pm 0.01 & \bfseries 0.33 \pm 0.05 & \color{blue}\bfseries 0.55 \pm 0.03 & \bfseries 0.55 \pm 0.03 \\
 & Episode cost & 1.60 \pm 0.12 & \bfseries 0.02 \pm 0.00 & 1.99 \pm 0.17 & \bfseries 0.19 \pm 0.04 & 1.86 \pm 0.18 & 1.68 \pm 0.14 & 1.47 \pm 0.13 & \bfseries 0.09 \pm 0.03 & \bfseries 0.02 \pm 0.01 & \color{blue}\bfseries 0.46 \pm 0.13 & \bfseries 0.46 \pm 0.13 \\
\cmidrule(lr){1-13}
\multirow{2}{*}{BallCircle} & Episode return & 0.92 \pm 0.01 & \color{blue}\bfseries 0.82 \pm 0.00 & 0.94 \pm 0.01 & \bfseries 0.56 \pm 0.02 & 0.90 \pm 0.00 & 0.84 \pm 0.02 & 0.90 \pm 0.00 & \bfseries 0.64 \pm 0.02 & \bfseries 0.41 \pm 0.04 & \bfseries 0.60 \pm 0.02 & \bfseries 0.65 \pm 0.02 \\
 & Episode cost & 1.44 \pm 0.02 & \color{blue}\bfseries 0.94 \pm 0.04 & 1.44 \pm 0.03 & \bfseries 0.48 \pm 0.05 & 1.34 \pm 0.02 & 1.23 \pm 0.04 & 1.42 \pm 0.01 & \bfseries 0.90 \pm 0.09 & \bfseries 0.32 \pm 0.12 & \bfseries 0.43 \pm 0.04 & \bfseries 0.54 \pm 0.05 \\
\cmidrule(lr){1-13}
\multirow{2}{*}{CarCircle} & Episode return & 0.86 \pm 0.04 & 0.77 \pm 0.01 & 0.92 \pm 0.01 & \bfseries 0.56 \pm 0.02 & 0.92 \pm 0.00 & 0.84 \pm 0.01 & 0.31 \pm 0.05 & 0.50 \pm 0.03 & 0.13 \pm 0.02 & \color{blue}\bfseries 0.70 \pm 0.01 & \bfseries 0.70 \pm 0.01 \\
 & Episode cost & 2.27 \pm 0.06 & 1.20 \pm 0.08 & 2.19 \pm 0.04 & \bfseries 0.52 \pm 0.16 & 2.15 \pm 0.02 & 1.85 \pm 0.08 & 1.41 \pm 0.12 & 1.14 \pm 0.08 & 1.52 \pm 0.78 & \color{blue}\bfseries 0.82 \pm 0.03 & \bfseries 0.82 \pm 0.03 \\
\cmidrule(lr){1-13}
\multirow{2}{*}{DroneCircle} & Episode return & 0.06 \pm 0.03 & -0.12 \pm 0.01 & 0.52 \pm 0.05 & \bfseries 0.18 \pm 0.01 & 0.78 \pm 0.01 & 0.36 \pm 0.06 & 0.38 \pm 0.03 & 0.04 \pm 0.03 & -0.07 \pm 0.08 & \color{blue}\bfseries 0.54 \pm 0.01 & \bfseries 0.54 \pm 0.01 \\
 & Episode cost & 1.25 \pm 0.13 & 1.12 \pm 0.06 & 1.77 \pm 0.12 & \bfseries 0.81 \pm 0.06 & 1.61 \pm 0.04 & 1.61 \pm 0.09 & 1.57 \pm 0.06 & 1.91 \pm 0.31 & 1.26 \pm 1.05 & \color{blue}\bfseries 0.80 \pm 0.04 & \bfseries 0.80 \pm 0.04 \\
\cmidrule(lr){1-13}
\multirow{2}{*}{AntCircle} & Episode return & 0.20 \pm 0.01 & \bfseries 0.09 \pm 0.01 & 0.45 \pm 0.01 & \bfseries 0.19 \pm 0.01 & 0.57 \pm 0.02 & \bfseries 0.20 \pm 0.01 & 0.35 \pm 0.01 & 0.16 \pm 0.04 & \bfseries 0.13 \pm 0.02 & \color{blue}\bfseries 0.35 \pm 0.01 & \bfseries 0.35 \pm 0.01 \\
 & Episode cost & 1.27 \pm 0.10 & \bfseries 0.47 \pm 0.10 & 1.22 \pm 0.08 & \bfseries 0.49 \pm 0.05 & 1.77 \pm 0.07 & \bfseries 0.94 \pm 0.05 & 1.17 \pm 0.09 & 1.27 \pm 0.38 & \bfseries 0.13 \pm 0.05 & \color{blue}\bfseries 0.48 \pm 0.06 & \bfseries 0.48 \pm 0.06 \\
\cmidrule(lr){1-13}
\multirow{2}{*}{MetaDrive-easysparse} & Episode return & 0.92 \pm 0.07 & \color{blue}\bfseries 0.38 \pm 0.05 & 0.99 \pm 0.00 & \bfseries 0.01 \pm 0.02 & 0.99 \pm 0.01 & 0.98 \pm 0.02 & 0.99 \pm 0.00 & \color{blue}\bfseries 0.31 \pm 0.25 & \bfseries -0.01 \pm 0.02 & \bfseries 0.23 \pm 0.07 & \bfseries 0.56 \pm 0.09 \\
 & Episode cost & 2.06 \pm 0.13 & \color{blue}\bfseries 0.50 \pm 0.07 & 2.11 \pm 0.06 & \bfseries 0.15 \pm 0.03 & 1.87 \pm 0.11 & 2.13 \pm 0.05 & 2.05 \pm 0.01 & \color{blue}\bfseries 0.02 \pm 0.00 & \bfseries 0.03 \pm 0.00 & \bfseries 0.33 \pm 0.05 & \bfseries 0.40 \pm 0.20 \\
\cmidrule(lr){1-13}
\multirow{2}{*}{MetaDrive-easymean} & Episode return & 0.97 \pm 0.01 & \bfseries 0.25 \pm 0.06 & 0.86 \pm 0.07 & \bfseries 0.53 \pm 0.03 & 0.86 \pm 0.09 & 0.93 \pm 0.01 & 0.84 \pm 0.01 & \color{blue}\bfseries 0.47 \pm 0.27 & \bfseries 0.13 \pm 0.11 & \color{blue}\bfseries 0.70 \pm 0.04 & \bfseries 0.73 \pm 0.01 \\
 & Episode cost & 2.01 \pm 0.02 & \bfseries 0.39 \pm 0.06 & 1.70 \pm 0.08 & \bfseries 0.33 \pm 0.02 & 1.66 \pm 0.25 & 1.98 \pm 0.03 & 1.86 \pm 0.03 & \color{blue}\bfseries 0.02 \pm 0.00 & \bfseries 0.03 \pm 0.00 & \color{blue}\bfseries 0.45 \pm 0.02 & \bfseries 0.51 \pm 0.03 \\
\cmidrule(lr){1-13}
\multirow{2}{*}{MetaDrive-easydense} & Episode return & 0.87 \pm 0.02 & \bfseries 0.48 \pm 0.02 & 0.61 \pm 0.03 & \color{blue}\bfseries 0.46 \pm 0.11 & 0.69 \pm 0.03 & 0.84 \pm 0.02 & 0.56 \pm 0.02 & \bfseries 0.26 \pm 0.09 & \bfseries 0.24 \pm 0.04 & \color{blue}\bfseries 0.56 \pm 0.04 & \bfseries 0.56 \pm 0.04 \\
 & Episode cost & 1.60 \pm 0.02 & \bfseries 0.49 \pm 0.02 & 1.35 \pm 0.04 & \color{blue}\bfseries 0.31 \pm 0.16 & 1.45 \pm 0.03 & 1.58 \pm 0.03 & 1.19 \pm 0.03 & \bfseries 0.03 \pm 0.00 & \bfseries 0.03 \pm 0.00 & \color{blue}\bfseries 0.25 \pm 0.04 & \bfseries 0.25 \pm 0.04 \\
\cmidrule(lr){1-13}
\multirow{2}{*}{MetaDrive-mediumsparse} & Episode return & 0.86 \pm 0.02 & 0.28 \pm 0.04 & 0.82 \pm 0.09 & \color{blue}\bfseries 0.24 \pm 0.08 & 0.81 \pm 0.03 & 0.78 \pm 0.09 & 0.87 \pm 0.02 & \color{blue}\bfseries 0.18 \pm 0.01 & \color{blue}\bfseries 0.11 \pm 0.11 & 0.79 \pm 0.05 & \bfseries 0.74 \pm 0.09 \\
 & Episode cost & 1.57 \pm 0.11 & 1.02 \pm 0.05 & 1.94 \pm 0.19 & \color{blue}\bfseries 0.45 \pm 0.08 & 1.94 \pm 0.11 & 1.44 \pm 0.21 & 2.16 \pm 0.01 & \color{blue}\bfseries 0.05 \pm 0.00 & \color{blue}\bfseries 0.05 \pm 0.00 & 1.29 \pm 0.08 & \bfseries 0.88 \pm 0.15 \\
\cmidrule(lr){1-13}
\multirow{2}{*}{MetaDrive-mediummean} & Episode return & 0.94 \pm 0.00 & \bfseries 0.05 \pm 0.06 & 0.66 \pm 0.14 & \color{blue}\bfseries 0.44 \pm 0.08 & 0.89 \pm 0.03 & 0.92 \pm 0.01 & 0.80 \pm 0.03 & \color{blue}\bfseries 0.21 \pm 0.18 & \color{blue}\bfseries 0.65 \pm 0.27 & \color{blue}\bfseries 0.51 \pm 0.14 & \bfseries 0.51 \pm 0.14 \\
 & Episode cost & 1.64 \pm 0.07 & \bfseries 0.25 \pm 0.09 & 1.70 \pm 0.31 & \color{blue}\bfseries 0.27 \pm 0.03 & 1.90 \pm 0.09 & 1.56 \pm 0.11 & 1.99 \pm 0.08 & \color{blue}\bfseries 0.05 \pm 0.00 & \color{blue}\bfseries 0.04 \pm 0.01 & \color{blue}\bfseries 0.82 \pm 0.14 & \bfseries 0.82 \pm 0.14 \\
\cmidrule(lr){1-13}
\multirow{2}{*}{MetaDrive-mediumdense} & Episode return & 0.85 \pm 0.02 & \bfseries 0.27 \pm 0.04 & 0.85 \pm 0.03 & \bfseries 0.16 \pm 0.04 & 0.88 \pm 0.01 & 0.89 \pm 0.00 & 0.42 \pm 0.08 & \bfseries 0.16 \pm 0.08 & \bfseries 0.06 \pm 0.12 & \color{blue}\bfseries 0.52 \pm 0.07 & \bfseries 0.53 \pm 0.08 \\
 & Episode cost & 1.81 \pm 0.05 & \bfseries 0.47 \pm 0.08 & 2.03 \pm 0.08 & \bfseries 0.05 \pm 0.01 & 1.98 \pm 0.04 & 1.77 \pm 0.04 & 1.08 \pm 0.14 & \bfseries 0.05 \pm 0.00 & \bfseries 0.05 \pm 0.00 & \color{blue}\bfseries 0.48 \pm 0.15 & \bfseries 0.73 \pm 0.14 \\
\cmidrule(lr){1-13}
\multirow{2}{*}{MetaDrive-hardsparse} & Episode return & 0.48 \pm 0.02 & \color{blue}\bfseries 0.24 \pm 0.08 & 0.56 \pm 0.01 & \color{blue}\bfseries 0.38 \pm 0.01 & 0.46 \pm 0.03 & 0.51 \pm 0.01 & 0.44 \pm 0.00 & \bfseries 0.18 \pm 0.08 & \color{blue}\bfseries 0.40 \pm 0.11 & \color{blue}\bfseries 0.35 \pm 0.02 & \bfseries 0.35 \pm 0.02 \\
 & Episode cost & 1.79 \pm 0.07 & \color{blue}\bfseries 0.44 \pm 0.11 & 2.31 \pm 0.12 & \color{blue}\bfseries 0.03 \pm 0.00 & 1.91 \pm 0.09 & 1.84 \pm 0.09 & 1.29 \pm 0.03 & \bfseries 0.05 \pm 0.00 & \color{blue}\bfseries 0.04 \pm 0.01 & \color{blue}\bfseries 0.13 \pm 0.03 & \bfseries 0.13 \pm 0.03 \\
\cmidrule(lr){1-13}
\multirow{2}{*}{MetaDrive-hardmean} & Episode return & 0.42 \pm 0.04 & \bfseries 0.05 \pm 0.05 & 0.44 \pm 0.03 & \color{blue}\bfseries 0.43 \pm 0.02 & 0.50 \pm 0.02 & 0.42 \pm 0.03 & 0.36 \pm 0.01 & \bfseries 0.17 \pm 0.10 & \bfseries 0.30 \pm 0.07 & 0.47 \pm 0.02 & \bfseries 0.47 \pm 0.01 \\
 & Episode cost & 1.66 \pm 0.09 & \bfseries 0.39 \pm 0.12 & 2.08 \pm 0.09 & \color{blue}\bfseries 0.45 \pm 0.02 & 1.97 \pm 0.08 & 1.62 \pm 0.08 & 1.91 \pm 0.04 & \bfseries 0.05 \pm 0.00 & \bfseries 0.05 \pm 0.00 & 1.76 \pm 0.12 & \bfseries 0.85 \pm 0.05 \\
\cmidrule(lr){1-13}
\multirow{2}{*}{MetaDrive-harddense} & Episode return & \color{blue}\bfseries 0.52 \pm 0.01 & \bfseries 0.28 \pm 0.02 & 0.52 \pm 0.01 & \bfseries 0.30 \pm 0.05 & 0.33 \pm 0.05 & 0.40 \pm 0.04 & 0.34 \pm 0.01 & \bfseries 0.17 \pm 0.11 & \bfseries 0.18 \pm 0.12 & \bfseries 0.34 \pm 0.04 & \bfseries 0.36 \pm 0.03 \\
 & Episode cost & \color{blue}\bfseries 0.98 \pm 0.04 & \bfseries 0.59 \pm 0.02 & 1.46 \pm 0.09 & \bfseries 0.52 \pm 0.05 & 1.47 \pm 0.19 & 1.31 \pm 0.10 & 1.61 \pm 0.05 & \bfseries 0.05 \pm 0.00 & \bfseries 0.05 \pm 0.00 & \bfseries 0.75 \pm 0.08 & \bfseries 0.65 \pm 0.10 \\
  \bottomrule
     \end{tabular}
      \end{adjustbox}
      \vspace{2pt}
      \caption{Per-task safety-performance comparison on all 38 re-curated DSRL tasks ($N_{\DS}=N_{\DP}=10$). Results are averaged over five seeds, with standard errors reported. The last two columns report DexDICE with the fixed default $\kappa=0.01$ and with the per-task best $\kappa$ selected from the sweep $\kappa\in\{0.001, 0.005, 0.01, 0.05, 0.1, 0.2\}$ (highest return among settings satisfying the cost constraint; the highest-return setting if none is feasible). \textbf{Bold} marks methods satisfying the cost constraint (mean cost $\le$ 1, normalized to the cost upper bound). Among safety-feasible results, we highlight in \textcolor{blue}{\textbf{blue}} those whose error bars overlap with the highest mean episode return; the per-task best-$\kappa$ column (right of the vertical rule) is excluded from this comparison and is only marked for constraint satisfaction. This table is the per-task complement of the aggregated results in Table~\ref{tab:main_comparison}.}
      \label{app:full_dsrl_table}
      \vspace{-12pt}
  \end{table*}

\subsection{Empirical Comparison with Offline Safe Methods in Different Settings}
\label{app:comparison_with_baselines}

In this section we compare DexDICE with a broader set of baselines from two families that operate under different supervision settings: DICE-based methods (SafeDICE~\citep{jang2023safedice}, COptiDICE~\citep{leecoptidice}, ICSDICE~\citep{quan2024learning}) and offline safe RL methods (CDT~\citep{liu2023constrained}, FISOR~\citep{zheng2024safe}, TraC~\citep{yang2024offline}). Despite their differing assumptions, all are evaluated under our safety and return metrics for conceptual and practical comparison.

\paragraph{Setup.}
SafeDICE treats unsafe demonstrations as non-preferred. COptiDICE, CDT, FISOR, and TraC use ground-truth reward $r(s,a)$ and cost $c(s,a)$ from the dataset. ICSDICE uses true rewards and safe demonstrations $\DS$. DexDICE is trained using only $\DS$, $\DP$, and $\DU$. Its per-task best $\kappa$ in Table~\ref{tab:combined-rl-performance} is selected post hoc using evaluation return and cost, separately from the fixed-$\kappa$ main result.

\paragraph{Results.}
With the fixed $\kappa=0.01$ used in the main result, DexDICE satisfies the cost constraint on $31$ of the $38$ tasks (Table~\ref{app:full_dsrl_table}).
Table~\ref{tab:combined-rl-performance} separately reports the exploratory comparison using DexDICE's post hoc per-task best $\kappa$.
Within the DICE family, SafeDICE violates the cost constraint on $20$ of the $38$ tasks since its non-preferred set does not adequately cover the unsafe support, leaving exploitable gaps. COptiDICE, despite ground-truth reward and cost access, violates the constraint on $19$ tasks, consistent with the observation in the DSRL paper~\citep{liu2024datasets} that behavior-distribution constraints (e.g., $D_f(d\|\dU)$) induce conservatism and brittle safety under lower-quality data. ICSDICE attains the highest returns on average but violates the constraint on $22$ tasks, concentrated in the velocity, Bullet Safety Gym, and MetaDrive tasks.
Among offline safe RL methods, FISOR, FISOR-IL, and TraC satisfy the constraint on $34$, $35$, and $30$ tasks, respectively, under full ground-truth supervision but pay a notable return cost; their constrained-RL formulations bias the policy toward conservative behavior even when the dataset already contains high-return safe trajectories. CDT attains higher returns but violates the constraint on $18$ tasks.

\clearpage
\begin{table*}[p]
  \centering
  \begin{adjustbox}{width=\textwidth, totalheight=0.92\textheight, keepaspectratio}
  \aboverulesep=0.1ex
  \belowrulesep=0.1ex
  \renewcommand{\arraystretch}{1.1}
  \begin{tabular}
  {l|l|
   S[table-format=1.2(1.2),mode=text]
   S[table-format=1.2(1.2),mode=text]
   S[table-format=1.2(1.2),mode=text]
   S[table-format=1.2(1.2),mode=text]
   S[table-format=1.2(1.2),mode=text]
   S[table-format=1.2(1.2),mode=text]
   S[table-format=1.2(1.2),mode=text]
   S[table-format=1.2(1.2),mode=text]
   }
     \toprule
  \textbf{Environment} & \textbf{Metric} & \textbf{COptiDICE} & \textbf{CDT} & \textbf{TraC} & \textbf{FISOR} & \textbf{FISOR-IL} & \textbf{SafeDICE} & \textbf{ICSDICE} & \textbf{DexDICE} \\
  \hline\hline
  \multicolumn{2}{c|}{\textbf{Training Inputs}} &
  \multicolumn{1}{c}{$r(s,a), c(s,a), \mathcal{D}_\rU$} &
  \multicolumn{1}{c}{$r(s,a), c(s,a), \mathcal{D}_\rU$} &
  \multicolumn{1}{c}{$r(s,a), c(s,a), \mathcal{D}_\rU$} &
  \multicolumn{1}{c}{$r(s,a), c(s,a), \mathcal{D}_\rU$} &
  \multicolumn{1}{c}{$\hat{c}(s,a), \mathcal{D}_\rU$} &
  \multicolumn{1}{c}{$\mathcal{D}_\mathrm{N}, \mathcal{D}_\rP$} &
  \multicolumn{1}{c}{$r(s,a), \mathcal{D}_\rS, \mathcal{D}_\rU$} &
  \multicolumn{1}{c}{$\mathcal{D}_\rP, \mathcal{D}_\rS, \mathcal{D}_\rU$} \\
  \hline
\multirow{2}{*}{PointButton1} & Episode return & \bfseries 0.13 \pm 0.01 & 0.59 \pm 0.01 & 0.14 \pm 0.03 & \bfseries 0.06 \pm 0.01 & \bfseries 0.03 \pm 0.01 & 0.11 \pm 0.01 & \bfseries 0.09 \pm 0.01 & \color{blue}\bfseries 0.15 \pm 0.02 \\
 & Episode cost & \bfseries 0.92 \pm 0.12 & 2.77 \pm 0.31 & 1.08 \pm 0.19 & \bfseries 0.42 \pm 0.03 & \bfseries 0.41 \pm 0.04 & 1.04 \pm 0.14 & \bfseries 0.69 \pm 0.08 & \color{blue}\bfseries 0.95 \pm 0.09 \\
\cmidrule(lr){1-10}
\multirow{2}{*}{PointButton2} & Episode return & \bfseries 0.22 \pm 0.02 & 0.55 \pm 0.03 & 0.20 \pm 0.04 & \bfseries 0.10 \pm 0.01 & \bfseries 0.09 \pm 0.01 & \bfseries 0.20 \pm 0.01 & \color{blue}\bfseries 0.25 \pm 0.02 & \color{blue}\bfseries 0.27 \pm 0.02 \\
 & Episode cost & \bfseries 0.71 \pm 0.04 & 1.72 \pm 0.10 & 1.11 \pm 0.42 & \bfseries 0.34 \pm 0.05 & \bfseries 0.32 \pm 0.04 & \bfseries 0.64 \pm 0.04 & \color{blue}\bfseries 0.64 \pm 0.07 & \color{blue}\bfseries 0.92 \pm 0.08 \\
\cmidrule(lr){1-10}
\multirow{2}{*}{PointCircle1} & Episode return & 0.74 \pm 0.03 & \color{blue}\bfseries 0.57 \pm 0.01 & \bfseries 0.52 \pm 0.01 & 0.40 \pm 0.03 & 0.46 \pm 0.01 & 0.62 \pm 0.03 & 0.67 \pm 0.02 & \bfseries 0.27 \pm 0.03 \\
 & Episode cost & 4.57 \pm 0.28 & \color{blue}\bfseries 0.75 \pm 0.05 & \bfseries 0.07 \pm 0.03 & 3.94 \pm 0.31 & 1.41 \pm 0.21 & 4.02 \pm 0.29 & 3.66 \pm 0.12 & \bfseries 0.27 \pm 0.07 \\
\cmidrule(lr){1-10}
\multirow{2}{*}{PointCircle2} & Episode return & 0.72 \pm 0.02 & \color{blue}\bfseries 0.64 \pm 0.01 & 0.64 \pm 0.01 & 0.74 \pm 0.01 & 0.71 \pm 0.01 & 0.72 \pm 0.02 & 0.64 \pm 0.02 & \bfseries 0.60 \pm 0.02 \\
 & Episode cost & 2.31 \pm 0.15 & \color{blue}\bfseries 0.93 \pm 0.03 & 1.19 \pm 0.06 & 2.23 \pm 0.13 & 2.05 \pm 0.12 & 2.48 \pm 0.08 & 1.97 \pm 0.17 & \bfseries 0.93 \pm 0.19 \\
\cmidrule(lr){1-10}
\multirow{2}{*}{PointGoal1} & Episode return & \color{blue}\bfseries 0.66 \pm 0.01 & 0.74 \pm 0.01 & \bfseries 0.35 \pm 0.08 & 0.55 \pm 0.01 & \bfseries 0.26 \pm 0.01 & \color{blue}\bfseries 0.65 \pm 0.01 & \bfseries 0.53 \pm 0.01 & \bfseries 0.41 \pm 0.02 \\
 & Episode cost & \color{blue}\bfseries 0.88 \pm 0.04 & 1.23 \pm 0.08 & \bfseries 0.27 \pm 0.16 & 1.06 \pm 0.07 & \bfseries 0.32 \pm 0.01 & \color{blue}\bfseries 0.97 \pm 0.05 & \bfseries 0.78 \pm 0.03 & \bfseries 0.44 \pm 0.05 \\
\cmidrule(lr){1-10}
\multirow{2}{*}{PointGoal2} & Episode return & \color{blue}\bfseries 0.51 \pm 0.02 & 0.73 \pm 0.02 & \bfseries 0.39 \pm 0.05 & \bfseries 0.17 \pm 0.02 & \bfseries 0.24 \pm 0.03 & 0.50 \pm 0.00 & \bfseries 0.38 \pm 0.02 & \bfseries 0.38 \pm 0.02 \\
 & Episode cost & \color{blue}\bfseries 0.90 \pm 0.03 & 1.55 \pm 0.11 & \bfseries 0.47 \pm 0.07 & \bfseries 0.16 \pm 0.05 & \bfseries 0.23 \pm 0.03 & 1.01 \pm 0.03 & \bfseries 0.75 \pm 0.05 & \bfseries 0.68 \pm 0.02 \\
\cmidrule(lr){1-10}
\multirow{2}{*}{PointPush1} & Episode return & \color{blue}\bfseries 0.23 \pm 0.00 & 0.29 \pm 0.02 & \bfseries 0.17 \pm 0.05 & \bfseries 0.06 \pm 0.01 & \bfseries 0.16 \pm 0.01 & \bfseries 0.18 \pm 0.01 & \bfseries 0.18 \pm 0.01 & \bfseries 0.19 \pm 0.01 \\
 & Episode cost & \color{blue}\bfseries 0.72 \pm 0.06 & 1.15 \pm 0.18 & \bfseries 0.54 \pm 0.20 & \bfseries 0.09 \pm 0.02 & \bfseries 0.35 \pm 0.03 & \bfseries 0.56 \pm 0.04 & \bfseries 0.51 \pm 0.04 & \bfseries 0.57 \pm 0.08 \\
\cmidrule(lr){1-10}
\multirow{2}{*}{PointPush2} & Episode return & \bfseries 0.13 \pm 0.01 & 0.23 \pm 0.03 & 0.13 \pm 0.02 & \bfseries 0.09 \pm 0.01 & \bfseries 0.13 \pm 0.02 & \bfseries 0.14 \pm 0.01 & \color{blue}\bfseries 0.16 \pm 0.01 & \color{blue}\bfseries 0.15 \pm 0.03 \\
 & Episode cost & \bfseries 0.85 \pm 0.04 & 1.28 \pm 0.22 & 1.32 \pm 0.38 & \bfseries 0.49 \pm 0.05 & \bfseries 0.74 \pm 0.05 & \bfseries 0.91 \pm 0.06 & \color{blue}\bfseries 0.94 \pm 0.03 & \color{blue}\bfseries 0.96 \pm 0.10 \\
\cmidrule(lr){1-10}
\multirow{2}{*}{CarButton1} & Episode return & \bfseries -0.07 \pm 0.02 & 0.24 \pm 0.03 & \color{blue}\bfseries 0.11 \pm 0.03 & \bfseries -0.16 \pm 0.07 & \bfseries -0.15 \pm 0.05 & \bfseries -0.05 \pm 0.03 & \bfseries 0.06 \pm 0.02 & \color{blue}\bfseries 0.13 \pm 0.02 \\
 & Episode cost & \bfseries 0.53 \pm 0.04 & 1.83 \pm 0.15 & \color{blue}\bfseries 0.78 \pm 0.17 & \bfseries 0.02 \pm 0.00 & \bfseries 0.05 \pm 0.01 & \bfseries 0.55 \pm 0.07 & \bfseries 0.59 \pm 0.02 & \color{blue}\bfseries 0.81 \pm 0.09 \\
\cmidrule(lr){1-10}
\multirow{2}{*}{CarButton2} & Episode return & \bfseries -0.09 \pm 0.02 & 0.25 \pm 0.04 & \bfseries -0.11 \pm 0.06 & \bfseries -0.13 \pm 0.04 & \bfseries -0.08 \pm 0.01 & \bfseries -0.16 \pm 0.04 & \bfseries 0.01 \pm 0.02 & \color{blue}\bfseries 0.08 \pm 0.02 \\
 & Episode cost & \bfseries 0.56 \pm 0.08 & 2.17 \pm 0.16 & \bfseries 0.55 \pm 0.09 & \bfseries 0.06 \pm 0.01 & \bfseries 0.10 \pm 0.03 & \bfseries 0.39 \pm 0.06 & \bfseries 0.60 \pm 0.05 & \color{blue}\bfseries 0.86 \pm 0.13 \\
\cmidrule(lr){1-10}
\multirow{2}{*}{CarCircle1} & Episode return & 0.71 \pm 0.00 & 0.74 \pm 0.02 & 0.56 \pm 0.01 & \bfseries 0.49 \pm 0.01 & \color{blue}\bfseries 0.56 \pm 0.01 & 0.71 \pm 0.01 & 0.62 \pm 0.01 & 0.56 \pm 0.03 \\
 & Episode cost & 2.43 \pm 0.06 & 1.94 \pm 0.12 & 1.31 \pm 0.15 & \bfseries 0.34 \pm 0.03 & \color{blue}\bfseries 0.97 \pm 0.04 & 2.33 \pm 0.06 & 2.31 \pm 0.13 & 1.37 \pm 0.17 \\
\cmidrule(lr){1-10}
\multirow{2}{*}{CarCircle2} & Episode return & 0.73 \pm 0.01 & 0.68 \pm 0.01 & 0.47 \pm 0.05 & \color{blue}\bfseries 0.43 \pm 0.01 & 0.56 \pm 0.01 & 0.71 \pm 0.02 & 0.70 \pm 0.01 & 0.63 \pm 0.01 \\
 & Episode cost & 3.03 \pm 0.16 & 2.04 \pm 0.24 & 1.25 \pm 0.13 & \color{blue}\bfseries 0.38 \pm 0.04 & 1.39 \pm 0.07 & 2.99 \pm 0.15 & 2.74 \pm 0.13 & 2.01 \pm 0.13 \\
\cmidrule(lr){1-10}
\multirow{2}{*}{CarGoal1} & Episode return & \color{blue}\bfseries 0.43 \pm 0.01 & 0.71 \pm 0.01 & \bfseries 0.19 \pm 0.08 & \bfseries 0.12 \pm 0.01 & \bfseries 0.25 \pm 0.01 & \bfseries 0.36 \pm 0.02 & \bfseries 0.34 \pm 0.02 & \bfseries 0.31 \pm 0.01 \\
 & Episode cost & \color{blue}\bfseries 0.60 \pm 0.04 & 1.44 \pm 0.07 & \bfseries 0.23 \pm 0.09 & \bfseries 0.01 \pm 0.00 & \bfseries 0.12 \pm 0.02 & \bfseries 0.46 \pm 0.02 & \bfseries 0.41 \pm 0.05 & \bfseries 0.32 \pm 0.02 \\
\cmidrule(lr){1-10}
\multirow{2}{*}{CarGoal2} & Episode return & \color{blue}\bfseries 0.25 \pm 0.01 & 0.65 \pm 0.01 & \color{blue}\bfseries 0.21 \pm 0.06 & \bfseries -0.02 \pm 0.01 & \bfseries 0.06 \pm 0.01 & \bfseries 0.24 \pm 0.01 & \bfseries 0.23 \pm 0.01 & \color{blue}\bfseries 0.26 \pm 0.01 \\
 & Episode cost & \color{blue}\bfseries 0.85 \pm 0.05 & 2.80 \pm 0.30 & \color{blue}\bfseries 0.76 \pm 0.21 & \bfseries 0.04 \pm 0.01 & \bfseries 0.15 \pm 0.02 & \bfseries 0.85 \pm 0.05 & \bfseries 0.83 \pm 0.08 & \color{blue}\bfseries 0.90 \pm 0.09 \\
\cmidrule(lr){1-10}
\multirow{2}{*}{CarPush1} & Episode return & \bfseries 0.20 \pm 0.01 & \color{blue}\bfseries 0.32 \pm 0.02 & \bfseries 0.12 \pm 0.04 & \bfseries 0.12 \pm 0.01 & \bfseries 0.15 \pm 0.00 & \bfseries 0.19 \pm 0.01 & \bfseries 0.20 \pm 0.01 & \bfseries 0.21 \pm 0.00 \\
 & Episode cost & \bfseries 0.41 \pm 0.04 & \color{blue}\bfseries 0.79 \pm 0.04 & \bfseries 0.24 \pm 0.19 & \bfseries 0.14 \pm 0.03 & \bfseries 0.18 \pm 0.03 & \bfseries 0.35 \pm 0.04 & \bfseries 0.47 \pm 0.06 & \bfseries 0.30 \pm 0.02 \\
\cmidrule(lr){1-10}
\multirow{2}{*}{CarPush2} & Episode return & \bfseries 0.09 \pm 0.01 & 0.19 \pm 0.01 & \bfseries 0.03 \pm 0.03 & \bfseries -0.00 \pm 0.01 & \bfseries 0.05 \pm 0.01 & \bfseries 0.10 \pm 0.01 & \bfseries 0.10 \pm 0.01 & \color{blue}\bfseries 0.12 \pm 0.01 \\
 & Episode cost & \bfseries 0.99 \pm 0.10 & 1.96 \pm 0.17 & \bfseries 0.74 \pm 0.19 & \bfseries 0.08 \pm 0.01 & \bfseries 0.17 \pm 0.03 & \bfseries 0.97 \pm 0.05 & \bfseries 0.89 \pm 0.06 & \color{blue}\bfseries 0.93 \pm 0.10 \\
\cmidrule(lr){1-10}
\multirow{2}{*}{SwimmerVelocity} & Episode return & 0.36 \pm 0.03 & \color{blue}\bfseries 0.69 \pm 0.00 & \bfseries 0.58 \pm 0.01 & -0.05 \pm 0.01 & \bfseries 0.22 \pm 0.04 & 0.38 \pm 0.06 & 0.44 \pm 0.04 & \bfseries 0.62 \pm 0.01 \\
 & Episode cost & 2.95 \pm 0.49 & \color{blue}\bfseries 0.88 \pm 0.00 & \bfseries 0.01 \pm 0.01 & 1.34 \pm 0.73 & \bfseries 0.16 \pm 0.09 & 2.80 \pm 0.80 & 3.26 \pm 0.46 & \bfseries 0.65 \pm 0.28 \\
\cmidrule(lr){1-10}
\multirow{2}{*}{HopperVelocity} & Episode return & 0.25 \pm 0.12 & 0.72 \pm 0.25 & \color{blue}\bfseries 0.63 \pm 0.07 & \bfseries 0.13 \pm 0.01 & \bfseries 0.47 \pm 0.03 & \bfseries 0.26 \pm 0.09 & 0.15 \pm 0.02 & \bfseries 0.47 \pm 0.07 \\
 & Episode cost & 1.16 \pm 0.45 & 1.11 \pm 0.10 & \color{blue}\bfseries 0.70 \pm 0.36 & \bfseries 0.05 \pm 0.03 & \bfseries 0.85 \pm 0.09 & \bfseries 0.98 \pm 0.31 & 2.55 \pm 0.33 & \bfseries 0.60 \pm 0.12 \\
\cmidrule(lr){1-10}
\multirow{2}{*}{HalfCheetahVelocity} & Episode return & \bfseries 0.83 \pm 0.04 & \color{blue}\bfseries 0.98 \pm 0.01 & 0.97 \pm 0.01 & \bfseries 0.87 \pm 0.00 & \bfseries 0.76 \pm 0.02 & \bfseries 0.77 \pm 0.08 & 0.94 \pm 0.01 & \bfseries 0.76 \pm 0.01 \\
 & Episode cost & \bfseries 0.78 \pm 0.20 & \color{blue}\bfseries 0.38 \pm 0.15 & 2.93 \pm 1.36 & \bfseries 0.00 \pm 0.00 & \bfseries 0.00 \pm 0.00 & \bfseries 0.82 \pm 0.12 & 4.80 \pm 0.69 & \bfseries 0.45 \pm 0.12 \\
\cmidrule(lr){1-10}
\multirow{2}{*}{Walker2dVelocity} & Episode return & \bfseries 0.78 \pm 0.00 & \bfseries 0.68 \pm 0.07 & \color{blue}\bfseries 0.79 \pm 0.00 & \bfseries 0.34 \pm 0.05 & \bfseries 0.75 \pm 0.01 & \bfseries 0.77 \pm 0.01 & \bfseries 0.77 \pm 0.01 & \color{blue}\bfseries 0.79 \pm 0.01 \\
 & Episode cost & \bfseries 0.69 \pm 0.30 & \bfseries 0.75 \pm 0.24 & \color{blue}\bfseries 0.00 \pm 0.00 & \bfseries 0.35 \pm 0.05 & \bfseries 0.33 \pm 0.10 & \bfseries 0.40 \pm 0.21 & \bfseries 0.57 \pm 0.22 & \color{blue}\bfseries 0.07 \pm 0.02 \\
\cmidrule(lr){1-10}
\multirow{2}{*}{AntVelocity} & Episode return & 0.98 \pm 0.00 & 0.99 \pm 0.01 & \color{blue}\bfseries 0.98 \pm 0.00 & \bfseries 0.87 \pm 0.00 & \bfseries 0.93 \pm 0.00 & 0.98 \pm 0.00 & 0.98 \pm 0.01 & \bfseries 0.96 \pm 0.00 \\
 & Episode cost & 3.93 \pm 0.35 & 1.11 \pm 0.08 & \color{blue}\bfseries 0.17 \pm 0.03 & \bfseries 0.00 \pm 0.00 & \bfseries 0.03 \pm 0.01 & 3.53 \pm 0.66 & 1.91 \pm 0.35 & \bfseries 0.15 \pm 0.03 \\
\cmidrule(lr){1-10}
\multirow{2}{*}{BallRun} & Episode return & 0.65 \pm 0.02 & \color{blue}\bfseries 0.53 \pm 0.01 & \bfseries 0.31 \pm 0.02 & \bfseries 0.07 \pm 0.07 & \bfseries 0.08 \pm 0.02 & 0.54 \pm 0.02 & 1.11 \pm 0.03 & \bfseries 0.41 \pm 0.04 \\
 & Episode cost & 1.40 \pm 0.04 & \color{blue}\bfseries 0.99 \pm 0.01 & \bfseries 0.39 \pm 0.08 & \bfseries 0.13 \pm 0.09 & \bfseries 0.08 \pm 0.03 & 1.17 \pm 0.05 & 2.18 \pm 0.03 & \bfseries 0.86 \pm 0.19 \\
\cmidrule(lr){1-10}
\multirow{2}{*}{CarRun} & Episode return & \bfseries 0.97 \pm 0.00 & \color{blue}\bfseries 0.99 \pm 0.00 & \bfseries 0.97 \pm 0.00 & \bfseries 0.54 \pm 0.11 & \bfseries -0.07 \pm 0.09 & \bfseries 0.97 \pm 0.00 & \bfseries 0.96 \pm 0.00 & \bfseries 0.97 \pm 0.01 \\
 & Episode cost & \bfseries 0.02 \pm 0.01 & \color{blue}\bfseries 0.92 \pm 0.04 & \bfseries 0.00 \pm 0.00 & \bfseries 0.19 \pm 0.12 & \bfseries 0.02 \pm 0.01 & \bfseries 0.01 \pm 0.01 & \bfseries 0.09 \pm 0.04 & \bfseries 0.79 \pm 0.30 \\
\cmidrule(lr){1-10}
\multirow{2}{*}{DroneRun} & Episode return & 0.67 \pm 0.01 & 0.64 \pm 0.02 & \bfseries 0.55 \pm 0.01 & \bfseries 0.27 \pm 0.03 & \bfseries 0.34 \pm 0.04 & 0.58 \pm 0.04 & 0.64 \pm 0.02 & \color{blue}\bfseries 0.60 \pm 0.01 \\
 & Episode cost & 1.82 \pm 0.07 & 1.12 \pm 0.16 & \bfseries 0.00 \pm 0.00 & \bfseries 0.22 \pm 0.06 & \bfseries 0.62 \pm 0.09 & 1.49 \pm 0.12 & 1.13 \pm 0.34 & \color{blue}\bfseries 0.68 \pm 0.20 \\
\cmidrule(lr){1-10}
\multirow{2}{*}{AntRun} & Episode return & 0.75 \pm 0.01 & \color{blue}\bfseries 0.71 \pm 0.02 & \bfseries 0.63 \pm 0.04 & \bfseries 0.33 \pm 0.01 & \bfseries 0.68 \pm 0.01 & 0.72 \pm 0.01 & 0.86 \pm 0.01 & \bfseries 0.55 \pm 0.03 \\
 & Episode cost & 1.88 \pm 0.04 & \color{blue}\bfseries 0.93 \pm 0.10 & \bfseries 0.45 \pm 0.03 & \bfseries 0.02 \pm 0.00 & \bfseries 0.38 \pm 0.05 & 1.20 \pm 0.11 & 2.99 \pm 0.10 & \bfseries 0.46 \pm 0.13 \\
\cmidrule(lr){1-10}
\multirow{2}{*}{BallCircle} & Episode return & 0.76 \pm 0.01 & \color{blue}\bfseries 0.85 \pm 0.00 & \bfseries 0.74 \pm 0.02 & \bfseries 0.26 \pm 0.01 & \bfseries 0.16 \pm 0.01 & \bfseries 0.73 \pm 0.02 & \bfseries 0.76 \pm 0.01 & \bfseries 0.65 \pm 0.02 \\
 & Episode cost & 1.00 \pm 0.05 & \color{blue}\bfseries 0.99 \pm 0.02 & \bfseries 0.63 \pm 0.05 & \bfseries 0.00 \pm 0.00 & \bfseries 0.00 \pm 0.00 & \bfseries 0.94 \pm 0.05 & \bfseries 0.96 \pm 0.04 & \bfseries 0.54 \pm 0.05 \\
\cmidrule(lr){1-10}
\multirow{2}{*}{CarCircle} & Episode return & 0.65 \pm 0.02 & 0.82 \pm 0.01 & 0.69 \pm 0.05 & \bfseries 0.21 \pm 0.02 & \bfseries 0.02 \pm 0.01 & 0.67 \pm 0.01 & 0.85 \pm 0.01 & \color{blue}\bfseries 0.70 \pm 0.01 \\
 & Episode cost & 1.57 \pm 0.06 & 1.22 \pm 0.06 & 1.26 \pm 0.45 & \bfseries 0.02 \pm 0.01 & \bfseries 0.03 \pm 0.02 & 1.40 \pm 0.04 & 2.02 \pm 0.07 & \color{blue}\bfseries 0.82 \pm 0.03 \\
\cmidrule(lr){1-10}
\multirow{2}{*}{DroneCircle} & Episode return & 0.70 \pm 0.01 & 0.68 \pm 0.00 & \color{blue}\bfseries 0.62 \pm 0.03 & \bfseries 0.35 \pm 0.01 & \bfseries 0.40 \pm 0.01 & 0.69 \pm 0.02 & 0.68 \pm 0.01 & \bfseries 0.54 \pm 0.01 \\
 & Episode cost & 1.06 \pm 0.04 & 1.00 \pm 0.02 & \color{blue}\bfseries 0.86 \pm 0.06 & \bfseries 0.00 \pm 0.00 & \bfseries 0.05 \pm 0.01 & 1.09 \pm 0.08 & 1.07 \pm 0.05 & \bfseries 0.80 \pm 0.04 \\
\cmidrule(lr){1-10}
\multirow{2}{*}{AntCircle} & Episode return & 0.60 \pm 0.02 & 0.60 \pm 0.02 & \color{blue}\bfseries 0.41 \pm 0.02 & \bfseries 0.11 \pm 0.01 & \bfseries 0.11 \pm 0.01 & 0.60 \pm 0.02 & 0.49 \pm 0.01 & \bfseries 0.35 \pm 0.01 \\
 & Episode cost & 2.05 \pm 0.08 & 1.94 \pm 0.17 & \color{blue}\bfseries 0.53 \pm 0.13 & \bfseries 0.00 \pm 0.00 & \bfseries 0.00 \pm 0.00 & 2.05 \pm 0.12 & 1.33 \pm 0.06 & \bfseries 0.48 \pm 0.06 \\
\cmidrule(lr){1-10}
\multirow{2}{*}{MetaDrive-easysparse} & Episode return & \bfseries 0.65 \pm 0.09 & \color{blue}\bfseries 0.99 \pm 0.01 & \bfseries 0.18 \pm 0.15 & \bfseries 0.35 \pm 0.02 & \bfseries 0.51 \pm 0.01 & \bfseries 0.44 \pm 0.07 & 0.88 \pm 0.06 & \bfseries 0.56 \pm 0.09 \\
 & Episode cost & \bfseries 0.94 \pm 0.26 & \color{blue}\bfseries 0.02 \pm 0.01 & \bfseries 0.03 \pm 0.00 & \bfseries 0.09 \pm 0.04 & \bfseries 0.29 \pm 0.05 & \bfseries 0.43 \pm 0.13 & 1.77 \pm 0.22 & \bfseries 0.40 \pm 0.20 \\
\cmidrule(lr){1-10}
\multirow{2}{*}{MetaDrive-easymean} & Episode return & \bfseries 0.53 \pm 0.10 & \color{blue}\bfseries 0.72 \pm 0.05 & \bfseries 0.19 \pm 0.06 & \bfseries 0.34 \pm 0.02 & \bfseries 0.47 \pm 0.02 & 0.71 \pm 0.08 & 0.91 \pm 0.02 & \color{blue}\bfseries 0.73 \pm 0.01 \\
 & Episode cost & \bfseries 0.69 \pm 0.24 & \color{blue}\bfseries 0.05 \pm 0.01 & \bfseries 0.03 \pm 0.00 & \bfseries 0.06 \pm 0.01 & \bfseries 0.44 \pm 0.09 & 1.09 \pm 0.18 & 1.80 \pm 0.07 & \color{blue}\bfseries 0.51 \pm 0.03 \\
\cmidrule(lr){1-10}
\multirow{2}{*}{MetaDrive-easydense} & Episode return & \bfseries 0.51 \pm 0.05 & \bfseries 0.48 \pm 0.03 & \color{blue}\bfseries 0.39 \pm 0.23 & \bfseries 0.39 \pm 0.02 & \bfseries 0.50 \pm 0.03 & \color{blue}\bfseries 0.60 \pm 0.05 & \color{blue}\bfseries 0.56 \pm 0.08 & \bfseries 0.56 \pm 0.04 \\
 & Episode cost & \bfseries 0.69 \pm 0.16 & \bfseries 0.04 \pm 0.00 & \color{blue}\bfseries 0.02 \pm 0.00 & \bfseries 0.04 \pm 0.01 & \bfseries 0.41 \pm 0.14 & \color{blue}\bfseries 0.79 \pm 0.16 & \color{blue}\bfseries 0.43 \pm 0.14 & \bfseries 0.25 \pm 0.04 \\
\cmidrule(lr){1-10}
\multirow{2}{*}{MetaDrive-mediumsparse} & Episode return & 0.82 \pm 0.04 & \bfseries 0.21 \pm 0.06 & \bfseries 0.32 \pm 0.16 & \bfseries 0.43 \pm 0.03 & \bfseries 0.65 \pm 0.06 & \bfseries 0.62 \pm 0.06 & 0.78 \pm 0.04 & \color{blue}\bfseries 0.74 \pm 0.09 \\
 & Episode cost & 1.11 \pm 0.17 & \bfseries 0.05 \pm 0.00 & \bfseries 0.05 \pm 0.00 & \bfseries 0.01 \pm 0.00 & \bfseries 0.54 \pm 0.16 & \bfseries 0.68 \pm 0.20 & 2.02 \pm 0.14 & \color{blue}\bfseries 0.88 \pm 0.15 \\
\cmidrule(lr){1-10}
\multirow{2}{*}{MetaDrive-mediummean} & Episode return & 0.78 \pm 0.05 & \bfseries 0.23 \pm 0.10 & \color{blue}\bfseries 0.66 \pm 0.18 & \bfseries 0.28 \pm 0.01 & \color{blue}\bfseries 0.71 \pm 0.05 & 0.78 \pm 0.04 & 0.82 \pm 0.02 & \bfseries 0.51 \pm 0.14 \\
 & Episode cost & 1.68 \pm 0.11 & \bfseries 0.05 \pm 0.00 & \color{blue}\bfseries 0.04 \pm 0.01 & \bfseries 0.00 \pm 0.00 & \color{blue}\bfseries 0.62 \pm 0.08 & 1.43 \pm 0.23 & 2.11 \pm 0.03 & \bfseries 0.82 \pm 0.14 \\
\cmidrule(lr){1-10}
\multirow{2}{*}{MetaDrive-mediumdense} & Episode return & 0.81 \pm 0.04 & \bfseries 0.40 \pm 0.30 & \bfseries 0.77 \pm 0.06 & \bfseries 0.36 \pm 0.04 & \color{blue}\bfseries 0.87 \pm 0.01 & 0.78 \pm 0.04 & 0.81 \pm 0.08 & \bfseries 0.53 \pm 0.08 \\
 & Episode cost & 1.45 \pm 0.11 & \bfseries 0.04 \pm 0.02 & \bfseries 0.04 \pm 0.01 & \bfseries 0.02 \pm 0.01 & \color{blue}\bfseries 0.96 \pm 0.17 & 1.47 \pm 0.24 & 1.14 \pm 0.24 & \bfseries 0.73 \pm 0.14 \\
\cmidrule(lr){1-10}
\multirow{2}{*}{MetaDrive-hardsparse} & Episode return & 0.47 \pm 0.03 & \color{blue}\bfseries 0.45 \pm 0.08 & \bfseries 0.15 \pm 0.05 & \bfseries 0.31 \pm 0.01 & \bfseries 0.38 \pm 0.01 & 0.49 \pm 0.02 & 0.48 \pm 0.03 & \bfseries 0.35 \pm 0.02 \\
 & Episode cost & 1.85 \pm 0.12 & \color{blue}\bfseries 0.03 \pm 0.01 & \bfseries 0.05 \pm 0.00 & \bfseries 0.01 \pm 0.01 & \bfseries 0.65 \pm 0.07 & 1.95 \pm 0.04 & 1.97 \pm 0.09 & \bfseries 0.13 \pm 0.03 \\
\cmidrule(lr){1-10}
\multirow{2}{*}{MetaDrive-hardmean} & Episode return & 0.49 \pm 0.02 & \bfseries 0.34 \pm 0.06 & \bfseries 0.13 \pm 0.07 & \bfseries 0.25 \pm 0.01 & \bfseries 0.41 \pm 0.02 & 0.46 \pm 0.03 & 0.41 \pm 0.03 & \color{blue}\bfseries 0.47 \pm 0.01 \\
 & Episode cost & 1.90 \pm 0.12 & \bfseries 0.05 \pm 0.01 & \bfseries 0.05 \pm 0.00 & \bfseries 0.01 \pm 0.01 & \bfseries 0.91 \pm 0.14 & 1.67 \pm 0.23 & 1.99 \pm 0.10 & \color{blue}\bfseries 0.85 \pm 0.05 \\
\cmidrule(lr){1-10}
\multirow{2}{*}{MetaDrive-harddense} & Episode return & 0.33 \pm 0.03 & \bfseries 0.19 \pm 0.01 & \bfseries 0.28 \pm 0.07 & \bfseries 0.31 \pm 0.02 & \color{blue}\bfseries 0.36 \pm 0.02 & 0.38 \pm 0.02 & 0.38 \pm 0.04 & \color{blue}\bfseries 0.36 \pm 0.03 \\
 & Episode cost & 1.30 \pm 0.10 & \bfseries 0.06 \pm 0.00 & \bfseries 0.05 \pm 0.00 & \bfseries 0.05 \pm 0.03 & \color{blue}\bfseries 0.54 \pm 0.10 & 1.26 \pm 0.20 & 1.60 \pm 0.13 & \color{blue}\bfseries 0.65 \pm 0.10 \\
\bottomrule
\end{tabular}
\end{adjustbox}
\caption{Performance comparison with related-but-distinct offline settings (COptiDICE, SafeDICE, ICSDICE) and offline safe RL baselines (CDT, TraC, FISOR, FISOR-IL) on all 38 re-curated DSRL tasks. DexDICE is reported with its post hoc per-task best $\kappa$ selected using evaluation return and cost, identical to the last column of Table~\ref{app:full_dsrl_table}; this differs from the fixed-$\kappa$ main result. Results are averaged over five seeds, with standard errors reported. \textbf{Bold} marks methods satisfying the cost constraint (mean cost $\le$ 1, normalized to the cost upper bound). Among safety-feasible results, we highlight in \textcolor{blue}{\textbf{blue}} those whose error bars overlap with the highest mean episode return.}
\label{tab:combined-rl-performance}
\end{table*}

\clearpage

\subsection{Hyperparameters}
\label{sec:app.hyperparameters}

\begin{table}[hbpt!]
    \centering
    \footnotesize
    \renewcommand{\arraystretch}{1}
    \vspace{0.3cm}
    \vspace{-10pt}
    \begin{adjustbox}{max width=\textwidth}
    \begin{tabular}{l|ccccccc}
    \hline \textbf { Hyperparameters } & \textbf{BC} & \textbf{FBC} & \textbf {JFBC} & \textbf {DemoDICE-S} & \textbf{DemoDICE-U} & {\textbf{DWBC-S}} & \textbf{DexDICE (Ours)} \\
    \hline\hline
    \text { Policy distribution }                           & \multicolumn{7}{c}{{Gaussian}}\\
    \text { Batch size }                                    & \multicolumn{7}{c}{{512} } \\
    \text { Discount factor $\gamma$ }                          & \multicolumn{7}{c}{{0.99}} \\
    \hline
    \text { Network sizes ($\pi$, $\nu$, $C_\rP$, $C_\rS$) }                            & \multicolumn{7}{c}{\parbox{12cm}{\centering $[256,256]$ \quad (Safe Navigation/Velocity, re-curated DSRL tasks) \\ $[1024, 1024, 512]$ \quad (Safe Manipulation tasks)}}\\
    \text { Training iteration }                              & \multicolumn{7}{c}{{1,000,000} } \\
    \text { Evaluation frequency }                              & \multicolumn{7}{c}{{10,000} } \\
    \hline
    \text{Algorithm specific configurations} \\
    \text { \quad $\pi$ learning rate }                          & \multicolumn{7}{c}{{$3\times 10^{-5}$} } \\
    \text { \quad $\alpha$  (regularization coefficient) }                               & - & -  & - & 0.1 & 0.1 &  7.5  & - \\
     {\text { \quad $\eta$ (class prior)}} &  - &  -  &  - &  - &  - &  0.5  &  - \\
    \text { \quad $\kappa$  update frequency  }              & - & 100 & 100 & - & - & - & 100 \\
    \text { \quad $\nu$ learning rate }                     & - & - & - & $3\times 10^{-5}$ & $3\times 10^{-5}$ & - & $3\times 10^{-5}$ \\
    \text { \quad $\nu$ gradient penalty coefficient }          & - & - & - & 10 & 10 & - & 10 \\
    \text { \quad $C_\rP$ gradient penalty coefficient }          & - & - & 0 & 10 & 10 & - & 10 \\
    \text { \quad $C_\rP$ learning rate }                         & - & - & $3\times 10^{-5}$ & $3\times 10^{-5}$ & $3\times 10^{-5}$ & $3\times 10^{-5}$ & $3\times 10^{-5}$ \\
    \text { \quad $C_\rS$ gradient penalty coefficient }          & - & 0 & 0 & - & - & - & 0 \\
    \text { \quad $C_\rS$ learning rate }                         & - & $3\times 10^{-5}$ &$3\times 10^{-5}$ & - & - & - & $3\times 10^{-5}$ \\
    \hline
    \end{tabular}
    \end{adjustbox}
    \vspace{3pt}
    \caption{Summary of hyperparameters used for each method. A fixed value of $\kappa=0.01$ is used for all tasks in both the re-curated DSRL and the controlled datasets without further tuning; the per-task best $\kappa$ from the sweep $\kappa\in\{0.001, 0.005, 0.01, 0.05, 0.1, 0.2\}$ is reported only in the last column of Table~\ref{app:full_dsrl_table} and in Table~\ref{tab:combined-rl-performance}.}
    \vspace{-10pt}
    \label{tab:hyperparams}
\end{table}

\subsection{Data Efficiency Analysis}
\label{sec:detail_plot_data_efficiency}
The main paper's Figure~\ref{fig:analysis_on_data_efficiency} reports six-task averages of data efficiency and quality robustness with respect to both the safe demonstration set $\DS$ and the performant demonstration set $\DP$. The remainder of this section (data efficiency) and Section~\ref{sec:detail_plot_data_quality} (data quality) expand these aggregated curves into per-task breakdowns; Section~\ref{sec:data_efficiency_overall} additionally reports a joint $\DS$/$\DP$ scaling analysis.

\subsubsection{Data Efficiency with Joint Scaling}
\label{sec:data_efficiency_overall}
\begin{figure}[hbt!]
    \vspace{5pt}
    \centering
    \includegraphics[width=0.5\linewidth]{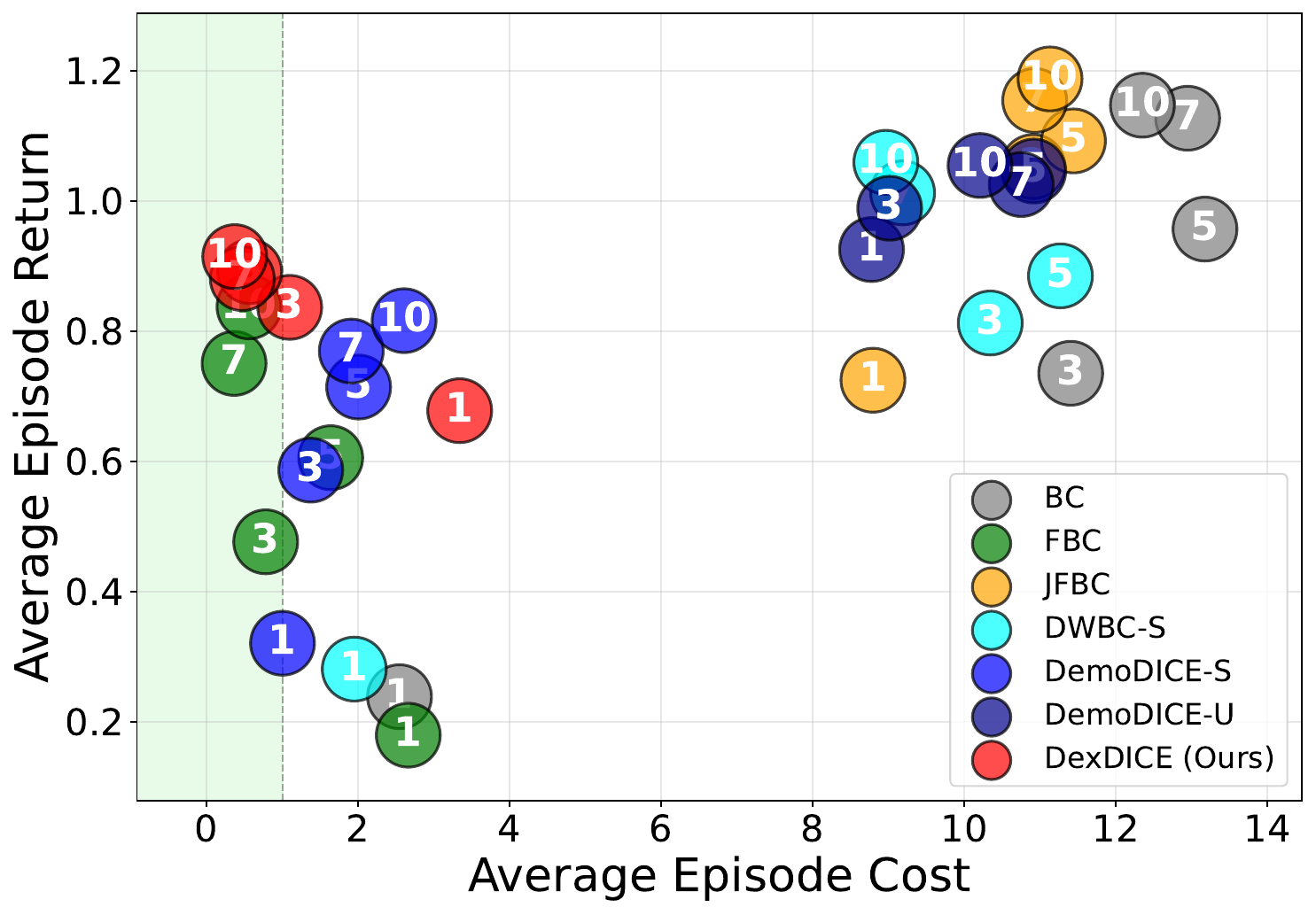}
    \caption{Overall data efficiency averaged over six tasks of each method by jointly varying $k$, where $N_{\DP}=N_{\DS}=2k$.}
    \label{fig:data_efficiency_overall}
\end{figure}

\begin{figure}[hbt!]
    \vspace{5pt}
    \centering
    \includegraphics[width=0.95\linewidth]{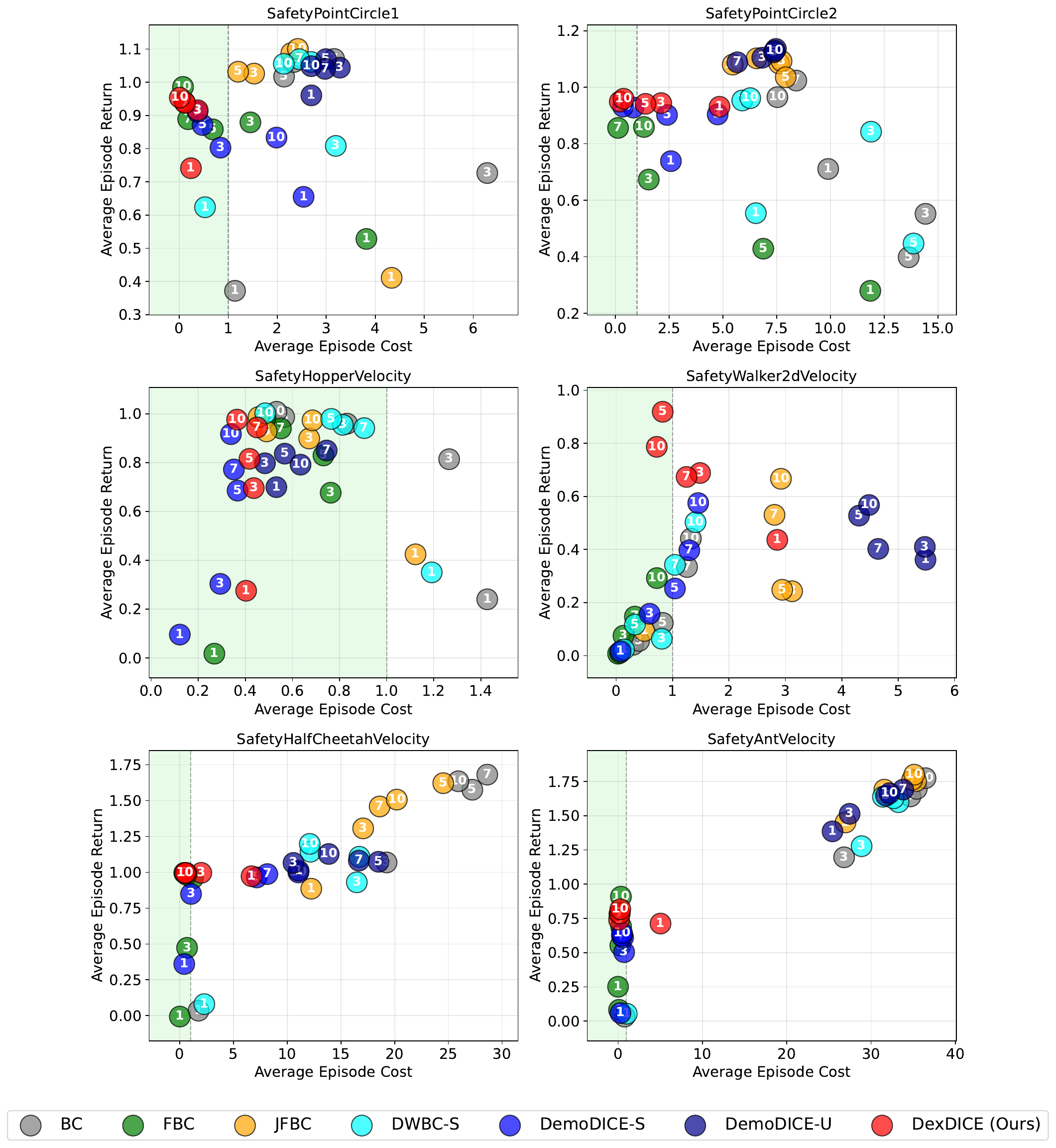}
    \caption{Per-task analysis on data efficiency with jointly varying $k$ where $N_{\DP}=N_{\DS}=2k$.}
    \label{fig:data_efficiency}
\end{figure}

In this section,
we analyze data efficiency by varying the dataset size parameter $k$, where $N_{\DS} = N_{\DP} = 2k$. Specifically, we vary the number of demonstrations as $k \in \{1, 3, 5, 7, 10\}$ while keeping $N_{\DU}=2000$. Figure~\ref{fig:data_efficiency_overall} presents the aggregated results across six benchmark tasks, providing a high-level comparison among the methods. Figure~\ref{fig:data_efficiency} reports the per-task results, illustrating how data efficiency differs across individual environments.

\clearpage
\subsubsection{Data Efficiency on $\DS$}
\label{sec:data_efficiency_on_DS}
\begin{figure}[hbt!]
    \vspace{5pt}
    \centering
    \includegraphics[width=1.0\linewidth]{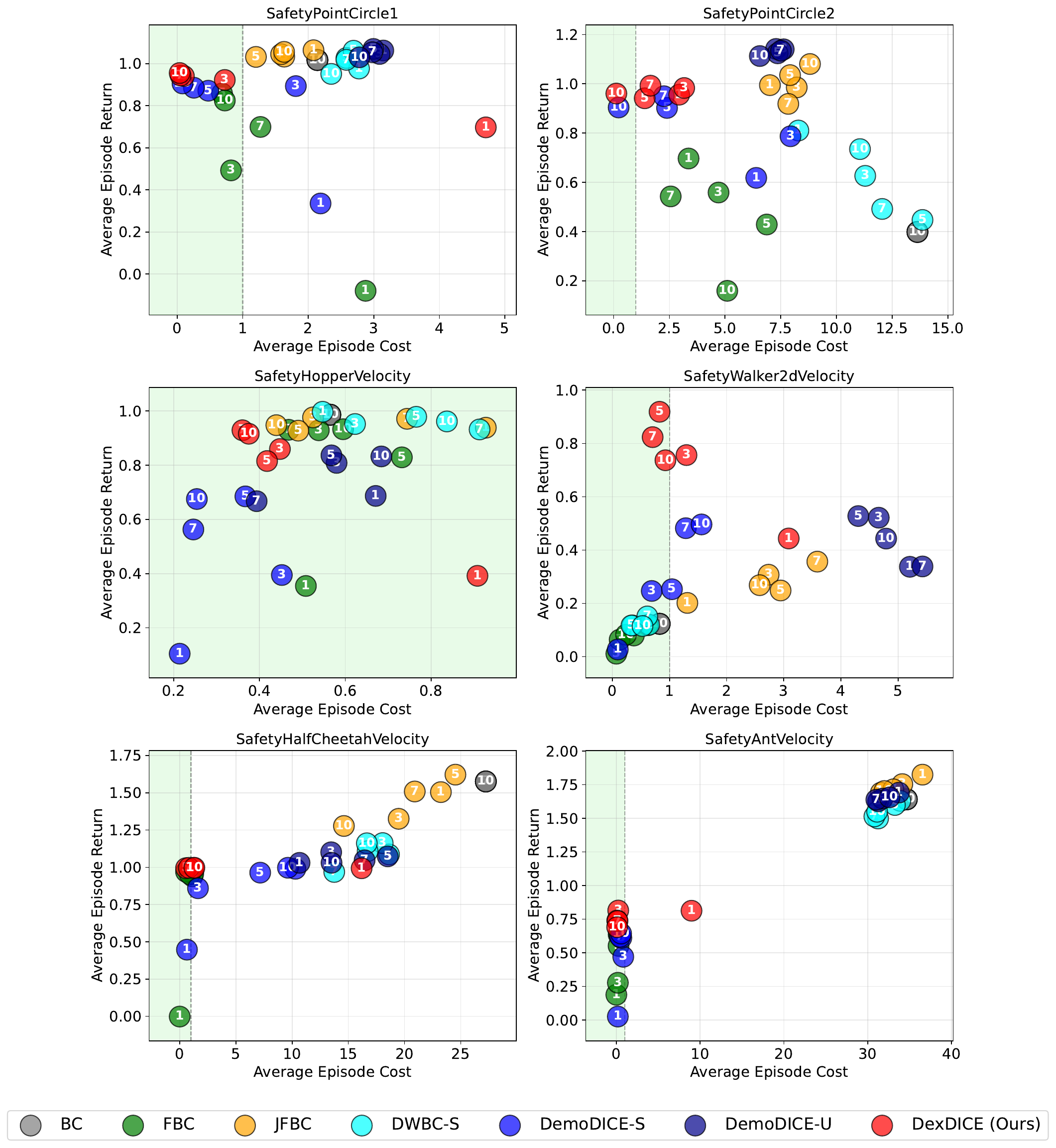}
    \caption{Per-task analysis on data efficiency of safe demonstrations $\DS$.}
    \label{fig:app_safe_data_efficiency}
\end{figure}

We sweep $N_{\DS}$ while keeping $N_{\DP} = 10$ and $N_{\DU} = 2000$ fixed. Figure~\ref{fig:data_efficiency_DS} reports the average across six tasks; Figure~\ref{fig:app_safe_data_efficiency} provides the per-task breakdown. %
\clearpage
\subsubsection{Data Efficiency on $\DP$}
\label{sec:data_efficiency_on_DP}
\begin{figure}[hbt!]
    \vspace{5pt}
    \centering
    \includegraphics[width=1.0\linewidth]{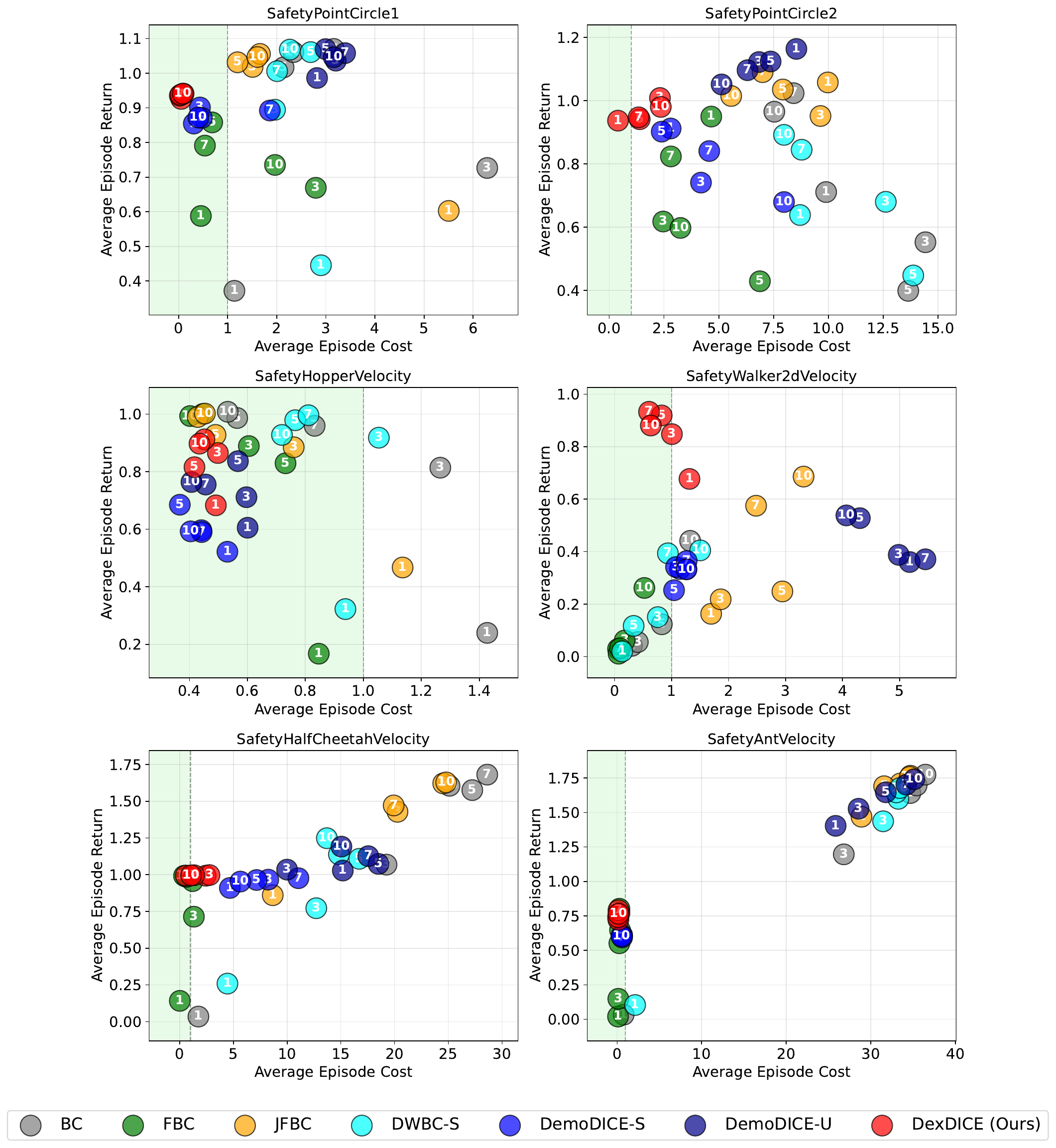}
    \caption{Per-task analysis on data efficiency of performant demonstrations $\DP$.}
    \label{fig:app_performant_data_efficiency}
\end{figure}

We sweep $N_{\DP}$ while keeping $N_{\DS} = 10$ and $N_{\DU} = 2000$ fixed. Figure~\ref{fig:data_efficiency_DP} reports the average across six tasks; Figure~\ref{fig:app_performant_data_efficiency} provides the per-task breakdown. Trends are consistent with those reported for DemoDICE-style methods: DexDICE and DemoDICE-U scale gracefully with $N_{\DP}$, while filtering-based baselines (FBC, JFBC) saturate or regress at small $\DP$.

\newpage

\subsection{Data Quality Analysis}
\label{sec:detail_plot_data_quality}
This section provides per-task breakdowns of the six-task data-quality averages reported in Section~\ref{sec:dataefficiency} (Figures~\ref{fig:data_quality_dS} and \ref{fig:data_quality_dP}). All sweeps fix $N_{\DP} = N_{\DS} = 20$ and $N_{\DU} = 2000$.

\subsubsection{Data Quality on $\DS$}
\begin{figure}[hpt!]
    \vspace{5pt}
    \centering
    \includegraphics[width=1.0\linewidth]{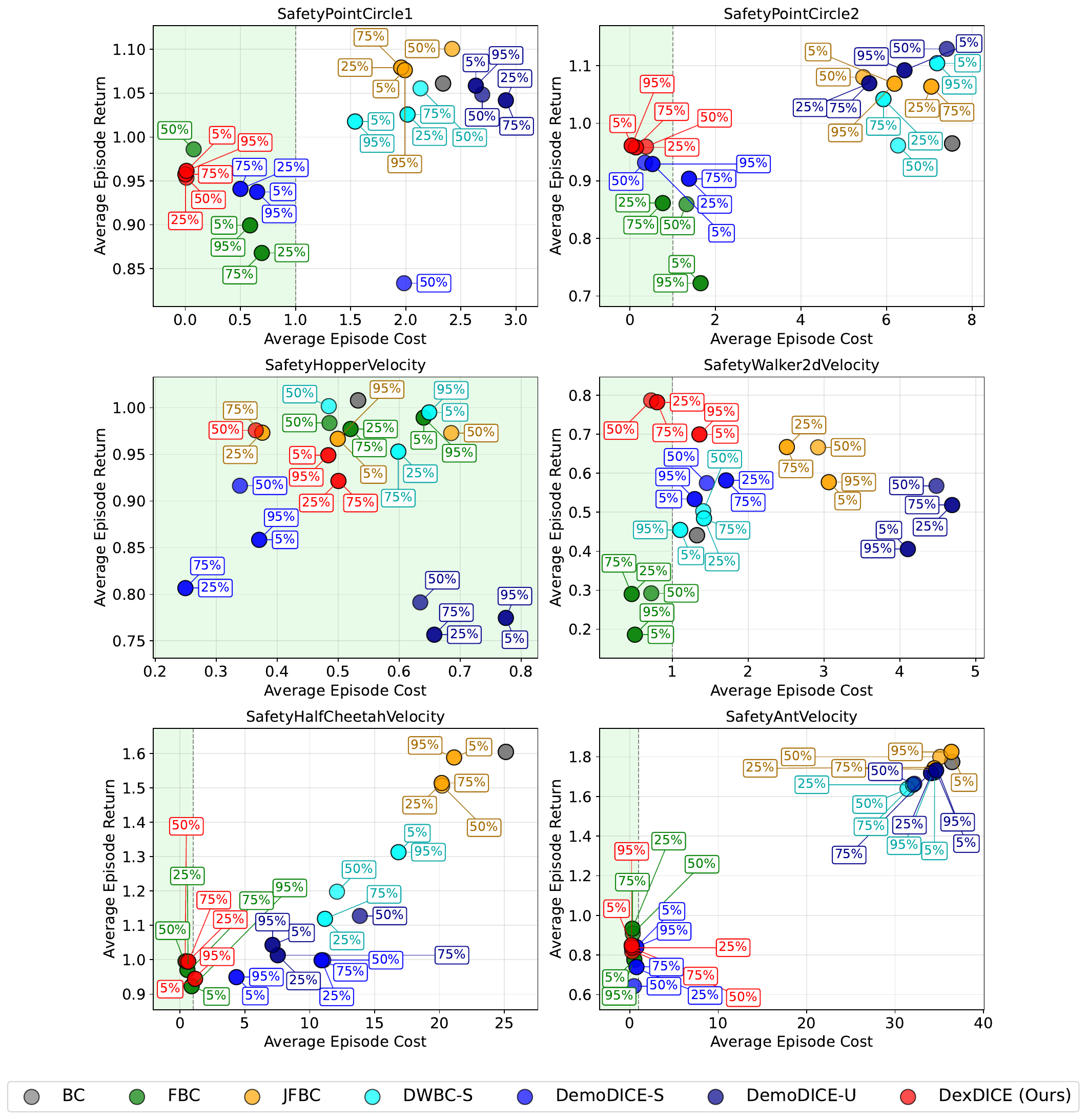}
    \caption{Per-task performance for dependencies on $\textbf{quality}(\DS)$.}
    \label{fig:app_safe_data_quality}
\end{figure}

We vary $\textbf{quality}(\DS) \in \{5\%, 25\%, 50\%, 75\%, 95\%\}$ while holding $\textbf{quality}(\DP) = 50\%$ fixed. Figure~\ref{fig:data_quality_dS} shows the average across six tasks; Figure~\ref{fig:app_safe_data_quality} provides the per-task breakdown.

\newpage

\clearpage

\subsubsection{Data Quality on $\DP$}
\label{app:data_quality_DP}
\begin{figure}[hbt!]
    \vspace{5pt}
    \centering
    \includegraphics[width=1.0\linewidth]{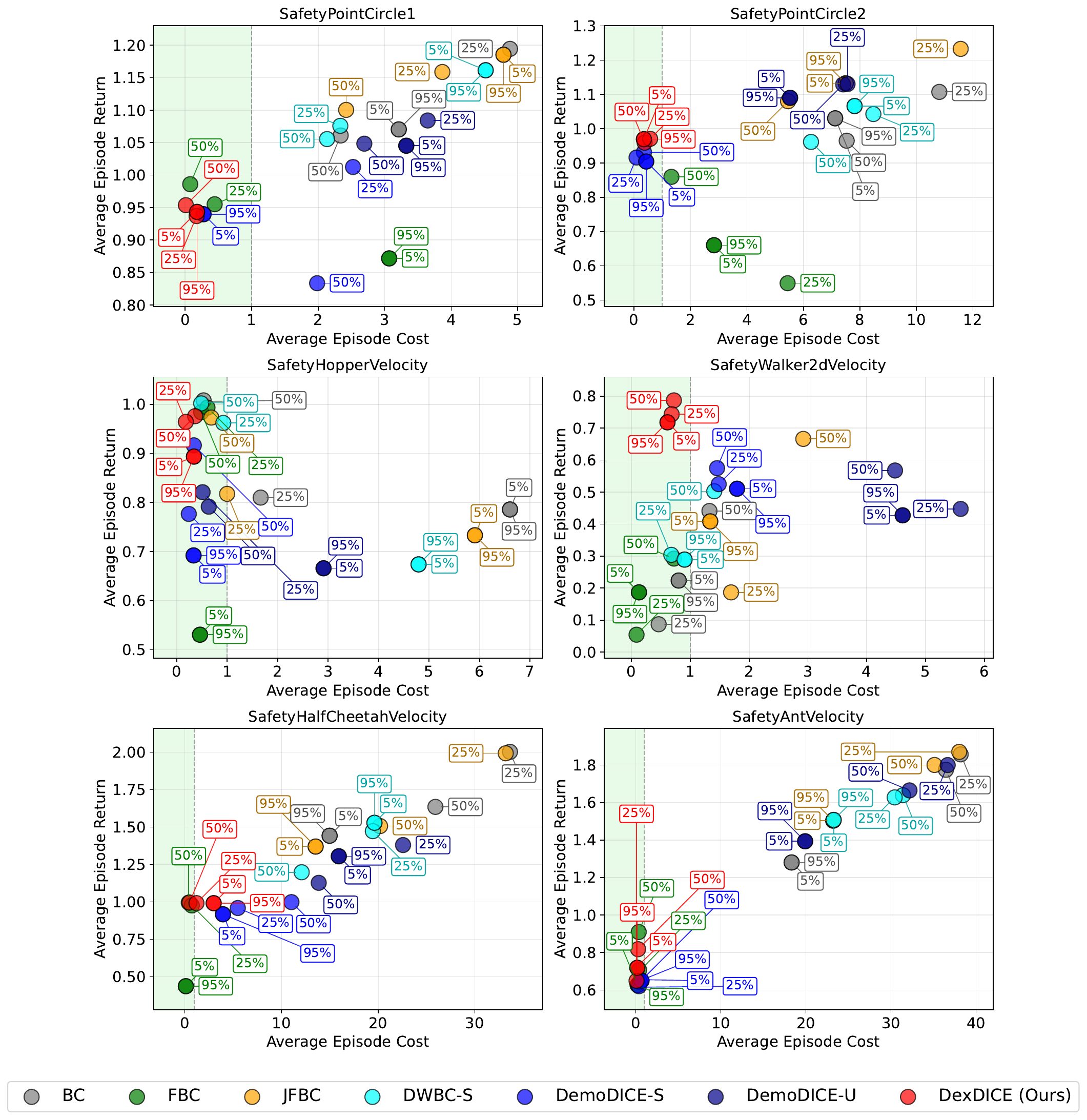}
    \caption{Per-task performance analysis for dependencies on $\textbf{quality}(\DP)$.}
    \label{fig:app_performant_data_quality}
\end{figure}

We vary $\textbf{quality}(\DP) \in \{5\%, 25\%, 50\%, 75\%, 95\%\}$ while holding $\textbf{quality}(\DS) = 50\%$ fixed. Figure~\ref{fig:data_quality_dP} shows the average across six tasks; Figure~\ref{fig:app_performant_data_quality} provides the per-task breakdown. DexDICE and DemoDICE-U track each other closely, with DexDICE additionally maintaining the cost constraint across all configurations except the most extreme low-quality regime.

\clearpage

\newpage
\subsection{Comparison on Support Estimators}
\label{sec:supp_estimation}
One of our core components is the support estimator constructed from unlabeled demonstrations proposed in Section~\ref{sec:safe_support_estimation}.
To assess effectiveness of decision functions, we compare our approach against the following established baselines:
\vspace{-7pt}
\begin{itemize}[leftmargin=5mm]
\item \textbf{GMM}: a density-estimation approach based on fitting Gaussian mixture models ~\citep{bishop2006pattern} to approximate $d_\rS$.\vspace{-5pt}
\item \textbf{LOF (Local Outlier Factor)} \citep{breunig2000lof}: a non-parametric method that detects outliers by measuring the local deviation of a sample’s density relative to that of its neighbors. \vspace{-5pt}
\item \textbf{OCSVM (One-class Support Vector Machine)} \citep{scholkopf2001estimating}: a kernel-based method that learns a hypersurface enclosing the majority of data, treating points outside the boundary as novelties.\vspace{-5pt}
\item \textbf{Deep SVDD (Deep Support Vector Data Description)} \citep{ruff2018deep}: a deep representation–based method that maps data into a latent space and minimizes the volume of a hypersphere enclosing the nominal samples.\vspace{-5pt}
\end{itemize}
\begin{table}[htbp!]
\centering
\begin{adjustbox}{width=0.75\textwidth}
\begin{tabular}%
{l|S[table-format=1.2(1.2),mode=text]
     S[table-format=1.2(1.2),mode=text]
     S[table-format=1.2(1.2),mode=text]
     S[table-format=1.2(1.2),mode=text]
     S[table-format=1.2(1.2),mode=text]}
\toprule
\textbf{Task} & \textbf{GMM} & \textbf{LOF} & \textbf{OCSVM} & \textbf{Deep SVDD} & \textbf{Ours} \\
\midrule\midrule
PointCircle1 & 0.62 \pm 0.01 & 0.46 \pm 0.00 & 0.43 \pm 0.00 & 0.56 \pm 0.00 & \bfseries 0.87 \pm 0.00 \\
PointCircle2 & 0.71 \pm 0.00 & 0.60 \pm 0.00 & 0.52 \pm 0.00 & 0.67 \pm 0.00 & \bfseries 0.80 \pm 0.00 \\
HopperVelocity & \bfseries 0.83 \pm 0.00 & 0.73 \pm 0.00 & 0.65 \pm 0.00 & 0.76 \pm 0.01 & 0.74 \pm 0.02 \\
Walker2dVelocity & 0.94 \pm 0.00 & 0.88 \pm 0.00 & 0.82 \pm 0.00 & 0.85 \pm 0.00 & \bfseries 0.95 \pm 0.00 \\
HalfCheetahVelocity & 0.56 \pm 0.00 & 0.58 \pm 0.00 & 0.49 \pm 0.00 & 0.45 \pm 0.00 & \bfseries 0.65 \pm 0.00 \\
CarCircle & 0.50 \pm 0.00 & 0.47 \pm 0.00 & 0.49 \pm 0.00 & 0.49 \pm 0.00 & \bfseries 0.72 \pm 0.00 \\
CarRun & 0.54 \pm 0.01 & 0.35 \pm 0.00 & 0.48 \pm 0.00 & 0.45 \pm 0.02 & \bfseries 0.64 \pm 0.00 \\
DroneCircle & 0.67 \pm 0.00 & 0.49 \pm 0.00 & 0.64 \pm 0.00 & 0.56 \pm 0.01 & \bfseries 0.90 \pm 0.00 \\
DroneRun & 0.77 \pm 0.00 & 0.81 \pm 0.00 & 0.35 \pm 0.00 & \bfseries 0.88 \pm 0.00 & 0.83 \pm 0.00 \\
AntCircle & 0.55 \pm 0.01 & 0.56 \pm 0.00 & 0.46 \pm 0.00 & 0.54 \pm 0.00 & \bfseries 0.69 \pm 0.00 \\
AntRun &  0.62 \pm 0.00 & 0.75 \pm 0.00 & 0.69 \pm 0.00 & 0.64 \pm 0.01 & \bfseries 0.76 \pm 0.01 \\
\bottomrule
\end{tabular}
\end{adjustbox}
\vspace{5pt}
\caption{AUROC comparison of different novelty detection methods on re-curated DSRL tasks.}
\label{tab:novelty}
\vspace{-15pt}
\end{table}
As shown in Table~\ref{tab:novelty}, our method achieves the highest AUROC on 8 of 11 tasks, indicating consistently superior novelty-detection performance. This highlights the benefit of leveraging density-ratio estimation for reliable support estimation.
We further examine how the choice of support estimator impacts safety and performance on DexDICE; the results are reported in Table~\ref{tab:safe_estimator_performance}.
\begin{table*}[hpt!]
    \centering
    \begin{adjustbox}{width=0.8\textwidth}
    \small
    \aboverulesep=0.1ex
    \belowrulesep=0.1ex
    \renewcommand{\arraystretch}{1.1}
    \begin{tabular}
    {l|l|
     S[table-format=1.2(1.2),mode=text]
     S[table-format=1.2(1.2),mode=text]
     S[table-format=1.2(1.2),mode=text]
     S[table-format=1.2(1.2),mode=text]
     S[table-format=1.2(1.2),mode=text]}
       \toprule
        \textbf{Task} & \textbf{Metric} &
        \small{\textbf{GMM}} & \small{\textbf{LOF}} &\small{\textbf{OCSVM}} &\small{\textbf{Deep SVDD}} & \small{\textbf{Ours}} \\
       \midrule\midrule
\multirow{2}{*}{PointCircle1} & Episode return & 0.64 \pm 0.03 & 0.80 \pm 0.01 & 0.79 \pm 0.01 & 0.87 \pm 0.01 & \color{blue}\bfseries 0.27 \pm 0.03 \\
 & Episode cost & 2.81 \pm 0.19 & 4.76 \pm 0.10 & 4.81 \pm 0.12 & 4.64 \pm 0.09 & \color{blue}\bfseries 0.27 \pm 0.07 \\
\cmidrule(lr){1-7}
\multirow{2}{*}{PointCircle2} & Episode return & 0.68 \pm 0.01 & \bfseries 0.03 \pm 0.12 & 0.79 \pm 0.01 & 0.73 \pm 0.03 & \color{blue}\bfseries 0.60 \pm 0.02 \\
 & Episode cost & 2.56 \pm 0.07 & \bfseries 0.65 \pm 0.37 & 3.10 \pm 0.06 & 2.67 \pm 0.19 & \color{blue}\bfseries 0.93 \pm 0.19 \\
\midrule
\multirow{2}{*}{HopperVelocity} & Episode return & \bfseries 0.13 \pm 0.06 & \bfseries 0.10 \pm 0.05 & \bfseries 0.12 \pm 0.04 & 0.20 \pm 0.06 & \color{blue}\bfseries 0.47 \pm 0.07 \\
 & Episode cost & \bfseries 0.17 \pm 0.04 & \bfseries 0.00 \pm 0.00 & \bfseries 0.63 \pm 0.20 & 2.10 \pm 0.60 & \color{blue}\bfseries 0.60 \pm 0.12 \\
\cmidrule(lr){1-7}
\multirow{2}{*}{HalfCheetahVelocity} & Episode return & 0.94 \pm 0.01 & 0.96 \pm 0.01 & 0.92 \pm 0.04 & 0.76 \pm 0.19 & \color{blue}\bfseries 0.76 \pm 0.01 \\
 & Episode cost & 1.89 \pm 0.49 & 3.09 \pm 0.91 & 1.48 \pm 0.32 & 2.02 \pm 0.60 & \color{blue}\bfseries 0.45 \pm 0.12 \\
\cmidrule(lr){1-7}
\multirow{2}{*}{Walker2dVelocity} & Episode return & \color{blue}\bfseries 0.80 \pm 0.01 & 0.88 \pm 0.01 & 0.79 \pm 0.01 & 0.81 \pm 0.01 & \bfseries 0.79 \pm 0.01 \\
 & Episode cost & \color{blue}\bfseries 0.52 \pm 0.10 & 3.38 \pm 0.16 & 1.33 \pm 0.32 & 1.72 \pm 0.44 & \bfseries 0.07 \pm 0.02 \\
\midrule
\multirow{2}{*}{CarRun} & Episode return & \color{blue}\bfseries 0.97 \pm 0.00 & \bfseries 0.97 \pm 0.00 & \color{blue}\bfseries 0.98 \pm 0.00 & \bfseries 0.97 \pm 0.00 & \color{blue}\bfseries 0.97 \pm 0.01 \\
 & Episode cost & \color{blue}\bfseries 0.37 \pm 0.22 & \bfseries 0.18 \pm 0.07 & \color{blue}\bfseries 0.41 \pm 0.27 & \bfseries 0.15 \pm 0.06 & \color{blue}\bfseries 0.79 \pm 0.30 \\
\cmidrule(lr){1-7}
\multirow{2}{*}{CarCircle} & Episode return & 0.58 \pm 0.05 & 0.83 \pm 0.01 & 0.82 \pm 0.01 & 0.87 \pm 0.01 & \color{blue}\bfseries 0.70 \pm 0.01 \\
 & Episode cost & 1.20 \pm 0.09 & 2.00 \pm 0.03 & 1.92 \pm 0.05 & 2.03 \pm 0.02 & \color{blue}\bfseries 0.82 \pm 0.03 \\
\cmidrule(lr){1-7}
\multirow{2}{*}{DroneRun} & Episode return & \color{blue}\bfseries 0.58 \pm 0.02 & 0.74 \pm 0.02 & 0.75 \pm 0.01 & 0.74 \pm 0.01 & \color{blue}\bfseries 0.60 \pm 0.01 \\
 & Episode cost & \color{blue}\bfseries 0.76 \pm 0.19 & 2.28 \pm 0.03 & 2.31 \pm 0.08 & 2.26 \pm 0.07 & \color{blue}\bfseries 0.68 \pm 0.20 \\
\cmidrule(lr){1-7}
\multirow{2}{*}{DroneCircle} & Episode return & \color{blue}\bfseries 0.54 \pm 0.02 & 0.69 \pm 0.01 & 0.76 \pm 0.01 & 0.77 \pm 0.01 & \color{blue}\bfseries 0.54 \pm 0.01 \\
 & Episode cost & \color{blue}\bfseries 0.89 \pm 0.03 & 1.62 \pm 0.06 & 1.53 \pm 0.06 & 1.53 \pm 0.05 & \color{blue}\bfseries 0.80 \pm 0.04 \\
\cmidrule(lr){1-7}
\multirow{2}{*}{AntRun} & Episode return & \color{blue}\bfseries 0.60 \pm 0.01 & 0.75 \pm 0.01 & 0.78 \pm 0.01 & 0.76 \pm 0.02 & \bfseries 0.55 \pm 0.03 \\
 & Episode cost & \color{blue}\bfseries 0.23 \pm 0.05 & 1.81 \pm 0.09 & 2.03 \pm 0.14 & 1.77 \pm 0.20 & \bfseries 0.46 \pm 0.13 \\
\cmidrule(lr){1-7}
\multirow{2}{*}{AntCircle} & Episode return & \bfseries 0.32 \pm 0.01 & 0.51 \pm 0.01 & 0.58 \pm 0.01 & 0.58 \pm 0.01 & \color{blue}\bfseries 0.35 \pm 0.01 \\
 & Episode cost & \bfseries 0.76 \pm 0.08 & 1.14 \pm 0.06 & 1.82 \pm 0.05 & 1.80 \pm 0.09 & \color{blue}\bfseries 0.48 \pm 0.06 \\
        \bottomrule
   \end{tabular}
    \end{adjustbox}
    \vspace{5pt}
    \caption{Safety-Performance comparison with different support estimators on re-curated DSRL tasks. Results are averaged over five seeds with standard errors. We follow the same evaluation protocol as in Table~\ref{app:full_dsrl_table}.}
    \label{tab:safe_estimator_performance}
    \vspace{-20pt}
\end{table*}

\clearpage

\subsection{Sensitivity Analysis on Target FNR $\kappa$}
\label{sec:sensitivity_kappa}

\begin{figure}[hpt!]
    \centering
    \includegraphics[width=0.85\linewidth]{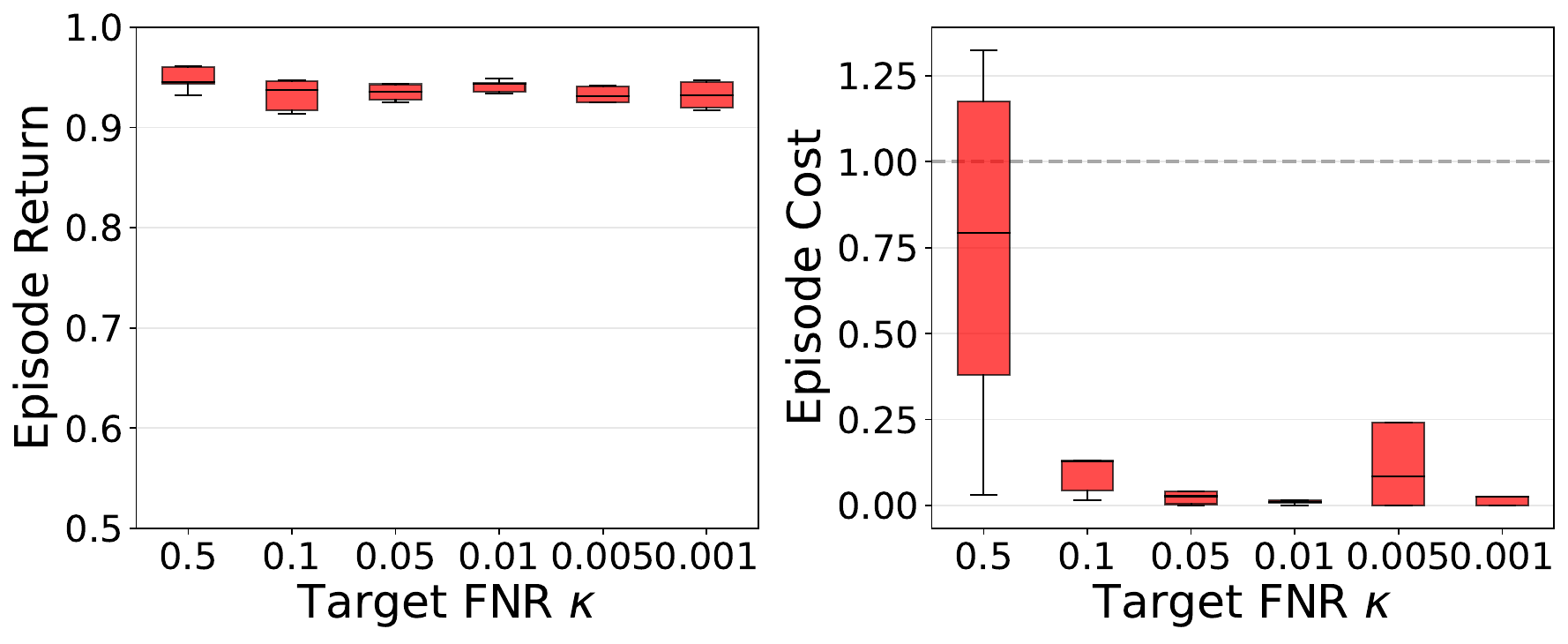}
    \caption{Sensitivity analysis of DexDICE with respect to the target FNR $\kappa$ used as the threshold for $\mathds{1}_{\dS}$. Experiments are conducted on controlled datasets of the PointCircle1 task with $N_{\DP} = N_{\DS} = 10$, over 5 random seeds.}
    \label{fig:FNR_sensitivity}
\end{figure}

The target FNR $\kappa$ determines the threshold for the estimated safe support: as $\kappa$ increases, the quantile threshold $\tau_\kappa$ rises and the set of accepted state-action pairs becomes smaller. Figure~\ref{fig:FNR_sensitivity} shows that episode return and cost remain stable across much of the tested range, with performance degrading at larger $\kappa$ values as the estimated safe support narrows.

\subsection{Robustness on Noisy Safe Demonstrations}
\label{sec:robust_to_noise}
\begin{figure}[hpt!]
    \vspace{5pt}
    \centering
    \includegraphics[width=0.99\linewidth]{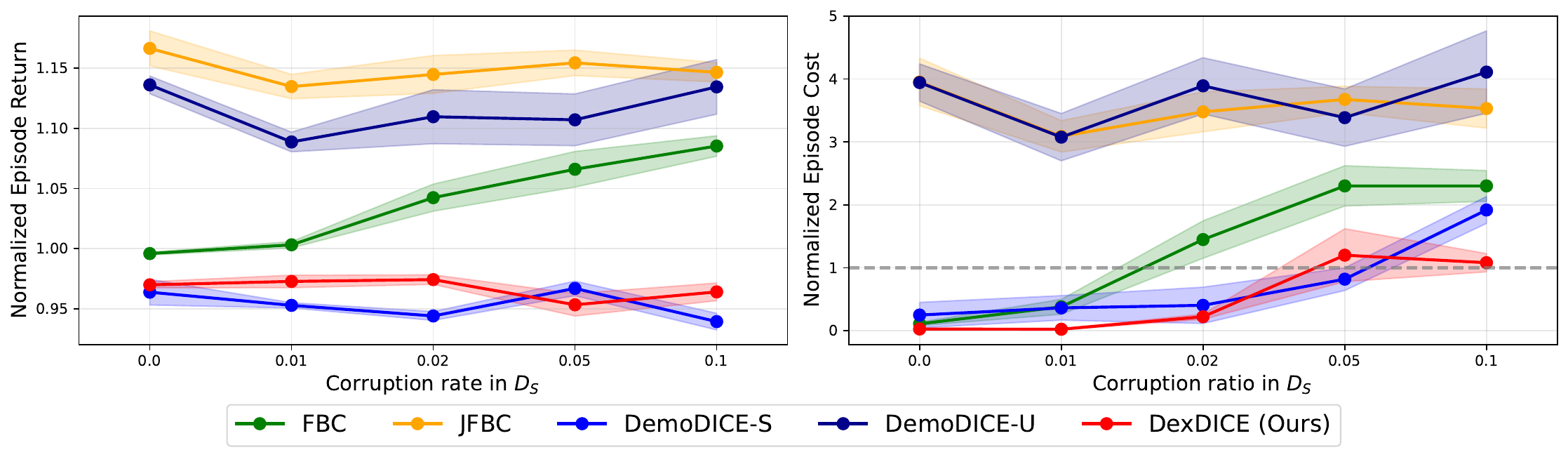}
    \caption{Stability under corruption of safe demonstrations.
We evaluate on the controlled dataset of the PointCircle1 task, using a sufficient number of demonstrations ($N_{\DP}=200$, $N_{\DS}=200$).
The x-axis indicates the corruption rate of $\DS$, i.e., the proportion of safe demonstrations replaced with unsafe demonstrations that exceed the episode cost upper bound.
Both DexDICE and DemoDICE-S maintain relatively high episode return and low episode cost even under moderate corruption,
while other methods degrade significantly as $\DS$ becomes increasingly corrupted.  }
    \label{fig:robust_nois}
\end{figure}

\rev{Figure~\ref{fig:robust_nois} evaluates DexDICE when a fraction of the safe demonstrations in $\DS$ is replaced with unsafe ones; DexDICE sustains high return and low cost under moderate corruption, while most baselines degrade sharply.}

\rev{
\subsection{Sensitivity to Dataset Construction Thresholds}
\label{sec:app.threshold_sensitivity}

The re-curated DSRL benchmark labels trajectories as safe or performant using two thresholds (Appendix~\ref{app:DSRLDatasets}): a cost threshold defining $\DS$ and a return threshold defining $\DP$. These thresholds determine, respectively, the demonstration-defined safety support and the performance target. We therefore provide a one-at-a-time sensitivity sweep over both, on six tasks with $N_{\DS}=N_{\DP}=10$ and 5 seeds, following the evaluation protocol of Table~\ref{tab:main_comparison}. (i) We vary the cost threshold defining $\DS$ from $0.5\times$ to $1.5\times$ the DSRL cost limit while holding the $\DP$ threshold at its default (Table~\ref{tab:threshold_sensitivity_DS}). (ii) We vary the return threshold defining $\DP$ over the 50th, 75th, 90th, and 95th percentiles of safe trajectory returns, plus a threshold above the maximum safe trajectory return, while holding the $\DS$ threshold at its default (Table~\ref{tab:threshold_sensitivity_DP}). These thresholds change only the dataset labels; every policy is evaluated against the same fixed normalized cost bar of $1.0$ defined by the original DSRL cost limit.

At the default settings ($1.0\times$ cost and 90th-percentile return), all six tasks remain below the cost bar. Across the full sweep, all 30 $\DS$-threshold cells and 27 of 30 $\DP$-threshold cells remain below it; the three gray entries occur only at non-default $\DP$ thresholds. Returns vary by task, but the default thresholds show no systematic degradation relative to neighboring settings.

\begin{table}[hbt!]
\centering
\small
\begin{adjustbox}{max width=\textwidth}
\aboverulesep=0.1ex
\belowrulesep=0.1ex
\renewcommand{\arraystretch}{1.1}
\begin{tabular}{l|l|cccccc}
\toprule
\textbf{Cost threshold} & \textbf{Metric} & \textbf{PointCircle1} & \textbf{DroneCircle} & \textbf{Walker2dVelocity} & \textbf{HopperVelocity} & \textbf{AntVelocity} & \textbf{HalfCheetahVelocity} \\
\midrule\midrule
\multirow{2}{*}{$0.5\times$} & Episode return & 0.26 $\pm$ 0.03 & 0.41 $\pm$ 0.04 & 0.78 $\pm$ 0.00 & 0.20 $\pm$ 0.07 & 0.93 $\pm$ 0.02 & 0.80 $\pm$ 0.02 \\
 & Episode cost & 0.12 $\pm$ 0.10 & 0.48 $\pm$ 0.04 & 0.02 $\pm$ 0.01 & 0.93 $\pm$ 0.39 & 0.10 $\pm$ 0.04 & 0.01 $\pm$ 0.00 \\
\cmidrule(lr){1-8}
\multirow{2}{*}{$0.75\times$} & Episode return & 0.46 $\pm$ 0.02 & 0.42 $\pm$ 0.04 & 0.71 $\pm$ 0.06 & 0.14 $\pm$ 0.06 & 0.96 $\pm$ 0.01 & 0.73 $\pm$ 0.05 \\
 & Episode cost & 0.10 $\pm$ 0.04 & 0.77 $\pm$ 0.02 & 0.27 $\pm$ 0.22 & 0.22 $\pm$ 0.08 & 0.17 $\pm$ 0.05 & 0.17 $\pm$ 0.16 \\
\cmidrule(lr){1-8}
\multirow{2}{*}{$1.0\times$ (default)} & Episode return & 0.27 $\pm$ 0.03 & 0.54 $\pm$ 0.01 & 0.79 $\pm$ 0.01 & 0.47 $\pm$ 0.07 & 0.96 $\pm$ 0.00 & 0.76 $\pm$ 0.01 \\
 & Episode cost & 0.27 $\pm$ 0.07 & 0.80 $\pm$ 0.04 & 0.07 $\pm$ 0.02 & 0.60 $\pm$ 0.12 & 0.15 $\pm$ 0.03 & 0.45 $\pm$ 0.12 \\
\cmidrule(lr){1-8}
\multirow{2}{*}{$1.25\times$} & Episode return & 0.31 $\pm$ 0.02 & 0.52 $\pm$ 0.02 & 0.79 $\pm$ 0.01 & 0.19 $\pm$ 0.02 & 0.96 $\pm$ 0.01 & 0.87 $\pm$ 0.01 \\
 & Episode cost & 0.18 $\pm$ 0.14 & 0.75 $\pm$ 0.06 & 0.06 $\pm$ 0.03 & 0.53 $\pm$ 0.16 & 0.32 $\pm$ 0.08 & 0.01 $\pm$ 0.01 \\
\cmidrule(lr){1-8}
\multirow{2}{*}{$1.5\times$} & Episode return & 0.34 $\pm$ 0.04 & 0.45 $\pm$ 0.04 & 0.77 $\pm$ 0.01 & 0.11 $\pm$ 0.03 & 0.98 $\pm$ 0.00 & 0.86 $\pm$ 0.00 \\
 & Episode cost & 0.43 $\pm$ 0.25 & 0.66 $\pm$ 0.04 & 0.00 $\pm$ 0.00 & 0.49 $\pm$ 0.16 & 0.24 $\pm$ 0.06 & 0.19 $\pm$ 0.06 \\
\bottomrule
\end{tabular}
\end{adjustbox}
\vspace{10pt}
\caption{Sensitivity analysis to the cost threshold used to label $\DS$, expressed as a multiple of the DSRL cost limit, with the $\DP$ threshold fixed at its default. Normalized episode return and cost are reported as 5-seed means $\pm$ standard errors. All settings satisfy the cost bar of $1.0$.}
\label{tab:threshold_sensitivity_DS}
\end{table}

\begin{table}[hbt!]
\centering
\small
\begin{adjustbox}{max width=\textwidth}
\aboverulesep=0.1ex
\belowrulesep=0.1ex
\renewcommand{\arraystretch}{1.1}
\begin{tabular}{l|l|cccccc}
\toprule
\textbf{Return threshold} & \textbf{Metric} & \textbf{PointCircle1} & \textbf{DroneCircle} & \textbf{Walker2dVelocity} & \textbf{HopperVelocity} & \textbf{AntVelocity} & \textbf{HalfCheetahVelocity} \\
\midrule\midrule
\multirow{2}{*}{50th} & Episode return & 0.44 $\pm$ 0.02 & 0.28 $\pm$ 0.02 & 0.78 $\pm$ 0.01 & 0.18 $\pm$ 0.07 & 0.86 $\pm$ 0.01 & 0.89 $\pm$ 0.01 \\
 & Episode cost & 0.22 $\pm$ 0.09 & 0.46 $\pm$ 0.10 & 0.34 $\pm$ 0.25 & 0.35 $\pm$ 0.16 & 0.21 $\pm$ 0.10 & \textcolor{gray}{1.80 $\pm$ 0.35} \\
\cmidrule(lr){1-8}
\multirow{2}{*}{75th} & Episode return & 0.47 $\pm$ 0.02 & 0.46 $\pm$ 0.05 & 0.80 $\pm$ 0.00 & 0.66 $\pm$ 0.03 & 0.88 $\pm$ 0.01 & 0.91 $\pm$ 0.01 \\
 & Episode cost & 0.13 $\pm$ 0.05 & 0.61 $\pm$ 0.07 & 0.03 $\pm$ 0.01 & 0.40 $\pm$ 0.16 & 0.07 $\pm$ 0.03 & 0.42 $\pm$ 0.11 \\
\cmidrule(lr){1-8}
\multirow{2}{*}{90th (default)} & Episode return & 0.27 $\pm$ 0.03 & 0.54 $\pm$ 0.01 & 0.79 $\pm$ 0.01 & 0.47 $\pm$ 0.07 & 0.96 $\pm$ 0.00 & 0.76 $\pm$ 0.01 \\
 & Episode cost & 0.27 $\pm$ 0.07 & 0.80 $\pm$ 0.04 & 0.07 $\pm$ 0.02 & 0.60 $\pm$ 0.12 & 0.15 $\pm$ 0.03 & 0.45 $\pm$ 0.12 \\
\cmidrule(lr){1-8}
\multirow{2}{*}{95th} & Episode return & 0.17 $\pm$ 0.04 & 0.55 $\pm$ 0.04 & 0.78 $\pm$ 0.00 & 0.34 $\pm$ 0.11 & 0.97 $\pm$ 0.00 & 0.92 $\pm$ 0.00 \\
 & Episode cost & 0.07 $\pm$ 0.02 & 0.91 $\pm$ 0.05 & 0.07 $\pm$ 0.06 & 0.70 $\pm$ 0.12 & 0.15 $\pm$ 0.04 & 0.06 $\pm$ 0.02 \\
\cmidrule(lr){1-8}
\multirow{2}{*}{$>$ max} & Episode return & 0.39 $\pm$ 0.03 & 0.56 $\pm$ 0.01 & 0.77 $\pm$ 0.01 & 0.26 $\pm$ 0.06 & 0.98 $\pm$ 0.01 & 0.69 $\pm$ 0.02 \\
 & Episode cost & 0.07 $\pm$ 0.03 & 0.78 $\pm$ 0.05 & 0.08 $\pm$ 0.05 & \textcolor{gray}{1.81 $\pm$ 0.26} & 0.47 $\pm$ 0.10 & \textcolor{gray}{1.56 $\pm$ 0.39} \\
\bottomrule
\end{tabular}
\end{adjustbox}
\vspace{10pt}
\caption{Sensitivity analysis to the return threshold used to label $\DP$, expressed as a percentile of safe-trajectory returns ($>$ max: above the maximum safe-trajectory return), with the $\DS$ threshold fixed at its default. Normalized episode return and cost are reported as 5-seed means $\pm$ standard errors. Gray values indicate costs above the bar of $1.0$; all three such cases occur at non-default thresholds.}
\label{tab:threshold_sensitivity_DP}
\end{table}

\subsection{Deployment-Time Projection onto the Estimated Safe Support}
\label{sec:app.projection}

Theorem~\ref{thm:guarantee-full} applies to the ideal support-constrained solution $\pi_\kappa$ with $\supp(d^{\pi_\kappa})\subseteq\hbS_{\tau_\kappa}$. The policy extracted in practice by weighted behavioral cloning (Appendix~\ref{app.policy_extraction}) is a Gaussian with full support over the action space and hence cannot be hard-supported on $\hbS_{\tau_\kappa}$. We quantify this gap by the \emph{support-mismatch ratio}, the fraction of executed online steps whose action lies outside $\hbS_{\tau_\kappa}$, i.e., $\hat r_\rS(s,a)<\tau_\kappa$.

To mitigate the gap, we implement a finite-budget, gradient-based action projection at deployment time. Starting from the proposed action, the filter ascends the safe-support discriminator's logit with respect to the action, clips every update to the action bounds, and terminates once $\hat r_\rS(s,a)\ge\tau_\kappa$; steps at which the search budget is exhausted are logged. To obtain smoother support-score gradients for the projection, the safe-support discriminator's gradient-penalty coefficient $C_\rS$ is raised from its default $0$ (Table~\ref{tab:hyperparams}) to $10$, matching the performance discriminator's setting $C_\rP=10$; both rows of Table~\ref{tab:projection} use this setting. Exact support inclusion under such filtering would require every reached state $s$ to admit at least one action $a$ with $(s,a)\in\hbS_{\tau_\kappa}$, which need not hold in general, and even then a finite-budget search may fail to find it. The residual mismatch after projection is therefore reported separately and is not covered by the exact guarantee. To obtain smoother support-score gradients for the projection, the safe-support discriminator's gradient-penalty coefficient $C_\rS$ is set to $10$ in this experiment (default $0$ in Table~\ref{tab:hyperparams}), matching the performance discriminator's setting $C_\rP=10$.

Table~\ref{tab:projection} reports results on six tasks with $N_{\DS}=N_{\DP}=10$ over 5 seeds. On the same checkpoints, projection reduces the task-average mismatch ratio from $14.88\%$ to $0.88\%$, with a modest reduction in mean return ($0.650\to0.623$) and a decrease in mean cost ($0.450\to0.357$).

\begin{table}[hbt!]
\centering
\small
\begin{adjustbox}{max width=\textwidth}
\aboverulesep=0.1ex
\belowrulesep=0.1ex
\renewcommand{\arraystretch}{1.1}
\begin{tabular}{l|l|cccccc}
\toprule
\textbf{Method} & \textbf{Metric} & \textbf{PointCircle1} & \textbf{DroneCircle} & \textbf{Walker2dVelocity} & \textbf{HopperVelocity} & \textbf{AntVelocity} & \textbf{HalfCheetahVelocity} \\
\midrule\midrule
\multirow{3}{*}{DexDICE ($C_\rS=10$)} & Episode return & 0.52 $\pm$ 0.02 & 0.61 $\pm$ 0.01 & 0.78 $\pm$ 0.01 & 0.22 $\pm$ 0.08 & 0.98 $\pm$ 0.00 & 0.79 $\pm$ 0.01 \\
 & Episode cost & 0.70 $\pm$ 0.22 & 0.72 $\pm$ 0.05 & 0.07 $\pm$ 0.02 & 0.67 $\pm$ 0.28 & 0.28 $\pm$ 0.06 & 0.26 $\pm$ 0.09 \\
 & Mismatch ratio & 9.69\% & 9.57\% & 3.26\% & 19.85\% & 8.68\% & 38.23\% \\
\cmidrule(lr){1-8}
\multirow{3}{*}{\ \ + projection filter} & Episode return & 0.46 $\pm$ 0.03 & 0.58 $\pm$ 0.00 & 0.77 $\pm$ 0.01 & 0.19 $\pm$ 0.06 & 0.97 $\pm$ 0.00 & 0.77 $\pm$ 0.01 \\
 & Episode cost & 0.55 $\pm$ 0.20 & 0.70 $\pm$ 0.03 & 0.10 $\pm$ 0.04 & 0.30 $\pm$ 0.07 & 0.24 $\pm$ 0.05 & 0.25 $\pm$ 0.10 \\
 & Mismatch ratio & 2.35\% & 0.28\% & 0.27\% & 2.31\% & 0.03\% & 0.07\% \\
\bottomrule
\end{tabular}
\end{adjustbox}
\vspace{3pt}
\caption{DexDICE with deployment-time projection onto the estimated safe support ($N_{\DS}=N_{\DP}=10$, 5 seeds). Normalized episode return and cost are reported as means $\pm$ standard errors; the mismatch ratio is the fraction of executed online steps whose action lies outside $\hbS_{\tau_\kappa}$.}
\label{tab:projection}
\end{table}
}

\section{Details of Real-World Robot Experiment}
\label{app:realworld}

\subsection{Hardware Specification}
\label{app:realworld_hardware}

\paragraph{Mobile platform.}
We use the Robotnik SUMMIT-XL-Steel, a four-wheel indoor mobile operated in omni-directional mode. Specifications relevant to our experiments are summarized in Table~\ref{tab:summit_specs}.

\begin{table}[hbpt!]
\centering
\small
\begin{tabular}{l l}
\toprule
Dimensions ($L \times W \times H$) & $750 \times 665 \times 510$\,mm \\
Drive &  Skid-steer / omni-directional (experiments: omni-directional) \\
Max speed & $3$\,m/s (experiments: $1.6$\,m/s, $3$\,rad/s) \\
LiDAR & Two SICK TIM571 2D scanners\\
\bottomrule
\end{tabular}
\vspace{10pt}
\caption{Specifications of the SUMMIT-XL-Steel.}
\label{tab:summit_specs}
\end{table}

\paragraph{Safety violation threshold.}
We define a safety violation as any timestep in which the closest LiDAR-measured obstacle distance falls below $d_\mathrm{SV}=0.65$\,m corresponding to the robot's bounding radius of $0.624$\,m plus a margin for localization noise. The \textit{violation rate} reported in Table~\ref{tab:realworld_results} is the per-trial fraction of timesteps below $d_\mathrm{SV}$.

\subsection{MDP Specification}

\paragraph{State.}
The state $s_t \in \mathbb{R}^{7}$ is expressed entirely in the robot body frame:
\begin{equation}
s_t \;=\; \bigl[\,
  v_x,\ v_y,\ \omega_z,\;
  \Delta x^{b},\ \Delta y^{b},\;
  \cos(\Delta\psi),\ \sin(\Delta\psi)
\,\bigr]^{\top},
\end{equation}
where $(v_x, v_y, \omega_z)$ is the body-frame linear/angular velocity and
$(\Delta x^{b}, \Delta y^{b}, \Delta\psi)$ is the goal pose relative to the
current robot pose. Letting $(x, y, \psi)$ and $(x_g, y_g, \psi_g)$ denote the
robot and goal poses in the world frame,
\begin{align}
\begin{bmatrix} \Delta x^{b} \\ \Delta y^{b} \end{bmatrix}
&= R(-\psi)\!\begin{bmatrix} x_g - x \\ y_g - y \end{bmatrix}, &
\Delta\psi &= \mathrm{wrap}_{[-\pi,\pi]}(\psi_g - \psi),
\end{align}
with $R(\theta) \in SO(2)$ the planar rotation. The yaw error
$\Delta\psi$ is encoded as $(\cos\Delta\psi, \sin\Delta\psi)$ to avoid the
$\pm\pi$ wrap-around discontinuity.

\paragraph{Observation (LiDAR).}
In addition to $s_t$, the policy receives a 2D LiDAR scan with $370$ beams over a $360^\circ$ field of view (SICK TIM571), with each ray clipped at $\ell_{\max} = 30\,\text{m}$. Following~\citet{tai2017virtual}, who used $10$ sub-sampled range readings for mapless navigation, we apply a sparse-$K$ encoding that retains only the $K = 10$ beams closest to the robot. Each retained beam $k$ is represented as $(\ell_k / \ell_{\max},\ \cos\phi_k,\ \sin\phi_k)$, where $\ell_k$ is the measured distance and $\phi_k$ is the beam angle in the body frame, yielding a permutation-invariant $3K$-dimensional set representation in the spirit of PointNet~\citep{qi2017pointnet}. This LiDAR feature is concatenated with $s_t$ as the policy input, giving an overall input dimension of $7 + 3K$.

\paragraph{Action.}
The action $a_t \in \mathbb{R}^{3}$ is a body-frame velocity command:
\begin{equation}
a_t \;=\; \bigl[\,
  v_x^{\text{cmd}},\ v_y^{\text{cmd}},\ \omega_z^{\text{cmd}}
\,\bigr]^{\top},
\qquad
\lvert v_x^{\text{cmd}}\rvert,\ \lvert v_y^{\text{cmd}}\rvert \le v_{\max},\quad
\lvert \omega_z^{\text{cmd}}\rvert \le \omega_{\max},
\end{equation}
with $v_{\max} = 1.6\,\text{m/s}$ and $\omega_{\max} = 3\,\text{rad/s}$.

\paragraph{Goal and termination.}
An episode terminates upon goal reaching, collision, or after
$T_{\max} = 500$ steps.
An episode is counted as a success when the robot reaches the goal pose within a position tolerance $\sqrt{\Delta x^2 + \Delta y^2} < 0.5\,\text{m}$ before the timeout.

\subsection{Training Setup}
\label{sec:training-setup}

\paragraph{Datasets, observation, action.}
We use the three datasets defined in Section~\ref{sec:exp_realworld} ($N_{\DP}{=}10$, $N_{\DS}{=}10$, $N_{\DU}{=}114$ including $\DP$ and $\DS$); observation and action spaces follow the MDP Specification above.

\paragraph{Baselines.}
DexDICE is compared against four baselines trained on the same datasets and spaces: \textbf{BC-Safe} (Gaussian BC on $\DS$), \textbf{FBC} (filtered BC with a discriminator-quantile filter), \textbf{DemoDICE-U} (DemoDICE with $\DU$ as the extraction set), and \textbf{FISOR-IL($\hat c$)}, the offline-IL variant of FISOR~\citep{zheng2024safe} fed with a pseudo cost from a $\DS$-vs-$(\DU\cup\DP)$ discriminator.

\paragraph{Architecture and optimization.}
Gaussian-policy methods (BC-Safe, FBC, DemoDICE-U, DexDICE) use an MLP actor with hidden sizes $[256,256]$\rev{, whose input is the concatenation of the sparse-$K$ LiDAR features and a $32$-dimensional encoding of $s_t$ produced by a two-layer MLP ($64$ and $32$ units)}; FISOR-IL($\hat c$) instead uses a DDPM actor that selects among $32$ candidate length-$8$ action chunks at inference. Each method is trained for $10^6$ Adam steps with batch size $512$ over $5$ seeds; DexDICE uses target FNR $\kappa{=}0.2$ and initial Lagrange multiplier $\lambda_0{=}5$.

\subsection{Evaluation Protocol}
\label{sec:eval-protocol}

\paragraph{Calibrated 2D simulator.}
Before real-robot deployment, candidate policies must be validated for safe and sensible behavior to avoid hardware damage or unsafe trials. We perform this validation in a 2D simulator whose first-order velocity dynamics are calibrated to the real platform from teleoperation logs and whose arena/obstacle layout matches the physical setup.

\paragraph{Trials.}
Simulator: $E{=}50$ episodes per (method, seed) under $\pm 0.20$\,m / $\pm 0.20$\,rad init-pose perturbations, terminating on goal reach, collision, or $T_\mathrm{max}{=}500$ steps ($50$\,s at $10$\,Hz); $5\times 50 = 250$ trials per method are aggregated for pre-deployment validation. Real robot: $10$ trials per method's validated checkpoint via the same $10$\,Hz inference loop, providing the main-paper results in Section~\ref{sec:exp_realworld}.

\paragraph{Metrics.}
\begin{itemize}[leftmargin=1.2em,topsep=2pt,itemsep=2pt]
\item \textbf{Success rate}: fraction of trials reaching the goal ($\|(\Delta x, \Delta y)\| < 0.50$\,m) without a collision (closest LiDAR beam $\ge d_\mathrm{coll}{=}0.15$\,m throughout).
\item \textbf{Time-to-goal}: mean episode duration over successful trials.
\item \textbf{Safety-violation rate}: fraction of timesteps with closest LiDAR beam below $d_\mathrm{SV}$ (defined in Appendix~\ref{app:realworld_hardware}), regardless of whether a collision occurs.
\end{itemize}

\subsection{Per-Trial Visualization}
\label{app:realworld_per_trial}

\begin{figure}[hpt!]
    \centering
    \includegraphics[width=1.0\linewidth]{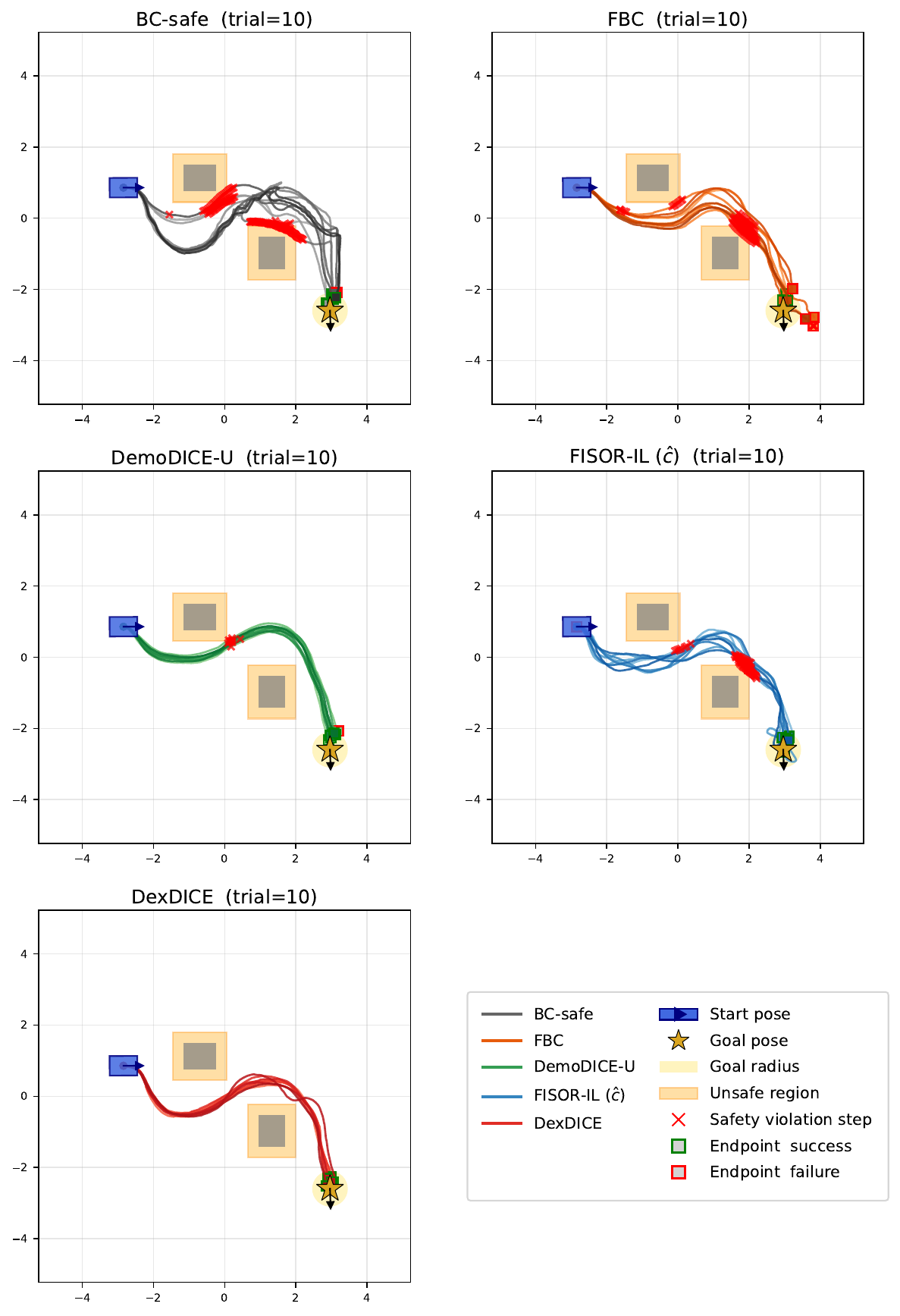}
    \caption{Visualization of per-trial trajectories for the five evaluated methods across $10$ deployment trials each. Each colored polyline represents one rollout; obstacle boxes are shown in gray.}
    \label{fig:realworld_per_trial}
\end{figure}

To complement the aggregate metrics in Table~\ref{tab:realworld_results} (Section~\ref{sec:exp_realworld}), Figure~\ref{fig:realworld_per_trial} overlays the trajectories of all 10 deployment trials per method on the navigation arena floorplan. The visualization makes the qualitative failure modes that drive the per-method differences directly inspectable.

\paragraph{Qualitative analysis.}
\textbf{BC-Safe} is the safest cloning baseline, but its conservatism slows trajectories to ${\sim}40\,\text{s}$ and stalls two trials before reaching the goal.
\textbf{FBC} matches expert speed but its support filter is too permissive on hardware, leading to obstacle clipping and collisions.
\textbf{FISOR-IL} with pseudo-labeled cost shows a similar fast-but-unsafe profile to FBC, with violations spread more uniformly across the trajectory.
\textbf{DemoDICE-U} is closest to DexDICE in both safety and speed; its two failures are timeouts driven by overly cautious detours, consistent with the simulation finding that distribution matching to $\dS$ biases the policy toward suboptimal safe routes.
\textbf{DexDICE (ours)} reaches the goal in all ten trials along nearly indistinguishable trajectories that combine $\dP$'s directness with $\dS$'s wide obstacle margins, with zero boundary violations and tight intra-method variance.

\section{Compute Resources}
\label{sec:compute_resources}

Experiments were run on (i) two workstation nodes (NVIDIA RTX~5090 32\,GiB, Intel Core Ultra~9~285K, 62\,GiB RAM), (ii) four NVIDIA DGX Spark units, and (iii) Google Cloud \texttt{n4} instances with Tesla T4 accelerators provisioned on demand for sweep parallelism. A single DexDICE training run on one re-curated DSRL task takes approximately $4$--$6$\,hours on an RTX~5090. Real-world experiments (Section~\ref{sec:exp_realworld}) ran inference on the SUMMIT-XL on-board compute and trained offline on the workstation nodes.

\section{Broader Impacts}
\label{sec:app.impact}

This paper aims to advance safe offline imitation learning for safety-critical applications. While our framework is designed to ensure safety, it could in principle be misused to circumvent safety requirements. Deployment in domains such as autonomous driving or healthcare without rigorous validation may cause harm; responsible use lies with practitioners and deploying organizations.

%% file: neurips_2026.bbl
\begin{thebibliography}{70}
\providecommand{\natexlab}[1]{#1}
\providecommand{\url}[1]{\texttt{#1}}
\expandafter\ifx\csname urlstyle\endcsname\relax
  \providecommand{\doi}[1]{doi: #1}\else
  \providecommand{\doi}{doi: \begingroup \urlstyle{rm}\Url}\fi

\bibitem[Altman(1999)]{altman1999constrained}
Eitan Altman.
\newblock \emph{Constrained Markov decision processes}.
\newblock Chapman and Hall/CRC, 1999.

\bibitem[Ames et~al.(2019)Ames, Coogan, Egerstedt, Notomista, Sreenath, and
  Tabuada]{ames2019control}
Aaron~D. Ames, Samuel Coogan, Magnus Egerstedt, Gennaro Notomista, Koushil
  Sreenath, and Paulo Tabuada.
\newblock Control barrier functions: Theory and applications.
\newblock In \emph{18th European Control Conference (ECC)}, 2019.

\bibitem[Bansal et~al.(2019)Bansal, Krizhevsky, and
  Ogale]{bansal2018chauffeurnet}
Mayank Bansal, Alex Krizhevsky, and Abhijit~S. Ogale.
\newblock {ChauffeurNet: L}earning to drive by imitating the best and
  synthesizing the worst.
\newblock In \emph{Robotics: Science and Systems XV}, 2019.

\bibitem[Bansal et~al.(2017)Bansal, Chen, Herbert, and
  Tomlin]{bansal2017hamilton}
Somil Bansal, Mo~Chen, Sylvia Herbert, and Claire~J. Tomlin.
\newblock {Hamilton--Jacobi} reachability: A brief overview and recent
  advances.
\newblock In \emph{IEEE 56th Annual Conference on Decision and Control (CDC)},
  2017.

\bibitem[Bekker and Davis(2020)]{bekker2020learning}
Jessa Bekker and Jesse Davis.
\newblock Learning from positive and unlabeled data: A survey.
\newblock \emph{Machine Learning}, 109, 2020.

\bibitem[Bishop(2006)]{bishop2006pattern}
Christopher~M. Bishop.
\newblock \emph{Pattern Recognition and Machine Learning}.
\newblock Springer, 2006.

\bibitem[Boyd and Vandenberghe(2004)]{boyd2004convex}
Stephen~P Boyd and Lieven Vandenberghe.
\newblock \emph{Convex optimization}.
\newblock Cambridge university press, 2004.

\bibitem[Breunig et~al.(2000)Breunig, Kriegel, Ng, and Sander]{breunig2000lof}
Markus~M Breunig, Hans-Peter Kriegel, Raymond~T Ng, and J{\"o}rg Sander.
\newblock {LOF: I}dentifying density-based local outliers.
\newblock In \emph{Proceedings of the 2000 ACM SIGMOD International Conference
  on Management of Data}, 2000.

\bibitem[Burnwal et~al.(2026)Burnwal, Bhatt, and Ravindran]{burnwal2026safemil}
Returaj Burnwal, Nirav~Pravinbhai Bhatt, and Balaraman Ravindran.
\newblock {SafeMIL: L}earning offline safe imitation policy from non-preferred
  trajectories.
\newblock In \emph{Proceedings of the AAAI Conference on Artificial
  Intelligence}, volume~40, 2026.

\bibitem[Campolettano et~al.(2025)Campolettano, Kusano, and
  Victor]{campolettano2025potential}
Eamon~T Campolettano, Kristofer~D Kusano, and Trent Victor.
\newblock Potential safety benefits associated with speed limit compliance in
  {S}an {F}rancisco and {P}hoenix.
\newblock \emph{Traffic Injury Prevention}, 26, 2025.

\bibitem[Cao et~al.(2025)Cao, Joa, and Borrelli]{cao2025simple}
Shengfan Cao, Eunhyek Joa, and Francesco Borrelli.
\newblock A simple approach to constraint-aware imitation learning with
  application to autonomous racing.
\newblock In \emph{IEEE/RSJ International Conference on Intelligent Robots and
  Systems (IROS)}, 2025.

\bibitem[Chen et~al.(2021)Chen, Lu, Rajeswaran, Lee, Grover, Laskin, Abbeel,
  Srinivas, and Mordatch]{chen2021decision}
Lili Chen, Kevin Lu, Aravind Rajeswaran, Kimin Lee, Aditya Grover, Misha
  Laskin, Pieter Abbeel, Aravind Srinivas, and Igor Mordatch.
\newblock Decision transformer: Reinforcement learning via sequence modeling.
\newblock In \emph{Advances in Neural Information Processing Systems},
  volume~34, 2021.

\bibitem[Christoffel et~al.(2016)Christoffel, Niu, and
  Sugiyama]{pmlr-v45-Christoffel15}
Marthinus Christoffel, Gang Niu, and Masashi Sugiyama.
\newblock Class-prior estimation for learning from positive and unlabeled data.
\newblock In \emph{Proceedings of the 7th Asian Conference on Machine
  Learning}, volume~45, 2016.

\bibitem[Ciftci et~al.(2025)Ciftci, Chiu, Feng, Sukhatme, and
  Bansal]{ciftci2025safe}
Yusuf~Umut Ciftci, Darren Chiu, Zeyuan Feng, Gaurav~S Sukhatme, and Somil
  Bansal.
\newblock {SAFE-GIL: S}afety guided imitation learning for robotic systems.
\newblock In \emph{IEEE International Conference on Robotics and Automation
  (ICRA)}, 2025.

\bibitem[Dadashi et~al.(2021)Dadashi, Hussenot, Geist, and
  Pietquin]{dadashiprimal}
Robert Dadashi, Leonard Hussenot, Matthieu Geist, and Olivier Pietquin.
\newblock Primal {W}asserstein imitation learning.
\newblock In \emph{The Ninth International Conference on Learning
  Representations}, 2021.

\bibitem[Elkan and Noto(2008)]{elkan2008learning}
Charles Elkan and Keith Noto.
\newblock Learning classifiers from only positive and unlabeled data.
\newblock In \emph{Proceedings of the 14th ACM SIGKDD International Conference
  on Knowledge Discovery and Data Mining}, 2008.

\bibitem[Fang et~al.(2019)Fang, Jia, Guo, Xu, Wen, and Sun]{fang2019survey}
Bin Fang, Shidong Jia, Di~Guo, Muhua Xu, Shuhuan Wen, and Fuchun Sun.
\newblock Survey of imitation learning for robotic manipulation.
\newblock \emph{International Journal of Intelligent Robotics and
  Applications}, 3, 2019.

\bibitem[Gong et~al.(2025)Gong, Kumar, and Varakantham]{yang2024offline}
Ze~Gong, Akshat Kumar, and Pradeep Varakantham.
\newblock Offline safe reinforcement learning using trajectory classification.
\newblock In \emph{Proceedings of the AAAI Conference on Artificial
  Intelligence}, volume~39, 2025.

\bibitem[Hawke et~al.(2020)Hawke, Shen, Gurau, Sharma, Reda, Nikolov, Mazur,
  Micklethwaite, Griffiths, Shah, et~al.]{hawke2020urban}
Jeffrey Hawke, Richard Shen, Corina Gurau, Siddharth Sharma, Daniele Reda,
  Nikolay Nikolov, Przemys{\l}aw Mazur, Sean Micklethwaite, Nicolas Griffiths,
  Amar Shah, et~al.
\newblock Urban driving with conditional imitation learning.
\newblock In \emph{IEEE International Conference on Robotics and Automation
  (ICRA)}, 2020.

\bibitem[Hido et~al.(2011)Hido, Tsuboi, Kashima, Sugiyama, and
  Kanamori]{hido2011statistical}
Shohei Hido, Yuta Tsuboi, Hisashi Kashima, Masashi Sugiyama, and Takafumi
  Kanamori.
\newblock Statistical outlier detection using direct density ratio estimation.
\newblock \emph{Knowledge and Information Systems}, 26, 2011.

\bibitem[Ho and Ermon(2016)]{ho2016generative}
Jonathan Ho and Stefano Ermon.
\newblock Generative adversarial imitation learning.
\newblock In \emph{Advances in Neural Information Processing Systems},
  volume~29, 2016.

\bibitem[Hoang et~al.(2024)Hoang, Mai, and Varakantham]{hoang2024sprinql}
Huy Hoang, Tien Mai, and Pradeep Varakantham.
\newblock {SPRINQL: S}ub-optimal demonstrations driven offline imitation
  learning.
\newblock In \emph{Advances in Neural Information Processing Systems},
  volume~37, 2024.

\bibitem[Hoang et~al.(2025)Hoang, Mai, and Varakantham]{hoang2025noexperts}
Huy Hoang, Tien Mai, and Pradeep Varakantham.
\newblock No experts, no problem: Avoidance learning from bad demonstrations.
\newblock In \emph{Advances in Neural Information Processing Systems},
  volume~38, 2025.

\bibitem[Hugessen et~al.(2024)Hugessen, Wiltzer, and
  Berseth]{hugessen2024simplifying}
Adriana Hugessen, Harley Wiltzer, and Glen Berseth.
\newblock Simplifying constraint inference with inverse reinforcement learning.
\newblock In \emph{Advances in Neural Information Processing Systems},
  volume~37, 2024.

\bibitem[Hussein et~al.(2017)Hussein, Gaber, Elyan, and
  Jayne]{hussein2017imitation}
Ahmed Hussein, Mohamed~Medhat Gaber, Eyad Elyan, and Chrisina Jayne.
\newblock Imitation learning: A survey of learning methods.
\newblock \emph{ACM Computing Surveys (CSUR)}, 50, 2017.

\bibitem[Jang et~al.(2023)Jang, Kim, Lee, Sohn, Kim, Lee, and
  Lee]{jang2023safedice}
Youngsoo Jang, Geon-Hyeong Kim, Jongmin Lee, Sungryull Sohn, Byoungjip Kim,
  Honglak Lee, and Moontae Lee.
\newblock {SafeDICE}: Offline safe imitation learning with non-preferred
  demonstrations.
\newblock In \emph{Advances in Neural Information Processing Systems},
  volume~36, 2023.

\bibitem[Jarrett et~al.(2020)Jarrett, Bica, and van~der
  Schaar]{jarrett2020strictly}
Daniel Jarrett, Ioana Bica, and Mihaela van~der Schaar.
\newblock Strictly batch imitation learning by energy-based distribution
  matching.
\newblock In \emph{Advances in Neural Information Processing Systems},
  volume~33, 2020.

\bibitem[Ji et~al.(2023)Ji, Zhang, Zhou, Pan, Huang, Sun, Geng, Zhong, Dai, and
  Yang]{ji2023safety}
Jiaming Ji, Borong Zhang, Jiayi Zhou, Xuehai Pan, Weidong Huang, Ruiyang Sun,
  Yiran Geng, Yifan Zhong, Josef Dai, and Yaodong Yang.
\newblock {Safety Gymnasium: A} unified safe reinforcement learning benchmark.
\newblock In \emph{Advances in Neural Information Processing Systems},
  volume~36, 2023.

\bibitem[Ji et~al.(2024)Ji, Zhou, Zhang, Dai, Pan, Sun, Huang, Geng, Liu, and
  Yang]{ji2024omnisafe}
Jiaming Ji, Jiayi Zhou, Borong Zhang, Juntao Dai, Xuehai Pan, Ruiyang Sun,
  Weidong Huang, Yiran Geng, Mickel Liu, and Yaodong Yang.
\newblock {OmniSafe: A}n infrastructure for accelerating safe reinforcement
  learning research.
\newblock \emph{Journal of Machine Learning Research}, 25, 2024.

\bibitem[Johns(2021)]{johns2021coarse}
Edward Johns.
\newblock Coarse-to-fine imitation learning: Robot manipulation from a single
  demonstration.
\newblock In \emph{IEEE International Conference on Robotics and Automation
  (ICRA)}, 2021.

\bibitem[Kim et~al.(2022)Kim, Seo, Lee, Jeon, Hwang, Yang, and
  Kim]{kim2022demodice}
Geon-Hyeong Kim, Seokin Seo, Jongmin Lee, Wonseok Jeon, HyeongJoo Hwang,
  Hongseok Yang, and Kee-Eung Kim.
\newblock {DemoDICE}: Offline imitation learning with supplementary imperfect
  demonstrations.
\newblock In \emph{The Tenth International Conference on Learning
  Representations}, 2022.

\bibitem[Kim and Park(2018)]{kim2018imitation}
Kee-Eung Kim and Hyun~Soo Park.
\newblock Imitation learning via kernel mean embedding.
\newblock In \emph{Proceedings of the AAAI Conference on Artificial
  Intelligence}, volume~32, 2018.

\bibitem[Kim et~al.(2023)Kim, Swamy, Liu, Zhao, Choudhury, and
  Wu]{kim2023learning}
Konwoo Kim, Gokul Swamy, Zuxin Liu, Ding Zhao, Sanjiban Choudhury, and Steven~Z
  Wu.
\newblock Learning shared safety constraints from multi-task demonstrations.
\newblock In \emph{Advances in Neural Information Processing Systems},
  volume~36, 2023.

\bibitem[Komorowski et~al.(2018)Komorowski, Celi, Badawi, Gordon, and
  Faisal]{komorowski2018artificial}
Matthieu Komorowski, Leo~A Celi, Omar Badawi, Anthony~C Gordon, and A~Aldo
  Faisal.
\newblock The artificial intelligence clinician learns optimal treatment
  strategies for sepsis in intensive care.
\newblock \emph{Nature Medicine}, 24, 2018.

\bibitem[Kwon et~al.(2024)Kwon, Lee, Lee, and Oh]{kwon2024safe}
Hyeokjin Kwon, Gunmin Lee, Junseo Lee, and Songhwai Oh.
\newblock {Safe CoR}: A dual-expert approach to integrating imitation learning
  and safe reinforcement learning using constraint rewards.
\newblock In \emph{IEEE/RSJ International Conference on Intelligent Robots and
  Systems (IROS)}, 2024.

\bibitem[Lasota et~al.(2014)Lasota, Rossano, and Shah]{lasota2014toward}
Przemyslaw~A Lasota, Gregory~F Rossano, and Julie~A Shah.
\newblock Toward safe close-proximity human-robot interaction with standard
  industrial robots.
\newblock In \emph{IEEE International Conference on Automation Science and
  Engineering (CASE)}, 2014.

\bibitem[Le~Mero et~al.(2022)Le~Mero, Yi, Dianati, and
  Mouzakitis]{le2022survey}
Luc Le~Mero, Dewei Yi, Mehrdad Dianati, and Alexandros Mouzakitis.
\newblock A survey on imitation learning techniques for end-to-end autonomous
  vehicles.
\newblock \emph{IEEE Transactions on Intelligent Transportation Systems}, 23,
  2022.

\bibitem[Lee et~al.(2021)Lee, Jeon, Lee, Pineau, and Kim]{lee2021optidice}
Jongmin Lee, Wonseok Jeon, Byungjun Lee, Joelle Pineau, and Kee-Eung Kim.
\newblock {OptiDICE}: Offline policy optimization via stationary distribution
  correction estimation.
\newblock In \emph{Proceedings of the 38th International Conference on Machine
  Learning}, 2021.

\bibitem[Lee et~al.(2022)Lee, Paduraru, Mankowitz, Heess, Precup, Kim, and
  Guez]{leecoptidice}
Jongmin Lee, Cosmin Paduraru, Daniel~J Mankowitz, Nicolas Heess, Doina Precup,
  Kee-Eung Kim, and Arthur Guez.
\newblock {COptiDICE}: Offline constrained reinforcement learning via
  stationary distribution correction estimation.
\newblock In \emph{The Tenth International Conference on Learning
  Representations}, 2022.

\bibitem[Liu et~al.(2025{\natexlab{a}})Liu, Xu, Liu, Gaurav, Subramanian, and
  Poupart]{liu2024comprehensive}
Guiliang Liu, Sheng Xu, Shicheng Liu, Ashish Gaurav, Sriram~Ganapathi
  Subramanian, and Pascal Poupart.
\newblock A comprehensive survey on inverse constrained reinforcement learning:
  Definitions, progress and challenges.
\newblock \emph{Transactions on Machine Learning Research}, 2025{\natexlab{a}}.

\bibitem[Liu et~al.(2025{\natexlab{b}})Liu, Fu, Tang, Chen, and
  Dong]{liu2025pngail}
Qiang Liu, Huiqiao Fu, Kaiqiang Tang, Chunlin Chen, and Daoyi Dong.
\newblock {PN-GAIL}: Leveraging non-optimal information from imperfect
  demonstrations.
\newblock In \emph{The Thirteenth International Conference on Learning
  Representations}, 2025{\natexlab{b}}.

\bibitem[Liu et~al.(2025{\natexlab{c}})Liu, Li, and Zhang]{liu2025c2iql}
Zifan Liu, Xinran Li, and Jun Zhang.
\newblock {C2IQL: C}onstraint-conditioned implicit {Q}-learning for safe
  offline reinforcement learning.
\newblock In \emph{Proceedings of the 42nd International Conference on Machine
  Learning}, 2025{\natexlab{c}}.

\bibitem[Liu et~al.(2023)Liu, Guo, Yao, Cen, Yu, Zhang, and
  Zhao]{liu2023constrained}
Zuxin Liu, Zijian Guo, Yihang Yao, Zhepeng Cen, Wenhao Yu, Tingnan Zhang, and
  Ding Zhao.
\newblock Constrained decision transformer for offline safe reinforcement
  learning.
\newblock In \emph{Proceedings of the 40th International Conference on Machine
  Learning}, 2023.

\bibitem[Liu et~al.(2024)Liu, Guo, Lin, Yao, Zhu, Cen, Hu, Yu, Zhang, Tan,
  et~al.]{liu2024datasets}
Zuxin Liu, Zijian Guo, Haohong Lin, Yihang Yao, Jiacheng Zhu, Zhepeng Cen,
  Hanjiang Hu, Wenhao Yu, Tingnan Zhang, Jie Tan, et~al.
\newblock Datasets and benchmarks for offline safe reinforcement learning.
\newblock \emph{Journal of Data-centric Machine Learning Research}, 2024.

\bibitem[Ma et~al.(2022)Ma, Shen, Jayaraman, and Bastani]{ma2022versatile}
Yecheng Ma, Andrew Shen, Dinesh Jayaraman, and Osbert Bastani.
\newblock Versatile offline imitation from observations and examples via
  regularized state-occupancy matching.
\newblock In \emph{Proceedings of the 39th International Conference on Machine
  Learning}, 2022.

\bibitem[Malik et~al.(2021)Malik, Anwar, Aghasi, and Ahmed]{malik2021inverse}
Shehryar Malik, Usman Anwar, Alireza Aghasi, and Ali Ahmed.
\newblock Inverse constrained reinforcement learning.
\newblock In \emph{Proceedings of the 38th International Conference on Machine
  Learning}, 2021.

\bibitem[Mandlekar et~al.(2021)Mandlekar, Xu, Wong, Nasiriany, Wang, Kulkarni,
  Fei-Fei, Savarese, Zhu, and Mart\'in-Mart\'in]{mandlekar2021matters}
Ajay Mandlekar, Danfei Xu, Josiah Wong, Soroush Nasiriany, Chen Wang, Rohun
  Kulkarni, Li~Fei-Fei, Silvio Savarese, Yuke Zhu, and Roberto
  Mart\'in-Mart\'in.
\newblock What matters in learning from offline human demonstrations for robot
  manipulation.
\newblock In \emph{Proceedings of the 5th Conference on Robot Learning}, volume
  164, 2021.

\bibitem[Nachum et~al.(2019)Nachum, Chow, Dai, and Li]{nachum2019dualdice}
Ofir Nachum, Yinlam Chow, Bo~Dai, and Lihong Li.
\newblock {DualDICE: B}ehavior-agnostic estimation of discounted stationary
  distribution corrections.
\newblock In \emph{Advances in Neural Information Processing Systems},
  volume~32, 2019.

\bibitem[Nemati et~al.(2016)Nemati, Ghassemi, and Clifford]{nemati2016optimal}
Shamim Nemati, Mohammad~M Ghassemi, and Gari~D Clifford.
\newblock Optimal medication dosing from suboptimal clinical examples: A deep
  reinforcement learning approach.
\newblock In \emph{38th Annual International Conference of the IEEE Engineering
  in Medicine and Biology Society (EMBC)}, 2016.

\bibitem[Qi et~al.(2017)Qi, Su, Mo, and Guibas]{qi2017pointnet}
Charles~R Qi, Hao Su, Kaichun Mo, and Leonidas~J Guibas.
\newblock {PointNet: D}eep learning on point sets for 3d classification and
  segmentation.
\newblock In \emph{Proceedings of the IEEE Conference on Computer Vision and
  Pattern Recognition}, 2017.

\bibitem[Quan et~al.(2024)Quan, Xu, and Liu]{quan2024learning}
Guorui Quan, Zhiqiang Xu, and Guiliang Liu.
\newblock Learning constraints from offline demonstrations via superior
  distribution correction estimation.
\newblock In \emph{Proceedings of the 41st International Conference on Machine
  Learning}, 2024.

\bibitem[Ray et~al.(2019)Ray, Achiam, and Amodei]{ray2019benchmarking}
Alex Ray, Joshua Achiam, and Dario Amodei.
\newblock Benchmarking safe exploration in deep reinforcement learning.
\newblock \emph{arXiv preprint arXiv:1910.01708}, 2019.

\bibitem[Ross and Bagnell(2010)]{pmlr-v9-ross10a}
Stephane Ross and Drew Bagnell.
\newblock Efficient reductions for imitation learning.
\newblock In \emph{Proceedings of the 13th International Conference on
  Artificial Intelligence and Statistics}, 2010.

\bibitem[Ruff et~al.(2018)Ruff, Vandermeulen, Goernitz, Deecke, Siddiqui,
  Binder, M{\"u}ller, and Kloft]{ruff2018deep}
Lukas Ruff, Robert Vandermeulen, Nico Goernitz, Lucas Deecke, Shoaib~Ahmed
  Siddiqui, Alexander Binder, Emmanuel M{\"u}ller, and Marius Kloft.
\newblock Deep one-class classification.
\newblock In \emph{Proceedings of the 35th International Conference on Machine
  Learning}, 2018.

\bibitem[Sch{\"o}lkopf et~al.(2001)Sch{\"o}lkopf, Platt, Shawe-Taylor, Smola,
  and Williamson]{scholkopf2001estimating}
Bernhard Sch{\"o}lkopf, John~C Platt, John Shawe-Taylor, Alex~J Smola, and
  Robert~C Williamson.
\newblock Estimating the support of a high-dimensional distribution.
\newblock \emph{Neural Computation}, 13, 2001.

\bibitem[Schulman et~al.(2017)Schulman, Wolski, Dhariwal, Radford, and
  Klimov]{schulman2017proximal}
John Schulman, Filip Wolski, Prafulla Dhariwal, Alec Radford, and Oleg Klimov.
\newblock Proximal policy optimization algorithms.
\newblock \emph{arXiv preprint arXiv:1707.06347}, 2017.

\bibitem[Scott and Blanchard(2009)]{scott2009novelty}
Clayton Scott and Gilles Blanchard.
\newblock Novelty detection: Unlabeled data definitely help.
\newblock In \emph{Proceedings of the 12th International Conference on
  Artificial Intelligence and Statistics}, 2009.

\bibitem[Seo et~al.(2024)Seo, Lee, Lee, Hwang, Yang, and
  Kim]{seo2024mitigating}
Seokin Seo, Byung-Jun Lee, Jongmin Lee, HyeongJoo Hwang, Hongseok Yang, and
  Kee-Eung Kim.
\newblock Mitigating covariate shift in behavioral cloning via robust
  stationary distribution correction.
\newblock In \emph{Advances in Neural Information Processing Systems},
  volume~37, 2024.

\bibitem[Shao et~al.(2024)Shao, Varakantham, and Cheng]{shao2024imitating}
Qian Shao, Pradeep Varakantham, and Shih-Fen Cheng.
\newblock Imitating cost-constrained behaviors in reinforcement learning.
\newblock In \emph{Proceedings of the International Conference on Automated
  Planning and Scheduling}, volume~34, 2024.

\bibitem[Subramanian et~al.(2024)Subramanian, Liu, Elmahgiubi, Rezaee, and
  Poupart]{subramanian2024confidence}
Sriram~Ganapathi Subramanian, Guiliang Liu, Mohammed Elmahgiubi, Kasra Rezaee,
  and Pascal Poupart.
\newblock Confidence aware inverse constrained reinforcement learning.
\newblock In \emph{Proceedings of the 41st International Conference on Machine
  Learning}, 2024.

\bibitem[Tai et~al.(2017)Tai, Paolo, and Liu]{tai2017virtual}
Lei Tai, Giuseppe Paolo, and Ming Liu.
\newblock Virtual-to-real deep reinforcement learning: Continuous control of
  mobile robots for mapless navigation.
\newblock In \emph{IEEE/RSJ International Conference on Intelligent Robots and
  Systems (IROS)}, 2017.

\bibitem[Tong(2013)]{tong2013plugin}
Xin Tong.
\newblock A plug-in approach to {N}eyman-{P}earson classification.
\newblock \emph{Journal of Machine Learning Research}, 14, 2013.

\bibitem[Xie et~al.(2020)Xie, Chowdhury, De~Paolis~Kaluza, Zhao, Wong, and
  Yu]{xie2020deep}
Fan Xie, Alexander Chowdhury, M~De~Paolis~Kaluza, Linfeng Zhao, Lawson Wong,
  and Rose Yu.
\newblock Deep imitation learning for bimanual robotic manipulation.
\newblock In \emph{Advances in Neural Information Processing Systems},
  volume~33, 2020.

\bibitem[Xu et~al.(2022)Xu, Zhan, Yin, and Qin]{xu2022discriminator}
Haoran Xu, Xianyuan Zhan, Honglei Yin, and Huiling Qin.
\newblock Discriminator-weighted offline imitation learning from suboptimal
  demonstrations.
\newblock In \emph{Proceedings of the 39th International Conference on Machine
  Learning}, 2022.

\bibitem[Yin et~al.(2022)Yin, Seiler, Jin, and Arcak]{yin2021imitation}
He~Yin, Peter Seiler, Ming Jin, and Murat Arcak.
\newblock Imitation learning with stability and safety guarantees.
\newblock \emph{IEEE Control Systems Letters}, 6, 2022.

\bibitem[Yu et~al.(2023)Yu, Yu, Song, Neiswanger, and Ermon]{yu2023offline}
Lantao Yu, Tianhe Yu, Jiaming Song, Willie Neiswanger, and Stefano Ermon.
\newblock Offline imitation learning with suboptimal demonstrations via relaxed
  distribution matching.
\newblock In \emph{Proceedings of the AAAI Conference on Artificial
  Intelligence}, volume~37, 2023.

\bibitem[Zare et~al.(2024)Zare, Kebria, Khosravi, and
  Nahavandi]{zare2024survey}
Maryam Zare, Parham~M Kebria, Abbas Khosravi, and Saeid Nahavandi.
\newblock A survey of imitation learning: Algorithms, recent developments, and
  challenges.
\newblock \emph{IEEE Transactions on Cybernetics}, 54, 2024.

\bibitem[Zhang et~al.(2020)Zhang, Li, Zhang, and Zhang]{zhang2020f}
Xin Zhang, Yanhua Li, Ziming Zhang, and Zhi-Li Zhang.
\newblock {$f$-GAIL}: Learning $f$-divergence for generative adversarial
  imitation learning.
\newblock In \emph{Advances in Neural Information Processing Systems},
  volume~33, 2020.

\bibitem[Zhao et~al.(2023)Zhao, Kumar, Levine, and Finn]{zhao2023learning}
Tony Zhao, Vikash Kumar, Sergey Levine, and Chelsea Finn.
\newblock Learning fine-grained bimanual manipulation with low-cost hardware.
\newblock \emph{Robotics: Science and Systems XIX}, 2023.

\bibitem[Zheng et~al.(2024)Zheng, Li, Yu, Yang, Li, Zhan, and
  Liu]{zheng2024safe}
Yinan Zheng, Jianxiong Li, Dongjie Yu, Yujie Yang, Shengbo~Eben Li, Xianyuan
  Zhan, and Jingjing Liu.
\newblock Safe offline reinforcement learning with feasibility-guided diffusion
  model.
\newblock In \emph{The Twelfth International Conference on Learning
  Representations}, 2024.

\end{thebibliography}
